\documentclass{article}

\makeatletter
\let\neuripsOriginalInputPath\input@path
\providecommand{\input@path}{}
\g@addto@macro\input@path{{neurips (arxiv)/}}
\makeatother

\usepackage[main, preprint]{neurips_2026}

\usepackage[utf8]{inputenc} %
\usepackage[T1]{fontenc}    %
\usepackage{url}            %
\usepackage{booktabs}       %
\usepackage{amsfonts}       %
\usepackage{nicefrac}       %
\usepackage{microtype}      %
\usepackage{xcolor}         %
\usepackage{tabularx}

\usepackage{titletoc}
\usepackage{caption}

\usepackage{amsmath, amsthm, amssymb} %
\usepackage{mathtools} %
\usepackage{thmtools} %
\makeatletter
\@ifundefined{newcounteralias}{}{%
  \renewcommand\thmt@autorefsetup{%
    \@xa\def\csname\thmt@envname autorefname\@xa\endcsname\@xa{\thmt@thmname}%
  }%
}
\makeatother
\usepackage{mleftright} %
\usepackage{bm} %

\usepackage[shortlabels]{enumitem} %

\usepackage{algorithm} %
\usepackage{algpseudocode} %

\usepackage{dsfont} %

\usepackage[hypertexnames=false]{hyperref}
\usepackage[capitalise]{cleveref} %
\usepackage{crossreftools}
\ExplSyntaxOn
\newcommand{\bookmarkCref}[1]{%
    \text_titlecase_first:n{\crtcrefcounter{#1}}~\crtrefnumber{#1}%
}
\ExplSyntaxOff
\Crefname{assumption}{Assumption}{Assumptions}
\Crefname{setting}{Setting}{Settings}

\crefname{section}{Section}{Sections}
\Crefname{section}{Section}{Sections}
\crefname{appendix}{Appendix}{Appendices}
\Crefname{appendix}{Appendix}{Appendices}

\theoremstyle{plain}
\newtheorem{lemma}{Lemma}[section]
\newtheorem{theorem}[lemma]{Theorem}
\newtheorem{corollary}[lemma]{Corollary}
\newtheorem{proposition}[lemma]{Proposition}

\theoremstyle{definition}
\newtheorem{definition}[lemma]{Definition}

\DeclareMathOperator*{\argmin}{arg\,min}
\DeclareMathOperator*{\argmax}{arg\,max} 
\DeclareMathOperator{\conv}{conv}

\let\vec\relax
\DeclareMathOperator{\vec}{vec}
\DeclareMathOperator{\E}{\mathbb{E}}

\DeclareMathOperator{\tr}{tr}

\DeclareMathOperator{\diag}{diag}

\DeclareMathOperator{\vnull}{null}
\DeclareMathOperator{\rank}{rank}

\DeclareMathOperator{\range}{range}

\newcommand{\R}{\mathbb{R}}

\newcommand{\PP}{\mathbb{P}}

\newcommand{\OO}{\mathbb{O}}

\newcommand{\eps}{\varepsilon}
\newcommand{\dd}{\mathrm{d}} %

\newcommand{\defeq}{\coloneqq} %

\mathchardef\mhyphen="2D

\newcommand{\DRO}{\mathsf{DRO}}

\newcommand{\MRO}{\mathsf{MRO}}

\newcommand{\ERM}{\mathsf{ERM}}

\newcommand{\SNR}{\mathsf{SNR}}

\newcommand{\bA}{\mathbf{A}}
\newcommand{\bB}{\mathbf{B}}
\newcommand{\bC}{\mathbf{C}}
\newcommand{\bD}{\mathbf{D}}
\newcommand{\bE}{\mathbf{E}}

\newcommand{\bG}{\mathbf{G}}
\newcommand{\bH}{\mathbf{H}}
\newcommand{\bI}{\mathbf{I}}
\newcommand{\bJ}{\mathbf{J}}
\newcommand{\bK}{\mathbf{K}}

\newcommand{\bM}{\mathbf{M}}
\newcommand{\bN}{\mathbf{N}}
\newcommand{\bO}{\mathbf{O}}
\newcommand{\bP}{\mathbf{P}}
\newcommand{\bQ}{\mathbf{Q}}

\newcommand{\bS}{\mathbf{S}}
\newcommand{\bT}{\mathbf{T}}
\newcommand{\bU}{\mathbf{U}}
\newcommand{\bV}{\mathbf{V}}
\newcommand{\bW}{\mathbf{W}}
\newcommand{\bX}{\mathbf{X}}
\newcommand{\bY}{\mathbf{Y}}
\newcommand{\bZ}{\mathbf{Z}}

\newcommand{\bSigma}{\mathbf{\Sigma}}

\newcommand{\bZero}{\mathbf{0}}
\newcommand{\bPi}{\mathbf{\Pi}}
\newcommand{\bDelta}{\mathbf{\Delta}}

\newcommand{\Sym}{\mathbb{S}}
\newcommand{\PSD}{\mathbb{S}_+}

\newcommand{\op}{\mathrm{op}}
\newcommand{\reff}{r_{\mathrm{eff}}}

\newcommand{\Wtwo}{\mathsf{W}_2}

\newcommand{\Frob}{\mathrm{F}}

\newcommand{\sB}{\mathsf{B}}

\newcommand{\sV}{\mathsf{V}}

\newcommand{\cA}{\mathcal{A}}
\newcommand{\cB}{\mathcal{B}}
\newcommand{\cC}{\mathcal{C}}

\newcommand{\cE}{\mathcal{E}}

\newcommand{\cG}{\mathcal{G}}

\newcommand{\cK}{\mathcal{K}}

\newcommand{\cN}{\mathcal{N}}

\newcommand{\cP}{\mathcal{P}}
\newcommand{\cQ}{\mathcal{Q}}

\def\left#1{\mleft#1}
\def\right#1{\mright#1}

\makeatletter
\let\input@path\neuripsOriginalInputPath
\makeatother

\title{Regret-Optimal Wasserstein-Robust\\ Linear Regression}

\author{%
  Sloan Nietert\\
  EPFL\\
  \texttt{sloan.nietert@epfl.ch}
  \\
  \And
  Daniel Kuhn\\
  EPFL\\
  \texttt{daniel.kuhn@epfl.ch}
}

\begin{document}

\maketitle

\begin{abstract}
Distributionally robust optimization (DRO) minimizes worst-case risk under distributional uncertainty, but its decisions can be suboptimal in terms of their excess risk, or regret. We study this gap for multivariate least-squares regression, assuming that the observed source law and unknown target law differ by an adversarial 2-Wasserstein perturbation of size $\eps$. We first identify the asymptotic complexity of robust decision-making in the $\eps \to 0$ limit, characterized by a problem-specific condition number $\kappa$. Here, a wide spectrum of simple algorithms, including DRO, achieve the instance-optimal rate of $O(\kappa \eps^2)$. For fixed $\eps > 0$, we prove that DRO still achieves the minimax rate if the problem has bounded residual effective rank. On the other hand, we identify significant failure modes where DRO is provably suboptimal by dimension-dependent factors. Fortunately, we prove that two alternative procedures, minimax regret optimization (MRO) and spectral-norm regularization (SNR), attain the minimax rate. We derive an exact convex reformulation of MRO and an accelerated
proximal algorithm for SNR, along with high-probability finite-sample
guarantees. We support these theoretical results with numerical experiments separating the performance of MRO and SNR from DRO.
\end{abstract}

\section{Introduction}
\label{sec:intro}

Distributionally robust optimization (DRO) is a flexible tool for hedging against model misspecification in ML, statistics, and operations research \citep{rahimian2019distributionally,lin2022distributionally,kuhn2025distributionally}. Wasserstein DRO is a well-studied variant that minimizes worst-case risk over all Wasserstein perturbations of the observed data distribution. For many tasks, this objective admits dual formulations, tractable implementations, and regularization interpretations \citep{kuhn2019wasserstein, gao2023distributionally, gao2024wasserstein}. Model misspecification comes in many flavors, e.g., sampling error, distribution shift, and data poisoning. In this work, we take an adversarial view: the learner observes a source law and seeks a predictor that performs well under a nearby but otherwise unrestricted target law. As we will see, robust risk at the source (the objective minimized by DRO) is fundamentally distinct from decision quality at the target. Our motivating question is the following:

\begin{quote}
    \centering\emph{When does Wasserstein DRO make good decisions in the face of adversarial data perturbations, and when can we efficiently outperform it?}
\end{quote}

We examine this for the setting of multivariate linear regression.
Fix dimensions $p,m\geq1$ and let $\cA\defeq\R^{m\times p}$ be the class of linear predictors.
Set $d \defeq p + m$ and write $\cP_2(\R^d)$ for the laws of
$Z=(X,Y)\in\R^p\times\R^m$ with finite second moment.
For $P\in\cP_2(\R^d)$, the least-squares risk of $\bA\in\cA$ is
$L(\bA,P)\defeq\E_P\|\bA X-Y\|^2$, and its \emph{excess risk},
or (ex-ante) \emph{regret}, is
\[
    R(\bA,P)
    \defeq L(\bA,P)-\inf_{\bA'\in\cA}L(\bA',P).
\]
This regret is ex-ante, rather than ex-post, because we benchmark against the best decision given knowledge of $P$ and not a realized pair $(X,Y) \sim P$. Both risk and regret depend on $P$ only through its second-moment matrix
$\bSigma_P\defeq\E_P[ZZ^\top]$, with blocks
$\bSigma_X^P$, $\bSigma_Y^P$, and
$\bSigma_{XY}^P=(\bSigma_{YX}^P)^\top$. Write
$\cP^*\defeq\{P\in\cP_2(\R^d):\bSigma_X^P\succ\bZero\}$
for the class of \emph{non-degenerate} laws.
For $P \in \cP^*$, its unique risk minimizer is
$\bA_P\defeq\bSigma_{YX}^P(\bSigma_X^P)^{-1}$ and regret has the closed form
\[
    R(\bA,P)
    =\bigl\|(\bA-\bA_P)(\bSigma_X^P)^{1/2}\bigr\|_\Frob^2.
\]
We write $\|\cdot\|_\Frob$ and $\|\cdot\|_\op$ for the Frobenius norm
and Euclidean operator norm, respectively.

In our population model, an adversary chooses a source--target pair
$P_{\mathrm{s}},P_{\mathrm{t}}\in\cP_2(\R^d)$ satisfying
$\Wtwo(P_{\mathrm{s}},P_{\mathrm{t}})\leq\eps$ and reveals only
the source $P_{\mathrm{s}}$.
Given the radius $\eps > 0$, the learner applies an estimator
$\hat{\bA}:\cP_2(\R^d)\to\cA$ and incurs target regret
$R(\hat{\bA}[P_{\mathrm{s}}],P_{\mathrm{t}})$.
We measure the \emph{robust regret} of $\hat{\bA}$ at a reference law
$P$ by allowing $P$ to occupy either the source or target role:
\[
    R_\eps(\hat{\bA},P)
    \defeq\sup_{Q\in\cB_\eps(P)}
    \max\bigl\{
        R(\hat{\bA}[P],Q),\,
        R(\hat{\bA}[Q],P)
    \bigr\},
\]
where
$\cB_\eps(P)\defeq\{Q\in\cP_2(\R^d):\Wtwo(P,Q)\leq\eps\}$.
The first term varies the target with fixed source $P$, while the second varies
the source with fixed target $P$.

\textbf{Robust estimators.} We primarily compare \emph{empirical risk minimization} (ERM), Wasserstein DRO,
\emph{minimax regret optimization} (MRO), and
\emph{spectral-norm regularization} (SNR):\footnote{\label{footnote:minimizer-existence}
All four optimization problems admit minimizers; see
\cref{app:minimizer-existence}. All of our regret bounds (upper and lower) hold for any minimizer selection rule.}
\begin{align*}
    \ERM[P]
    &\in\argmin_{\bA\in\cA}L(\bA,P),\\
    \DRO_\eps[P]
    &\in\argmin_{\bA\in\cA}
        \sup_{Q\in\cB_\eps(P)}L(\bA,Q),\\
    \MRO_\eps[P]
    &\in\argmin_{\bA\in\cA}
        \sup_{Q\in\cB_\eps(P)}R(\bA,Q),\\
    \SNR_\eps[P]
    &\in\argmin_{\bA\in\cA}
        L(\bA,P)+\eps^2\|\bA\|_\op^2.
\end{align*}
Here, ERM fits the source law directly, returning $\bA_P$ when $P$ is
non-degenerate. MRO replaces the risk objective in DRO with regret.
Finally, SNR adds a squared operator-norm penalty to ERM.

\textbf{Local guarantees.}
To quantify estimation complexity, we first study the
local minimax benchmark
\begin{equation}
\label{eq:local-minimax}
    R_\eps^\star(P)
    \defeq\inf_{\hat{\bA}}R_\eps(\hat{\bA},P)
    =\inf_{\bA\in\cA}
      \sup_{Q\in\cB_\eps(P)}R(\bA,Q).
\end{equation}
Equality follows by considering $P$ in the source role for the ``$\geq$'' direction and considering constant estimators for the ``$\leq$'' direction.
For non-degenerate $P$, let
$\bSigma_E^P$ denote the second-moment matrix of the residual
$Y - \bA_P X$. Our first result shows that the following limit exists and computes it exactly:
\[
    \kappa_P
    \defeq\lim_{\eps\downarrow0}\eps^{-2}R_\eps^\star(P)
    =1+\lambda_{\max}\!\left(
        \bA_P\bA_P^\top
        +\frac{\bSigma_E^P}{\lambda_{\min}(\bSigma_X^P)}
    \right).
\]
Thus, the \emph{condition number} $\kappa_P$ governs the optimal local
regret, $\kappa_P\eps^2+o(\eps^2)$.
It reflects the size of the regression coefficients and that of the residual
second moment relative to the smallest feature eigenvalue.
All four estimators attain this local
rate up to absolute multiplicative constants as $\eps\downarrow0$.

\textbf{Global guarantees.}
For fixed $\eps > 0$, we tightly characterize estimator performance as a function of $\eps$ and the condition number. For a class $\cG \subseteq \cP_2(\R^d)$, define the \emph{worst-case} and \emph{minimax} regrets
$R_\eps(\hat{\bA},\cG)\defeq\sup_{P\in\cG}R_\eps(\hat{\bA},P)$ and $R_\eps^\star(\cG)\defeq\inf_{\hat{\bA}}R_\eps(\hat{\bA},\cG)$, so that $R_\eps^\star(P) = R_\eps^\star(\{P\})$. For the well-conditioned family $\cG(\kappa) \defeq \{P \in \cP^*:\kappa_P\leq\kappa\}$, our main findings are summarized in \cref{tab:main-rates}. Namely, the minimax rate is $O(\kappa \eps^2)$ for all $\eps > 0$, and this is attained by MRO and SNR. In contrast, ERM can suffer arbitrarily large regret even for well-conditioned instances. DRO lies between these extremes; when $m = O(1)$ or, more generally, when the residual second moment $\bSigma_E^P$ has low effective rank, we prove that it achieves the minimax rate. For $\kappa \geq 2$, however, its worst-case robust regret is suboptimal by a multiplicative factor of $\sqrt{\min\{p,m\}} + \sqrt{m/\kappa}$.

\begin{table}[t]
    \centering
    \small
    \setlength{\tabcolsep}{3pt}
    \renewcommand{\arraystretch}{1.2}
    \begin{tabular*}{\linewidth}{@{\extracolsep{\fill}}lcccc@{}}
        \toprule
        Estimator $\hat{\bA}$ & $\ERM$ & $\DRO_\eps$ & $\MRO_\eps$ & $\SNR_\eps$ \\
        \midrule
        Robust regret
        & $\infty$
        & $\bigl(\sqrt{\min\{p,m\}}\,\kappa+\sqrt{m\kappa}\bigr)\eps^2$
        & $\kappa\eps^2$
        & $\kappa\eps^2$ \\[3pt]
        Minimax optimal?
        & {\color{red!70!black}$\times$} & {\color{red!70!black}$\times$} & {\color{green!60!black}$\checkmark$} & {\color{green!60!black}$\checkmark$} \\[3pt]
        Computation
        & \begin{tabular}[c]{@{}c@{}}Closed\\form\end{tabular}
        & \begin{tabular}[c]{@{}c@{}}Convex\\reformulation\end{tabular}
        & \begin{tabular}[c]{@{}c@{}}(New) convex\\reformulation\end{tabular}
        & \begin{tabular}[c]{@{}c@{}}Accelerated proximal\\gradient method (FISTA)\end{tabular} \\
        \bottomrule
    \end{tabular*}
    \vspace{0.3mm}

    \caption{Worst-case robust regret
    $R_\eps(\hat{\bA},\cG(\kappa))$ for our running estimators, up to absolute multiplicative constants, for $\eps > 0$ and $\kappa \geq 2$. Computation uses second-moment matrices.}
    \label{tab:main-rates}
\end{table}

For upper bounds, our analysis employs the Bures--$\Wtwo$ geometry of second-moment matrices to prove a key regret stability lemma. In particular, we show that
$R(\bA_P,Q)\lesssim\kappa_P\Wtwo(P,Q)^2$ for every non-degenerate
$P$ and arbitrary $Q\in\cP_2(\R^d)$.
More generally, our stability lemma controls changes in square-root
regret through the predictor's operator norm, explaining the role of
the SNR penalty. For lower bounds, we provide explicit constructions where these upper bounds are tight.

\textbf{Computation and finite-sample statistics.}
We derive a novel, exact convex reformulation of MRO, building upon previous results for DRO \citep{shafieezadeh2018wasserstein}, along with an efficient proximal
algorithm for SNR, based on FISTA \citep{beck2009fista}.
Given a fixed source second-moment matrix and radius $\eps>0$, our SNR algorithm reaches objective gap $\delta>0$ in $O(\delta^{-1/2})$ iterations,
each using a singular value decomposition.
We also extend our results to two finite-sample models:
adversarial corruption of independent and identically
distributed (i.i.d.) target observations, and i.i.d.\ source
observations with a Wasserstein-perturbed population target. Here, we prove high-probability regret bounds under fourth-moment assumptions, with improved dependence on the confidence level under sub-Gaussian tails.
Finally, we provide synthetic numerical experiments illustrating the predicted separation of MRO and SNR from DRO.

\subsection{Related Work}

Wasserstein DRO (WDRO) for regression and other quadratic models has been studied extensively through the lens of duality, regularization, and robust risk minimization; see, e.g., \cite{nguyen2018distributionally,nguyen2021mean,taskesen2023distributionally} in addition to the works cited above. \cite{shafieezadeh2018wasserstein} explicitly develop WDRO algorithms for our multivariate linear regression setting, applied to a sequential Kalman filtering problem. More broadly, WDRO has gained interest in ML from the perspective of adversarial training \citep{sinha2017certifying}, and Wasserstein-robust risk has been well explored by several communities \citep{lee2018minimax,blanchet2019quantifying}.

Our focus is different: we evaluate estimators by \emph{excess} risk at the unknown target law, rather than by risk at the observed source or worst-case law. This distinction is essential, since regret-optimal procedures need not coincide with DRO. MRO, also called distributionally robust regret optimization, has garnered significant study. \cite{agarwal2022minimax} compare MRO and DRO with general ambiguity sets and show that heterogeneity in irreducible risk can cause DRO to incur larger regret. In operations research and control, related formulations have been studied for newsvendor models \citep{perakis2008regret,chen2021regret}, more general Wasserstein ambiguity sets \citep{cho2024wasserstein,fiechtner2025wasserstein}, and multistage stochastic control \citep{fiechtner2026distributionally}. 
However, these works do not establish worst-case robust regret rates for estimator comparison.

\cite{fiechtner2025wasserstein} introduce a first-order sensitivity parameter which nullifies for our setting. Theorem 3 in Version 5 of their paper (posted after the conference submission of this work) establishes a second-order sensitivity expansion whose coefficient in fact specializes to our condition number. We provide a self-contained Bures-geometric derivation, an explicit formula for multivariate least squares, and connect this local quantity to global stability bounds. Their all-radius optimality result for ERM concerns losses jointly quadratic in the decision and uncertainty, excluding our setting.

Our model is closely related to adversarially robust statistics under Wasserstein perturbations. Scalar-response linear regression has been studied under 1-Wasserstein \citep{zhu2022generalized,liu2023robust} and 2-Wasserstein contamination
\citep{chao2023statistical}. These works establish robust guarantees under different distributional and design assumptions; our global comparisons are instead indexed by the local minimax condition number. Moreover, moving to multivariate responses is crucial for us to separate the worst-case regret of DRO from the minimax rates of MRO and SNR.

\subsection{Preliminaries}
\label{ssec:prelims}

\textbf{Notation.} Unsubscripted vector norms are Euclidean. Write $\PSD^d$ for the cone of positive
semidefinite $d \times d$ matrices and $\OO^d \defeq \{ \bU \in \R^{d \times d} : \bU \bU^\top = \bU^\top \bU = \bI_d \}$ for the family of orthogonal matrices.
For symmetric matrices $\bM,\bN$ of the same size, write
$\bM\preceq\bN$ ($\bM\prec\bN$) when $\bN-\bM$ is positive
semidefinite (positive definite), and define $\succeq,\succ$ analogously.
Let $\lambda_{\min}(\bM)$ and $\lambda_{\max}(\bM)$ denote the
smallest and largest eigenvalues of a symmetric matrix $\bM$.
For a real matrix $\bM$, let $\bM^\dagger$ denote its
Moore--Penrose pseudoinverse. If square, write $\tr(\bM)$ for its trace. Let $\bZero$ and $\bI$ denote the zero and identity matrices, respectively, indexed by dimension when needed. The notation $\diag$ constructs a diagonal matrix from scalar
arguments and a block-diagonal matrix from square-matrix arguments. For $\bM\succeq\bZero$, $\bM^{1/2}$ denotes its unique positive
semidefinite square root. Write $a \lesssim b$ if $a \leq Cb$ for an absolute constant $C > 0$. Write $a \gtrsim b$ if $b \lesssim a$ and $a \asymp b$ if $a \lesssim b$ and $b \lesssim a$.

\textbf{Bures--Wasserstein geometry.} We employ the \emph{2-Wasserstein distance} defined for $P,Q \!\in\! \cP_2(\R^d)$ by
\begin{equation*}
    \Wtwo(P,Q) \defeq \inf_{\pi \in \Pi(P,Q)} \E_{(Z,Z') \sim \pi}\bigl[\|Z - Z'\|^2\bigr]^{\frac{1}{2}},
\end{equation*}
where $\Pi(P,Q)$ denotes the set of joint probability laws on $\R^d \times \R^d$ with marginals $P$ and $Q$. Since our least-squares risk only depends on (uncentered) second moments, it is convenient to switch from the full $\Wtwo$ geometry on probability measures to its induced geometry on second-moment matrices. In particular, we turn to the \emph{Bures distance}, defined between positive semidefinite $\bA,\bB \succeq \bZero$ by
\begin{equation*}
    \sB(\bA,\bB)^2 \defeq \tr(\bA) + \tr(\bB) - 2\tr\bigl((\bA^{\frac{1}{2}}\bB\bA^{\frac{1}{2}})^{\frac{1}{2}}\bigr).
\end{equation*}
We refer the reader to \cite{bhatia2019bures} for a recent survey of this metric and recall several of its important properties below, deferring proofs to \cref{app:Bures-geometry}. We often use an equivalent formulation of the Bures distance as the minimum Frobenius norm between square-root factors of $\bA$ and $\bB$.

\begin{lemma}[Procrustes formulation of $\sB$]
\label{lem:Bures-Procrustes}
For all $\bA,\bB \in \PSD^d$, we have
\begin{equation*}
    \sB(\bA,\bB) = \min_{\bU \in \OO^d} \|\bA^{\frac{1}{2}} - \bB^{\frac{1}{2}}\bU\|_\Frob = \min_{\substack{\bS,\bT \in \R^{d \times d}\\ \bA = \bS \bS^\top\!\!, \,\bB = \bT \bT^\top}} \|\bS - \bT\|_\Frob.
\end{equation*}
Thus, $\sB(\bA,\bB) \leq \eps$ if and only if there exists $\bDelta \in \R^{d \times d}$ with $\|\bDelta\|_\Frob \leq \eps$ such that $\bA = (\bB^{1/2} + \bDelta)(\bB^{1/2} + \bDelta)^\top$. The minimum is attained by $\bU = \bI_d$ if $\bA$ and $\bB$ are simultaneously diagonalizable.
\end{lemma}

More importantly, this metric is precisely that induced by $\Wtwo$ on the space of second-moment matrices. Namely, setting $\cB_\eps(\bSigma) \defeq \{ \bSigma' \in \PSD^d : \sB(\bSigma',\bSigma) \leq \eps \}$, we have the following.

\begin{lemma}[$\Wtwo$--Bures equivalence]
\label{lem:W2-Bures-equivalence}
For all $P \in \cP_2(\R^d)$ and $\eps \geq 0$, we have
\begin{equation*}
   \{ \bSigma_Q : Q \in \cB_\eps(P) \} = \cB_\eps(\bSigma_P) = \left\{ (\bSigma_P^{1/2} + \bDelta)(\bSigma_P^{1/2} + \bDelta)^\top : \|\bDelta\|_\Frob \leq \eps \right\}.
\end{equation*}
\end{lemma}

Throughout, we refer to $\PSD^d$ equipped with $\sB$ as \emph{Bures space}.

\section{Local Asymptotic Complexity and the Condition Number}
\label{sec:condition-number}

We begin by exploring the fundamental limits of robust estimation in the asymptotic $\eps \to 0$ regime. Recall the local minimax robust regret $R_\eps^\star(P)$, defined in \eqref{eq:local-minimax}.
In general, this quantity depends on the complex interaction between $P$, $R$, and the $\Wtwo$ geometry. However, as $\eps \to 0$, we show that the complexity of robust prediction is dictated by the following limiting quantity, which we compute in closed form under a mild non-degeneracy assumption.

\begin{definition}
\label{def:condition-number}
Define the \emph{condition number} of $P \in \cP^*$ by $\kappa_P \defeq \limsup_{\eps \downarrow 0} \eps^{-2} R^\star_\eps(P)$. For convenience, we set $\kappa_P \defeq \infty$ for $P \notin \cP^*$.\footnote{This definition and our resulting non-asymptotic analysis can be adapted to the setting where $\bSigma^P_X$ is singular but the condition number limit is finite (which occurs when the optimal risk is 0). We omit this edge case for ease of presentation.}
\end{definition}

\begin{theorem}[Condition number formula]
\label{thm:condition-number-formula}
If $P \in \cP^*$, then $R^\star_\eps(P) = \kappa_P \eps^2 + o(\eps^2)$ as $\eps \downarrow 0$ and
\begin{equation*}
    \kappa_P = 1 + \lambda_{\max}\bigl(\bA_P \bA_P^\top + \lambda_{\min}(\bSigma_X^P)^{-1}\bSigma_E^P\,\bigr).
\end{equation*}
This implies the useful bounds
\begin{equation*}
     1 + \max\left\{\|\bA_P\|_\op^2, \frac{\lambda_{\max}(\bSigma_E^P)}{\lambda_{\min}(\bSigma_X^P)}\right\}  \leq \kappa_P \leq  1 + \|\bA_P\|_\op^2 + \frac{\lambda_{\max}(\bSigma_E^P)}{\lambda_{\min}(\bSigma_X^P)}.
\end{equation*}
\end{theorem}

We derive this formula in Appendix~\ref{prf:condition-number-formula} by separately considering perturbations to the feature coordinates $X$ and the response coordinates $Y$. The additive $1$ term arises due to response perturbations which leave $X$ fixed, while the remaining eigenvalue term results from feature perturbations which leave $Y$ fixed. This latter term is determined by both the size of the optimal coefficient matrix and the scale of the residual variation compared to that of the features. 

The proof equates $\kappa_P$ to a second-order sensitivity of $R(\bA_Q,P)$ as $Q$ varies in the neighborhood of $P$. In Bures space, this amounts to a squared operator norm of a suitable Euclidean derivative, which we compute with some calculus and a scalarization trick that simplifies tensor calculations.

\section{Non-Asymptotic Regret Bounds}
\label{sec:non-asymptotic}

Although $\kappa$ is defined by an $\eps \to 0$ limit, we seek regret bounds which hold for all $\eps > 0$. To derive these, we prove a key lemma allowing one to transfer regret bounds between nearby distributions.

\begin{lemma}[Regret stability]
\label{lem:regret-stability}
For all $\bA \in \cA$, $P \in \cP^*$, and $Q \in \cP_2(\R^d)$, we have
\begin{align*}
    \left|\sqrt{R(\bA,P)} - \sqrt{R(\bA,Q)}\,\right| \leq (2\sqrt{\kappa_P} + \|\bA\|_\op)\Wtwo(P,Q).
\end{align*}
Moreover, for $\bA = \bA_P$, we have $R(\bA_P,Q) \leq 4 \kappa_P \Wtwo(P,Q)^2$.
\end{lemma}

The proof in Appendix~\ref{prf:regret-stability} performs some finite-difference error analysis using a closed form for the regret in terms of Euclidean projections, valid even for degenerate $Q$. Here, (a variant of) the Davis--Kahan--Wedin sine theorem \citep{wedin1972perturbation} serves as a key technical tool, controlling the Frobenius discrepancy between projection matrices onto the row spaces of nearby matrices.

Regret stability implies several important regret bounds for our main estimators.

\begin{lemma}[Stability implications]
\label{lem:stability-implications}
Let $P \in \cP^*$ and $\eps > 0$. We then have
\begin{align*}
    R_\eps(\MRO_\eps,P) \lesssim \kappa_P \eps^2, \qquad R_\eps(\SNR_\eps,P) \lesssim \kappa_P \eps^2.
\end{align*}
In contrast, our ERM and DRO guarantees adapt to the largest condition number in a ball around $P$:
\begin{equation*}
    R_\eps(\ERM,P) \lesssim \sup_{Q \in \cB_\eps(P)} \kappa_Q \eps^2 \qquad R_\eps(\DRO_\eps,P) \lesssim \sup_{Q \in \cB_{2\eps}(P)} \kappa_Q \eps^2
\end{equation*}
If $\eps^2 \leq \lambda_{\min}(\bSigma_X^P)/16$, then these two upper bounds are at most $O(\kappa_P \eps^2)$.
\end{lemma}

We give the proof for MRO and SNR below, deferring the ERM and DRO arguments to Appendix~\ref{prf:stability-implications-ERM-DRO}.

\begin{proof}[Proof of \cref{lem:stability-implications} (MRO and SNR)]
Let $Q\in\cB_\eps(P)$ and fix source--target pair
$(P_{\mathrm{s}},P_{\mathrm{t}})$ equal to either $(P,Q)$ or $(Q,P)$. For MRO, we set $\hat{\bA}\defeq\MRO_\eps[P_{\mathrm{s}}]$ and bound
\begin{align*}
    R(\hat{\bA},P_{\mathrm{t}})
    &\leq
    \sup_{Q'\in\cB_\eps(P_{\mathrm{s}})}R(\hat{\bA},Q') \leq
    \sup_{Q'\in\cB_\eps(P_{\mathrm{s}})}R(\bA_P,Q')
    \leq 16\kappa_P\eps^2.
\end{align*}
The first inequality uses
$P_{\mathrm{t}}\in\cB_\eps(P_{\mathrm{s}})$, the second uses
MRO optimality, and the last uses
$\cB_\eps(P_{\mathrm{s}})\subseteq\cB_{2\eps}(P)$ and
\cref{lem:regret-stability}. Supremizing over $Q$ and source--target pairs gives $R_\eps(\MRO_\eps,P) \lesssim \kappa_P \eps^2$.

For SNR, take instead $\hat{\bA}\defeq\SNR_\eps[P_{\mathrm{s}}]$.
Since risk and regret differ by a constant independent of $\bA$,
\begin{align*}
    R(\hat{\bA},P_{\mathrm{s}})
    +\eps^2\|\hat{\bA}\|_\op^2 \leq
    R(\bA_P,P_{\mathrm{s}})
    +\eps^2\|\bA_P\|_\op^2 \leq 5\kappa_P\eps^2,
\end{align*}
where the first inequality uses SNR optimality and the second follows from
$P_{\mathrm{s}}\in\cB_\eps(P)$,
\cref{lem:regret-stability}, and
$\|\bA_P\|_\op^2\leq\kappa_P$ (by \cref{thm:condition-number-formula}).
Applying \cref{lem:regret-stability} again yields
\begin{align*}
    \sqrt{R(\hat{\bA},P_{\mathrm{t}})}
    &\leq
    \sqrt{R(\hat{\bA},P_{\mathrm{s}})}
    +\bigl(\|\hat{\bA}\|_\op+2\sqrt{\kappa_P}\bigr)\eps\\
    & \leq
    \sqrt{2}\sqrt{
        R(\hat{\bA},P_{\mathrm{s}})
        +\eps^2\|\hat{\bA}\|_\op^2
    }
    +2\sqrt{\kappa_P}\eps \leq 6 \sqrt{\kappa_P}\eps.
\end{align*}
The second inequality is Cauchy--Schwarz, and the last uses the
objective comparison above. Squaring this SNR bound and supremizing over $Q$ and source--target pairs gives $R_\eps(\SNR_\eps,P) \lesssim \kappa_P \eps^2$.
\end{proof}

By combining the DRO guarantee above with existing DRO regularizer bounds \citep{gao2024wasserstein,lee2018minimax}, we obtain an all-radius guarantee
adapted to the residual effective rank. Writing
$\reff(\bM)\defeq\tr(\bM)/\|\bM\|_\op$ for $\bM \succeq \bZero$, with
$\reff(\bZero)\defeq0$, we prove the following in Appendix~\ref{prf:DRO-rank-adaptive}.

\begin{lemma}[Rank-adaptive DRO bound]
\label{lem:DRO-rank-adaptive}
For $P\in\cP^*$ and $\eps > 0$, set
$r\defeq\reff(\bSigma_E^P)$. Then
\[
    R_\eps(\DRO_\eps,P)
    \lesssim
    \left(\sqrt{\min\{p,r\}}(\kappa_P-1) + \sqrt{r(\kappa_P-1)} + \kappa_P\right) \eps^2.
\]
\end{lemma}

In particular, DRO achieves regret $O(\kappa_P\eps^2)$ whenever the residual second moment has effective rank $O(1)$. In the worst case, however, the effective rank can be as large as $m$.

Combining the above, we derive worst-case rates for the family $\cG(\kappa) = \{ P \in \cP^*: \kappa_P \leq \kappa \}$.

\begin{theorem}[Population-limit summary rates]
\label{thm:pop-limit-risk-bds}
Fix $\eps > 0$ and $\kappa \geq 1$. Then the minimax rate for the class $\cG = \cG(\kappa)$ is $R_\eps^\star(\cG) \asymp \kappa \eps^2$, and this is achieved by both $\MRO_\eps$ and $\SNR_\eps$. In contrast, we have $R_\eps(\ERM,\cG) = \infty$ and
\begin{equation*}
    R_\eps(\DRO_\eps,\cG) \asymp \left(\sqrt{\min\{p,m\}}(\kappa-1) + \sqrt{m(\kappa-1)} + \kappa\right) \eps^2.
\end{equation*}
In particular, if $\kappa \geq 2$ and $p \geq m$, then $R_\eps(\DRO_\eps,\cG) \asymp \sqrt{m} \kappa \eps^2$.
\end{theorem}

In summary, MRO and SNR are always minimax optimal, ERM can suffer unbounded regret, and DRO is provably suboptimal by a dimension-dependent factor for $\kappa \geq 2$. The upper bounds in Appendix~\ref{prf:pop-limit-risk-bds} apply \cref{lem:stability-implications,lem:DRO-rank-adaptive}. The minimax lower bound follows by rescaling the
local result from \cref{thm:condition-number-formula}; this works because the condition number is invariant to uniform rescaling.

For ERM, we consider a target law $P$ with identically zero responses. A nearby source law with a nearly collapsed feature and
a response correlated with that feature forces the ERM solution to diverge, yielding unbounded target regret even though $\kappa_P=1$.
The DRO lower bound requires more care. We employ two
constructions with $\lambda_{\min}(\bSigma_X^P)=\eps^2$ and
independent residual noise spread across many response coordinates.
In each, the adversary contracts one feature coordinate while
leaving the responses unchanged. We then exploit rotational symmetry and a strong duality formula to characterize the DRO minimizers explicitly.
The noisy response coordinates determine the minimum value, and we tune the contraction so that attaining this value forces an
inflated coefficient on the contracted feature and hence large
target regret. This demonstrates how independent
residual noise, which contributes an additive constant to target
risk and thus cancels from regret, can nevertheless lead to
suboptimal DRO decisions.

\section{Efficient Computation}
\label{sec:computation}

We next treat computation, given an observed second-moment matrix $\tilde{\bSigma} \in \PSD^d$ and radius $\eps > 0$. Since risk and regret only depend on laws via their second
moments, we define $L(\bA,\bSigma) \defeq L(\bA,P)$ for any law $P$ such that $\bSigma_P = \bSigma$, and likewise for $R(\bA,\bSigma)$. By the Bures--$\Wtwo$ equivalence (\cref{lem:W2-Bures-equivalence}), the optimization problems
defining our estimators admit equivalent formulations over second-moment matrices, used below. We defer sampling considerations to
\cref{sec:finite-sample}.

ERM admits the canonical minimum-Frobenius-norm solution $\smash{\tilde{\bSigma}}_{YX}\smash{\tilde{\bSigma}}_X^{\dagger}$, which can be computed via eigendecomposition of $\smash{\tilde{\bSigma}_X}$. Alternatively, one may apply any standard convex solver to the objective $\bA \mapsto L(\bA,\tilde{\bSigma}) = \tr(\bA^\top \bA \smash{\tilde{\bSigma}}_X) - 2\tr(\bA \smash{\tilde{\bSigma}}_{XY}) + \tr(\smash{\tilde{\bSigma}}_{Y})$.

\cite{shafieezadeh2018wasserstein} developed a convex reformulation for the DRO problem, accompanied by a Frank--Wolfe method with convergence guarantees (see Theorem 3.3 therein). The convex reformulation na\"ively requires solving a large semidefinite program, so their Frank--Wolfe method is a noted improvement, albeit with some dependencies on $\lambda_{\min}(\smash{\tilde{\bSigma}})^{-1}$ in their convergence rate. We do not seek any further improvements, given the stronger performance of MRO and SNR.

The MRO objective $\bA \mapsto \sup_{\bSigma \in \cB_\eps(\tilde{\bSigma})} R(\bA,\bSigma)$ is convex as the pointwise supremum of convex functions in $\bA$. However, regret is convex in $\bSigma$, as the pointwise supremum over $\bA'$ of $L(\bA,\bSigma) - L(\bA',\bSigma)$ which is linear in $\bSigma$. This appears problematic, since the inner problem requires maximizing over $\bSigma$. Nevertheless, we show that it admits a convex reformulation. Here $\tr_p(\bH)$ denotes the sum of the $p$ largest eigenvalues of a symmetric matrix $\bH$, counted with multiplicity.

\begin{theorem}[Convex reformulation for MRO]
\label{thm:MRO-convex-reformulation}
Let $\eps > 0$, $\tilde{\bSigma} \in \PSD^d$, and $\tilde{\bS} \in \R^{d \times d}$ such that $\tilde{\bSigma} = \tilde{\bS} \tilde{\bS}^\top$. Decompose
\begin{equation*}
    \tilde{\bS} =
    \begin{pmatrix}
        \tilde{\bS}_X\\
        \tilde{\bS}_Y
    \end{pmatrix},
    \qquad
    \tilde{\bS}_X \in \R^{p\times d},
    \qquad
    \tilde{\bS}_Y \in \R^{m\times d}.
\end{equation*}
For $\bA \in \cA$, consider the MRO objective $F(\bA) \defeq \sup_{\bSigma\in\cB_\eps(\tilde{\bSigma})}R(\bA,\bSigma)$ and the residual factor $\tilde{\bE}_{\bA} \defeq \tilde{\bS}_Y - \bA\tilde{\bS}_X \in \R^{m \times d}$. Define the extended-real-valued function $D:\cA \times \R \to \R \cup \{+\infty\}$ by
\begin{equation*}
    D(\bA,\eta) \defeq \eta\left(
        \eps^2-\|\tilde{\bS}_X\|_\Frob^2
        +
        \tr_p\left(
            \tilde{\bS}_X^\top\tilde{\bS}_X
            +
            \tilde{\bE}_{\bA}^\top
            \bigl((\eta-1)\bI_m-\bA\bA^\top\bigr)^{-1}
            \tilde{\bE}_{\bA}
        \right)
    \right)
\end{equation*}
if $\eta > 1 + \|\bA\|_\op^2$ and $+\infty$ otherwise.
Then $D$ is convex as an extended-value function on
\(\cA\times\R\) and $F(\bA) = \inf_{\eta \in \R}D(\bA,\eta)$. Consequently, MRO admits the lifted convex reformulation
\[
    \inf_{\bA\in\cA}F(\bA)
    = \inf_{\bA \in \cA} \sup_{\bSigma \in \cB_\eps(\tilde{\bSigma})} R(\bA,\bSigma) = 
    \inf_{\bA\in\cA,\ \eta\in\R}D(\bA,\eta).
\]
\end{theorem}

This proof in \cref{prf:MRO-convex-reformulation} extends a known dual formulation for the inner DRO problem
\begin{equation}
\label{eq:DRO-inner-strong-duality}
    \sup_{\bSigma \in \cB_\eps(\tilde{\bSigma})} L(\bA,\bSigma) =  \inf_{\eta > 1 + \|\bA\|_\op^2} \eta \left(\eps^2 + \tr\left(\tilde{\bE}_{\bA}^\top\bigl((\eta-1) \bI_m - \bA \bA^\top\bigr)^{-1} \tilde{\bE}_{\bA} \right)\right),
\end{equation}
which underlies the Frank--Wolfe implementation in \cite{shafieezadeh2018wasserstein}. In both cases, the variable $\eta$ is introduced as a Lagrange multiplier for the Bures ball constraint. However, our route to the MRO formula involves an additional maximization problem over rank-$p$ projection matrices. Initially outside the infimum, we employ a non-convex minimax theorem to bring the supremum inside, with the $\tr_p$ term arising due to the Ky Fan variational principle.

\cref{thm:MRO-convex-reformulation} suggests solving the MRO problem by applying a subgradient method to minimize $D$ over $\bA,\eta$. Indeed, subgradients can be computed via eigendecomposition, due to the Ky Fan variational formula for $\tr_p$ and Danskin's theorem. Moreover, if $\hat{\bA},\hat{\eta}$ are a feasible, $\delta$-approximate minimizing pair, then $\smash{\hat{\bA}}$ is a $\delta$-approximate MRO solution. We defer a complete convergence analysis to future work, since the minimax rate of MRO is shared by the much simpler SNR, explored next.

In contrast, the SNR objective $G(\bA) \!\defeq\! L(\bA,\tilde{\bSigma}) + \eps^2\|\bA\|_\op^2$ is convex and enjoys simpler subgradient computations. Indeed, for unit singular vectors $u \!\in\! \R^m$, $v \!\in\! \R^p$ satisfying $\bA v = \|\bA\|_\op u$, we have\footnote{The first two terms for this subgradient representative are obtained directly by differentiating $L(\cdot,\tilde{\bSigma})$. The last term follows, for example, from the characterization of the spectral-norm subdifferential set in Example 2.8 of \citet{lewis1995convex}.}
\begin{equation*}
    2\bA \tilde{\bSigma}_X - 2 \tilde{\bSigma}_{YX} + 2 \eps^2 \|\bA\|_\op uv^\top \in \partial G(\bA).
\end{equation*}

Some simple bounds on the optimizer norm yield the following for a projected subgradient method.

\begin{proposition}[SNR subgradient convergence]
\label{prop:SNR-subgradient-convergence}
Let $\tilde{\bSigma} \in \PSD^d$ and $\eps > 0$. If $\tilde{\bSigma}_Y = \bZero$, then $\bZero$ minimizes $G$. Otherwise, define radius $B \defeq (d \tr \tilde{\bSigma}_Y)^{1/2}/\eps$, Lipschitz bound $M \defeq 2(B \|\tilde{\bSigma}_X\|_\op +  \|\tilde{\bSigma}_{YX}\|_\Frob +  \eps^2 B)$, and iteration count $T \defeq \lceil M^2 B^2 \delta^{-2} \rceil$, for accuracy $\delta > 0$. Consider the update $\bA_{t+1} \!\gets\! \mathrm{proj}_B(\bA_t \!-\! \lambda \bG_t)$, where $\bA_1 = \bZero$, $\bG_t \in \partial G(\bA_t)$, $\lambda \defeq B/(M\sqrt{T})$, and $\mathrm{proj}_B$ is the Euclidean projection onto the Frobenius ball of radius $B$ centered at $\bZero$. Then, the averaged iterate $\bar{\bA}_T = T^{-1} \sum_{t=1}^{T} \bA_t$ achieves $G(\bar{\bA}_T) \leq \min_\bA G(\bA) + \delta$.
\end{proposition}

The proof in \cref{prf:SNR-subgradient-convergence} is routine. Hence, we can solve SNR to accuracy $\delta$ with $O(\delta^{-2})$ iterations, each computing a top singular vector. To accelerate this rate, we use that our nonsmooth penalty admits an efficiently computable proximal operator, enabling the application of FISTA \citep{beck2009fista}. In \cref{prf:SNR-FISTA-convergence}, we prove an improved rate for the resulting \cref{alg:SNR-FISTA}.\footnote{Faster linear rates are available if $L(\cdot,\tilde{\bSigma})$ is strongly convex, but this requires $\lambda_{\min}(\tilde{\bSigma}_X)$ to be bounded away from $0$, which is not guaranteed.} 

\begin{proposition}[SNR FISTA convergence]
\label{prop:SNR-FISTA-convergence}
Fix $\tilde{\bSigma} \in \PSD^d$, $\eps > 0$, and $\delta > 0$.
Then \cref{alg:SNR-FISTA} achieves $G(\bA_T) \leq \min_\bA G(\bA) + \delta$ after $T = \lceil (4 d\lambda_{\max}(\tilde{\bSigma}_X) \tr(\tilde{\bSigma}_Y))^{1/2} \eps^{-1} \cdot \delta^{-1/2} \rceil$ iterations.
\end{proposition}

In particular, FISTA achieves accuracy $\delta$ using $O(\delta^{-1/2})$ full singular value decompositions.

\algtext*{EndFor}
\algtext*{EndIf}
\begin{algorithm}[t]
\caption{FISTA for SNR}
\label{alg:SNR-FISTA}
\begin{algorithmic}[1]
\Require Observed second moments $\tilde{\bSigma}_X \in \R^{p \times p}$, $\tilde{\bSigma}_{YX} \in \mathbb R^{m \times p}$, penalty $\eps > 0$, iteration count $T$
\State Compute $M=2\lambda_{\max}(\tilde{\bSigma}_X)$
\If{$M = 0$} \Return $\bA_T = \bZero$
\EndIf
\State Set $\alpha=\eps^2/M$, initialize $\bA_0=\bZero_{m\times p}$, $\bY_0=\bA_0$, $t_0=1$
\For{$k=0,1,\dots,T-1$}
    \State $\bB_k = \bY_k - \frac{2}{M} \bigl(\bY_k\tilde{\bSigma}_X-\tilde{\bSigma}_{YX}\bigr)$ \Comment{gradient step for the smooth loss}
    \State $\bA_{k+1} = \operatorname{\texttt{ProxSpecSq}}(\bB_k,\alpha)$ \Comment{proximal step}
    \State $t_{k+1} = (1+\sqrt{1+4t_k^2}\,)/2$
    \State $\bY_{k+1} = \bA_{k+1} + \frac{t_k-1}{t_{k+1}} \left(\bA_{k+1}-\bA_k\right)
    $ \Comment{Nesterov update}
\EndFor
\State \Return $\bA_T$
\end{algorithmic}
\end{algorithm}

\begin{algorithm}[t]
\caption{\texttt{ProxSpecSq}: proximal map of $\alpha\|\cdot\|_{\op}^2$}
\label{alg:prox-spec-sq}
\begin{algorithmic}[1]
\Require Matrix $\bB \in \R^{m \times p}$, parameter $\alpha > 0$
\State Compute thin SVD $\bB = \bP\diag(\sigma_1,\dots,\sigma_q)\bQ^\top$, $\sigma_1 \geq \cdots \geq \sigma_q \geq 0$, $q=\min\{m,p\}$
\If{$\sigma_1=0$}
    \Return $\bB$
\EndIf
\State Define $\sigma_{q+1}=0$. Find the unique $k_\star\in\{1,\dots,q\}$ such that
\begin{equation*}
    \sigma_{k_\star} > \tau_{k_\star} \geq \sigma_{k_\star+1},
    \qquad
    \tau_k
    = \textstyle\sum_{i=1}^k\sigma_i/(k+2\alpha).
\end{equation*}
\State Set $\tau=\tau_{k_\star}$ and define clipped singular values $s_i = \min\{\sigma_i,\tau\}$ for $i \in \{1,\dots,q\}$
\State \Return $\:\bP\diag(s_1,\dots,s_q)\bQ^\top$
\end{algorithmic}
\end{algorithm}

\section{Finite-Sample Statistics}
\label{sec:finite-sample}

We now turn to the finite-sample setting. Let $Z_1,\ldots,Z_n\overset{\mathrm{i.i.d.}}{\sim}P$ and $P_n\defeq n^{-1}\sum_{i=1}^n\delta_{Z_i}$.
We treat two models, each specifying an observed empirical source measure $P_{\mathrm{s},n}$ and an unknown population target $P_{\mathrm t}$. In the \emph{sample-corruption model}, $P_{\mathrm t} = P$ and $P_{\mathrm s,n} = n^{-1}\sum_{i=1}^n\delta_{\tilde Z_i}$ for an adversarially perturbed sample $\{\tilde{Z}_i\}_{i=1}^n$ only guaranteed to satisfy $n^{-1} \sum_{i=1}^n\|\tilde Z_i-Z_i\|^2\leq\eps^2\rule[-1ex]{0pt}{1ex}$. Naturally, this implies $\Wtwo(P_{\mathrm s,n},P_n) \leq \eps$. In the \emph{distribution-shift model}, the learner instead observes $P_{\mathrm{s},n}=P_n$, and $P_{\mathrm t}=Q$ for a perturbed population law $Q\in\cB_\eps(P)$. Let $\mathbb M_{\eps,n}(P)$ denote the family of joint laws
of $(P_{\mathrm{s},n},P_{\mathrm t})$ arising from either
model, over all admissible, possibly randomized,
sample-corruption rules and all fixed $Q\in\cB_\eps(P)$. Seeking robust, high-probability regret bounds, we thus define
\[
    R_{\eps,n,\delta}(\hat{\bA}_n,P)
    \defeq
    \sup_{\PP\in\mathbb M_{\eps,n}(P)}
    \inf\left\{
        \tau\geq0:
        \PP\left(
            R(\hat{\bA}_n[P_{\mathrm{s},n}],P_{\mathrm t})
            \leq\tau
        \right)\geq1-\delta
    \right\},
\]
where $\delta \in (0,1)$ and $\hat{\bA}_n$ is any (measurable) estimator mapping $P_{\mathrm{s},n}$ to $\smash{\hat{\bA}_n[P_{\mathrm{s},n}]} \in \cA$.

To control empirical convergence of second-moment matrices, we introduce the \emph{Bures variance proxy}, a rescaled fourth-moment bound given by
\begin{equation}
\label{eq:Bures-variance-proxy}
    \sV_{\sB}(P) \defeq \E_P\left[\left\|(ZZ^\top - \bSigma_P)\bSigma_P^{\dagger/2}\right\|_\Frob^2\right] = \E_P\bigl[\|Z\|^2Z^\top\bSigma_P^\dagger Z\bigr]
     -\tr(\bSigma_P).
\end{equation}
Writing $\cP_4(\R^d)$ for the set of laws with finite fourth moments, we prove the following in \cref{prf:Bures-variance-properties}.

\begin{proposition}[$\sV_\sB$ properties]
\label{prop:Bures-variance-properties}
Let $P \in \cP_4(\R^d)$. Then $\sV_\sB(P) < \infty$, and $\sV_\sB(P) = 0$ if and only if $ZZ^\top = \bSigma_P$ for almost all $Z \sim P$. If $\bSigma_P \succ 0$, then $\sV_\sB(P) \leq \E_P[\|Z\|^4]/\lambda_{\min}(\bSigma_P)$. For Gaussian $P = \cN(0,\bSigma_P)$, we have $\sV_\sB(P) = (\rank(\bSigma_P) + 1)\tr(\bSigma_P)$, and if further $P \in \cP^*$ is non-degenerate, then $\sV_\sB(P) \geq p\tr(\bSigma_E^P)$. Crucially, we have $\E[\sB(\bSigma_P,\bSigma_{P_n})^2] \leq \sV_\sB(P)/n$.
\end{proposition}

Hence, we can extend population-limit results to the finite-sample setting by treating sampling as an adversarial Bures perturbation and slightly inflating $\eps$. However, this approach uses Markov's inequality, leading to sample complexity with inverse polynomial dependence on the failure probability. Alternatively, sub-Gaussian concentration controls the empirical
condition number directly, avoiding
radius inflation. For $K > 0$, recall that a random vector $W$ on $\R^d$ is $K$-sub-Gaussian if $\E[\exp((u^\top W)^2/K^2)] \leq 2$ for each unit vector $u \in \R^d$. We prove the following in \cref{prf:finite-sample-regret-bds}.

\begin{theorem}[Finite-sample regret bounds]
\label{thm:finite-sample-regret-bds}
Let $P\in\cP^*$, $\eps>0$, and $\delta\in(0,1)$.
Suppose first that $P\in\cP_4(\R^d)$, and fix
$V\geq\sV_\sB(P)$. Then, for sample size $n\geq V\delta^{-1}\eps^{-2}$ and expanded radius $\eps_n\defeq\eps+\sqrt{V/(n\delta)}$, we have
\[
        R_{\eps,n,\delta}(\MRO_{\eps_n},P),
        R_{\eps,n,\delta}(\SNR_{\eps_n},P) \lesssim\kappa_P\eps^2.
\]

Alternatively, let $(X,Y)\sim P$, write $E=Y-\bA_PX$,
and suppose that the whitened random vector $((\bSigma_X^P)^{-1/2}X,(\bSigma_E^P)^{\dagger/2}E)$ is $K$-sub-Gaussian. Then, for $n \gtrsim K^4 (d+ \log \frac{1}{\delta})(1 + \tr(\bSigma_E^P)\eps^{-2})$,
\[
R_{\eps,n,\delta}(\MRO_{\eps},P), R_{\eps,n,\delta}(\SNR_{\eps},P) \lesssim \kappa_P \eps^2.
\]
\end{theorem}

Notably, the sub-Gaussian guarantee improves the polynomial dependence on $\delta^{-1}$ of the first
sample-size bound to logarithmic, and it requires
no radius inflation. Still, these sample complexities are not comparable in general, and we defer the question of optimal sample complexity to future work.

\section{Experiments}
\label{sec:experiments}
Finally, we implement our estimators and compare their performance on synthetic finite-sample instances. All methods use the empirical source second-moment matrix
$\bSigma_{P_{\mathrm{s},n}}$. For ERM, we compute the closed-form solution using NumPy. For MRO, we implement the lifted objective $D(\bA,\eta)$ from \cref{thm:MRO-convex-reformulation} in PyTorch, evaluating its $\tr_p$ term by eigendecomposition and computing subgradients by automatic differentiation. We use the same template for DRO, replacing $D(\bA,\eta)$ with the simpler dual objective in \eqref{eq:DRO-inner-strong-duality}. For both methods, we use the normalized subgradient method with $k$th iteration step size $k^{-1/2}$, initialize $\bA$ at the ERM solution and $\eta=2+\|\bA\|_\op^2$, and repair feasibility after each step by taking $\smash{\eta\gets\max\{\eta,1+\|\bA\|_\op^2+\zeta\}}$ for a numerical buffer $\zeta = 0.001$. We return the final iterate after $T = 800$ iterations.  For SNR, we implement the FISTA method in \cref{alg:SNR-FISTA}
using NumPy, with the same ERM initialization and $T = 60$ iterations. The population benchmarks discussed below served as numerical diagnostics for selecting iteration counts and tolerances.

\begin{figure}
    \centering
    \includegraphics[width=1\linewidth]{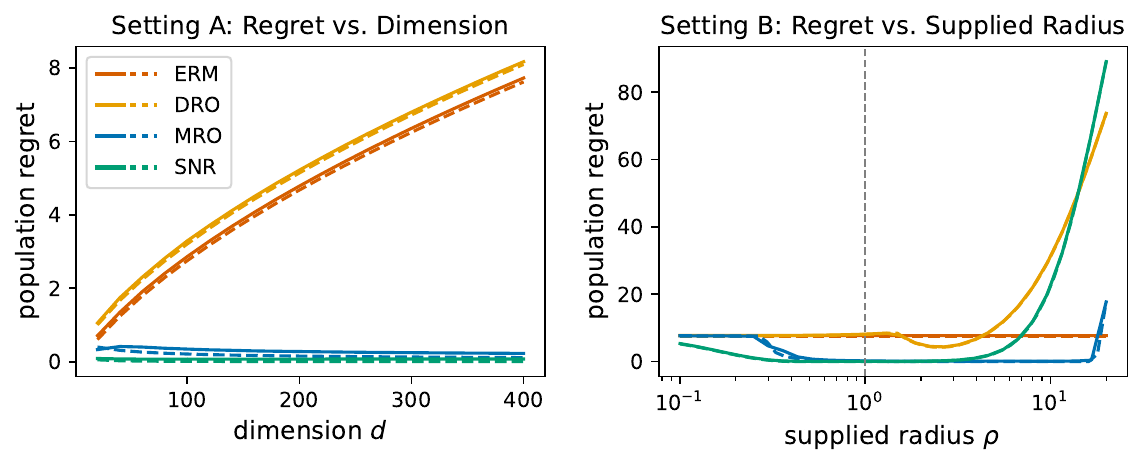}
    \caption{Target population regret $R(\hat{\bA}[P_{\mathrm s, n}],P)$ for Settings A (left) and B (right). Solid curves show finite-sample averages, while dashed curves show population benchmarks. For Setting B, the dashed vertical line marks the corruption budget $\eps=1$.}
    \label{fig:plots}
\end{figure}

In \cref{fig:plots}, we consider two experimental settings based on the \cref{lem:DRO-LB-A} construction. Both use corruption budget $\eps = 1$ and dimensions $p=m=d/2 \geq 2$. We take target measure $P = P_{\mathrm t}$ as the law of $(X,Y)$, where $X\sim \cN(0,\bI_m)$ and $Y=X+U$, and $U \sim \cN(0,\diag(0,\bI_{m-1}))$ is independent of $X$. Then $\bA_P=\bSigma_X^P=\bI_m$,
$\bSigma_E^P=\diag(0,\bI_{m-1})$, and hence $\kappa_P=3$. Target population regret is therefore $R(\bA,P) = \|\bA - \bI_m\|_\Frob^2$. Drawing $\{(X^{(i)},Y^{(i)})\}_{i=1}^n$ i.i.d.\ from $P$, we replace each $X^{(i)}_1$
by $m^{-1/4} X^{(i)}_1$, stopping if needed before the first replacement that would exceed a total squared perturbation budget of $n\eps^2$. All other feature coordinates, responses, and unselected
observations remain unchanged.
Writing $\tilde X^{(i)}$ for the resulting features, the observed empirical source $P_{\mathrm{s},n}
=n^{-1}\sum_{i=1}^n\delta_{(\tilde X^{(i)},Y^{(i)})}$
satisfies $\Wtwo(P_{\mathrm{s},n},P_n)^2 \leq \frac1n\sum_{i=1}^n \|\tilde X^{(i)}-X^{(i)}\|^2 \leq\eps^2$, obeying our \cref{sec:finite-sample} sample-corruption model. We use sample size $n=10m^2$ to match, for fixed confidence, the
$d(1+\tr(\bSigma_E^P))\asymp m^2$ dimensional dependence
in the sub-Gaussian bound of
\cref{thm:finite-sample-regret-bds}.

We also compute population benchmarks using the source law
$P_{\mathrm s}$ of $(\diag(m^{-1/4},\bI_{m-1})X,Y)$ for $(X,Y)\sim P$,
corresponding to contracting every observation. This choice is justified because the average cost of contracting every observation converges almost surely to $(1-m^{-1/4})^2 < \eps^2$ as $n \to \infty$.
The estimators then receive $\bSigma_{P_{\mathrm s}}$
instead of $\bSigma_{P_{\mathrm{s},n}}$. Appendix~\ref{app:experiment-reductions} uses rotational symmetry to reduce the  population problems to optimization over the two-parameter family $\bA(\alpha,\beta)=\diag(\alpha,\beta\bI_{m-1})$. Under this reduction, ERM and SNR admit closed-form solutions, while DRO and MRO admit constant-size semidefinite programs which we solve using CVXPY \citep{diamond2016cvxpy,agrawal2018rewriting}. These computations benchmark the estimators at exact population second moments and provide diagnostics for the finite-sample optimization routines. For the population MRO benchmark, we break ties by minimizing the source risk $L(\cdot,P_{\mathrm s})$.

For \textbf{Setting A}, we supply radius $\rho = \eps = 1$, vary $d$ between $20$ and $400$, and plot the target population
regret $R(\hat{\bA}[P_{\mathrm s,n}],P)$ for $\hat{\bA} \in \{\ERM,\DRO_\rho,\MRO_\rho,\SNR_\rho\}$.  This setting illustrates the dimension-free
behavior of MRO and SNR and $\sqrt{d}$ growth of DRO regret.  For \textbf{Setting B}, we fix $d=400$ and vary the supplied radius $\rho\in[0.1\eps,20\eps]$. For small $\rho$, DRO, MRO, and SNR perform similarly
to ERM, and their relative ranking at $\rho=\eps$ matches
Setting A. As the radius $\rho$ varies, the best method depends on its
scale; SNR is strongest for radii below the budget $\eps$, MRO is strongest for most larger radii, and for very large radii
excessive regularization or robustification causes ERM to dominate.
This precise behavior is beyond the scope of our current
theoretical results and warrants further study.

For both settings, the plots report target population regret averaged over
$10$ independent trials, with error bars showing the minimum and maximum
over trials, and dashed curves showing the population benchmarks.  Error bars are nearly imperceptible due to low variance at the chosen sample size. 

See \cref{app:experiment-extras} for additional details and code, an alternative to Setting A where $d$ is fixed but the effective rank of $\bSigma_E^{P}$ is varied, and a final experiment with real-world network traffic data \citep{cesnettimeseries24}. We emphasize that the observed separations between MRO/SNR and ERM/DRO need not hold universally, with ample room remaining for finer-grained instance-dependent analysis.

\section{Concluding Remarks}

In this paper, we introduced a new framework for regret-optimal Wasserstein-robust decision-making, focused on the setting of least-squares regression. We fully characterized the complexity of robust estimation in the $\eps \to 0$ limit, via the condition number, and extended these asymptotic guarantees to all $\eps > 0$ using regret stability. While DRO is a natural robust algorithm, we proved and verified empirically that it can be outperformed by MRO and our new procedure, SNR, for high dimensional tasks. Moreover, we developed a new convex reformulation for MRO and an efficient proximal gradient method for SNR. 
Moving forward, we hope that this initial work spurs more research into characterizing and efficiently achieving regret-optimal performance for robust stochastic optimization.

\section*{Acknowledgements and AI Usage Statement}

This work was supported as a part of NCCR Automation, a National Centre
of Competence in Research, funded by the Swiss National Science Foundation (grant number
51NF40 225155). We also thank the Optimal Transport and Optimization Workshop (OT+OPT '26) attendees for their helpful feedback on an early version of this work.

We used ChatGPT (versions 5-6 Pro) for editing, checking and refining mathematical arguments, and tuning lower bound constructions. Version 5.5 suggested an improvement to \cref{lem:regret-stability} that allowed us to extend our analysis from MRO to SNR. It also suggested the argument used in Step 1 of the proof of \cref{thm:MRO-convex-reformulation}, enabling our convex reformulation of MRO. We have independently verified all mathematical arguments and take full responsibility for the content and correctness of this work.

\newpage

\bibliographystyle{apalike}
\IfFileExists{references.bib}{\bibliography{references}}{\bibliography{../references}}

\appendix
\crefalias{section}{appendix}
\crefname{appendix}{Appendix}{Appendices}
\Crefname{appendix}{Appendix}{Appendices}

\newpage

\startcontents[appendix]
\section*{Appendix Contents}
\vspace{-2mm}
\begingroup
\setlength{\parskip}{5pt}
\printcontents[appendix]{}{1}{}
\endgroup

\section{Linear Algebra Preliminaries}
\label{app:extra-prelims}

This appendix collects linear algebraic tools used repeatedly in the proofs.  We include short derivations when they are more convenient than a citation, and otherwise point to standard references.

\textbf{Additional notation.}
Write $\Sym^n$ for the space of symmetric $n \times n$ matrices. For a matrix $\bM\in\R^{a\times b}$, let $\range(\bM)$ denote its column space, so that $\range(\bM^\top)$ is its
row space, and write $\rank(\bM)\defeq\dim\range(\bM)$. Write $\vec(\bM) \in \R^{ab}$ for its column-stacked vectorization and
$\otimes$ for the Kronecker product.
For matrices of the same size, we write
$\langle\bM,\bN\rangle_\Frob\defeq\tr(\bM^\top\bN)$ for their Frobenius inner product.
We order the singular values of $\bM\in\R^{a\times b}$ as
$\sigma_1(\bM)\geq\cdots\geq\sigma_{\min\{a,b\}}(\bM)\geq0$,
including zeros, and write
$\sigma_{\min}(\bM)\defeq\sigma_{\min\{a,b\}}(\bM)$.
For $\bH\in\Sym^d$ and $k\in\{1,\ldots,d\}$,
let $\tr_k(\bH)$ denote the sum of its $k$ largest eigenvalues and write $\bH_+$ for its positive
semidefinite part, obtained by replacing negative eigenvalues by zero. Write $(t)_+\defeq\max\{t,0\}$ for scalars.
We write $0_n$ for the zero vector in $\R^n$ and $e_i$ for the $i$th
standard basis vector, with its ambient dimension specified
when needed.

\subsection{Basic Matrix Inequalities and Identities}

\begin{lemma}[Basic norm and projection inequalities]
\label{lem:basic-matrix-norms}
For dimension-compatible matrices,
\begin{align*}
    \|\bA\bB\bC\|_\Frob
    &\leq \|\bA\|_\op\,\|\bB\|_\Frob\,\|\bC\|_\op,\\
    \|\bA\bB\|_\op
    &\leq \|\bA\|_\op\|\bB\|_\op,\\
    \|\bA\bB\|_\Frob
    &\leq \|\bA\|_\op\|\bB\|_\Frob,\\
    \|\bA\|_\op &\leq \|\bA\|_\Frob, \qquad
    \|\bA\|_\Frob \leq \sqrt{\rank(\bA)}\,\|\bA\|_\op.
\end{align*}
If $\bPi$ is an orthogonal projection, then
\begin{equation*}
    \|\bM\bPi\|_\Frob\leq \|\bM\|_\Frob,
    \qquad
    \|\bM\|_\Frob^2
    =
    \|\bM\bPi\|_\Frob^2+
    \|\bM(\bI-\bPi)\|_\Frob^2.
\end{equation*}
Finally,
\begin{equation*}
    |\langle \bA,\bB\rangle_\Frob|
    \leq \|\bA\|_\Frob\|\bB\|_\Frob.
\end{equation*}
\end{lemma}

\begin{proof}
The first inequality holds for all unitarily invariant matrix norms, including the Frobenius norm; see, e.g., \citealp[Section 3.5, Exercise 4]{Horn_1991_Matrix}. The second is a standard submultiplicativity property; see, e.g.,  \citealp[Section 3.5, Exercise 3]{Horn_1991_Matrix}. The third inequality follows from the first by taking $\bC = \bI$. The comparisons between Frobenius and operator norms follow from the singular-value identities $\|\bA\|_\op=\sigma_1(\bA)$ and $\|\bA\|_\Frob^2=\sum_i\sigma_i(\bA)^2$.  The projection identities follow from $\bPi(\bI-\bPi)=\bZero$ and the Pythagorean theorem for the Frobenius inner product.  The last inequality is Cauchy--Schwarz for the Frobenius inner product.
\end{proof}

\begin{lemma}[Pseudoinverse and projection identities]
\label{lem:pseudoinverse-projection-facts}
Let $\bM\in\R^{r\times d}$.  Then $\bM^\dagger\bM$ is the orthogonal projection onto the row space of $\bM$, and $\bM\bM^\dagger$ is the orthogonal projection onto the column space of $\bM$. We always have $\bM^\dagger=\bM^\top(\bM\bM^\top)^\dagger$. If $\bM$ has full row rank, then
\begin{equation*}
    \bM^\dagger=\bM^\top(\bM\bM^\top)^{-1},
    \qquad
    \bM\bM^\dagger=\bI_r,
    \qquad
    \|\bM^\dagger\|_\op=\sigma_{\min}(\bM)^{-1}.
\end{equation*}
If $\bB\succeq\bZero$, then $\bB^\dagger\bB^2=\bB$. 
\end{lemma}

\begin{proof}
Take a compact singular value decomposition $\bM=\bU\bD\bV^\top$ over its nonzero singular values.  Then $\bM^\dagger=\bV\bD^{-1}\bU^\top$, so $\bM^\dagger\bM=\bV\bV^\top$, $\bM\bM^\dagger=\bU\bU^\top$. These are exactly the claimed row-space and column-space projections. Further, $\bM\bM^\top = \bU \bD^2 \bU^\top$, and so
\begin{equation*}
    \bM^\top(\bM\bM^\top)^\dagger = \bV \bD \bU^\top \left(\bU \bD^{-2} \bU^\top\right) = \bV \bD^{-1} \bU^\top = \bM^\dagger 
\end{equation*}
If $\bM$ has full row rank, then $\bM\bM^\top$ is invertible and the displayed formulas follow directly. The claim for $\bB\succeq\bZero$ follows from the eigenvalue decomposition of $\bB$.
\end{proof}

\begin{lemma}[Schur complement]
\label{lem:schur-complement}
Let
\begin{equation*}
    \bQ=\begin{pmatrix}
        \bA & \bB\\
        \bB^\top & \bC
    \end{pmatrix}
\end{equation*}
be symmetric.  If $\bA\succ\bZero$, then
\begin{equation*}
    \bQ\succeq\bZero
    \quad\Longleftrightarrow\quad
    \bC-\bB^\top\bA^{-1}\bB\succeq\bZero.
\end{equation*}
The analogous statement holds with $\bC\succ\bZero$ and Schur complement $\bA-\bB\bC^{-1}\bB^\top$.
\end{lemma}

\begin{proof}
Use the congruence
\begin{equation*}
\begin{pmatrix}
\bI&\bZero\\ -\bB^\top\bA^{-1}&\bI
\end{pmatrix}
\begin{pmatrix}
\bA&\bB\\ \bB^\top&\bC
\end{pmatrix}
\begin{pmatrix}
\bI&-\bA^{-1}\bB\\ \bZero&\bI
\end{pmatrix}
=
\begin{pmatrix}
\bA&\bZero\\ \bZero&\bC-\bB^\top\bA^{-1}\bB
\end{pmatrix}.
\end{equation*}
Congruence by an invertible matrix preserves positive (semi)definiteness.
\end{proof}

\begin{lemma}[Rectangular resolvent identities]
\label{lem:rectangular-resolvent-identities}
Let $\bM\in\R^{r\times d}$ and $\eta>\|\bM\|_\op^2$.  Then
\begin{align*}
    (\eta\bI_d-\bM^\top\bM)^{-1}\bM^\top
    &=
    \bM^\top(\eta\bI_r-\bM\bM^\top)^{-1},\\
    \bM(\eta\bI_d-\bM^\top\bM)^{-1}
    &=
    (\eta\bI_r-\bM\bM^\top)^{-1}\bM,\\
    \bM^\top\bM(\eta\bI_d-\bM^\top\bM)^{-1}
    &=
    \bM^\top(\eta\bI_r-\bM\bM^\top)^{-1}\bM,\\
    \eta(\eta\bI_d-\bM^\top\bM)^{-1}
    &=
    \bI_d+
    \bM^\top(\eta\bI_r-\bM\bM^\top)^{-1}\bM.
\end{align*}
\end{lemma}
\begin{proof}
We start by computing
\begin{align*}
    \bM^\top (\eta \bI_r - \bM \bM^\top) = (\eta \bI_d - \bM^\top \bM) \bM^\top.
\end{align*}
The terms in parentheses are both invertible due to our lower bound on $\eta$, so suitable left/right multiplication gives the first identity. The second identity follows by taking transposes, and the third follows by left multiplication of the second identity by $\bM^\top$. The fourth identity follows by combining the third with the identity $\eta(\eta\bI-\bC)^{-1}=\bI+\bC(\eta\bI-\bC)^{-1}$, which can be confirmed by multiplying both expressions on the right by $\eta \bI - \bC$.
\end{proof}

\begin{lemma}[Lipschitz continuity of the smallest singular value]
\label{lem:factor-perturbation}
Let $\bS,\bar{\bS}\in\R^{r\times d}$. Then
\begin{equation*}
    |
        \sigma_{\min}(\bar{\bS})
        -
        \sigma_{\min}(\bS)
    |
    \leq
    \|\bar{\bS}-\bS\|_\op.
\end{equation*}
\end{lemma}

\begin{proof}
This is the standard singular-value perturbation inequality; see, e.g., Theorem~3.3.16(c) of 
\cite{Horn_1991_Matrix}.
\end{proof}

\begin{lemma}[Row-space projection perturbation]
\label{lem:sin-thm-variant}
Let $\bM_0 \in \R^{r \times d}$ have full row rank, and let $\bM_1 = \bM_0 + \bDelta$ for some $\bDelta \in \R^{r \times d}$. Let $\bPi_0,\bPi_1 \in \R^{d \times d}$ denote orthogonal projections onto the row spaces of $\bM_0,\bM_1$, respectively. Then,
\begin{equation*}
    \|(\bI_d - \bPi_0)\bPi_1\|_\Frob \leq \|(\bI_d - \bPi_1)\bPi_0\|_\Frob \leq \frac{\|\bDelta\|_\Frob}{\sigma_{\min}(\bM_0)}.
\end{equation*}
\end{lemma}

\begin{proof}
Although we have not seen this precise statement in the literature, the left-hand side connects to the principal angles between row spaces and has been well examined, with the closely related Davis--Kahan--Wedin sine theorem providing a similar bound (see, e.g., Section 3 of \citealp{wedin1972perturbation}).

For the proof, define the complementary projections $\bPi_{i}^\perp \defeq \bI_d - \bPi_i$ for $i \in \{0,1\}$. To prove the first inequality, we compute
\begin{align*}
    \|\bPi_{0}^\perp\bPi_1\|_\Frob^2 & = \tr((\bI_d - \bPi_0)\bPi_1) \tag{$\bPi_0,\bPi_1$ projections}\\
    &= \tr(\bPi_1) - \tr(\bPi_0 \bPi_1)\\
    &\leq \tr(\bPi_0) - \tr(\bPi_0 \bPi_1)\\
    &= \|\bPi_{1}^\perp\bPi_0\|_\Frob^2.
\end{align*}
The inequality uses that $\tr(\bPi_1) \leq r = \tr(\bPi_0)$, since $\bM_0$ has full row rank.
Thus, it suffices to bound $\|\bPi_{1}^\perp\bPi_0\|_\Frob^2$.\medskip

Next, since $\bM_0$ is full row rank, we can take a thin singular value decomposition
\begin{equation*}
\bM_0 = \bU \bD \bV^\top,
\end{equation*}
where $\bU\in\OO^r$ is orthogonal, $\bD\in\R^{r\times r}$ is diagonal with strictly positive entries, and $\bV\in\R^{d\times r}$ has orthonormal columns. By this choice, $\bV = \bM_0^\top \bU \bD^{-1}$ and $\bPi_0 = \bV \bV^\top$, implying that
\begin{equation*}
    \|\bPi_1^\perp \bPi_0\|_\Frob^2 = \tr(\bV^\top \bPi_1^\perp \bV)  = \|\bPi_1^\perp \bV\|_\Frob^2.
\end{equation*}
Since $\bPi_1^\perp$ is the orthogonal projection onto the orthogonal complement of the row space $R_1 \subseteq \R^d$ of $\bM_1$, we have
\begin{equation*}
    \|\bPi_1^\perp \bV\|_\Frob = \min\bigl\{ \|\bV-\bW\|_\Frob : \text{columns of $\bW$ lie in } R_1 \bigr\}.
\end{equation*}
Consider the specific competitor $\bW \defeq \bM_1^\top \bU \bD^{-1}$, whose columns lie in $R_1$. We then have
\begin{equation*}
\bV - \bW
= (\bM_0^\top - \bM_1^\top)\bU\bD^{-1}
= -\bDelta^\top \bU\bD^{-1}.
\end{equation*}
Therefore,
\begin{align*}
\|\bPi_1^\perp \bV\|_\Frob &\leq \|\bDelta^\top \bU\bD^{-1}\|_\Frob \leq \|\bDelta\|_\Frob \|\bD^{-1}\|_\op
= \frac{\|\bDelta\|_\Frob}{\sigma_{\min}(\bM_0)}.
\end{align*}
Combining the steps above, we obtain
\begin{equation*}
\|(\bI_d-\bPi_0)\bPi_1\|_\Frob = \|\bPi_{0}^\perp\bPi_1\|_\Frob \leq \|\bPi_1^\perp \bPi_0\|_\Frob
= \|\bPi_1^\perp \bV\|_\Frob \le \frac{\|\bDelta\|_\Frob}{\sigma_{\min}(\bM_0)},
\end{equation*}
as desired.
\end{proof}

\subsection{Scalar, Tensor, and Spectral Variational Facts}

\begin{lemma}[Elementary scalar identities]
\label{lem:elementary-scalar-ineqs}
For $a_1, \dots, a_k \geq 0$,
\begin{equation*}
    \left(\sum_{i=1}^k a_i\right)^2
    \leq
    k\sum_{i=1}^k a_i^2.
\end{equation*}
For $A,B\geq0$,
\begin{equation*}
    \inf_{\delta>0}(A+\delta)\left(1+\frac{B}{\delta}\right)
    =
    (\sqrt A+\sqrt B)^2,
\end{equation*}
with optimizer $\delta=\sqrt{AB}$ when $AB>0$.  Finally, for any norm $\|\cdot\|$, we have
\begin{equation*}
    \inf_x \max\{\|x-x_0\|^2,\|x-x_1\|^2\}
    =
    \frac14\|x_0-x_1\|^2.
\end{equation*}
\end{lemma}

\begin{proof}
The first inequality is Cauchy--Schwarz. For the second identity, expand
\begin{equation*}
    (A+\delta)\left(1+\frac{B}{\delta}\right)
    =A+B+\delta+\frac{AB}{\delta}.
\end{equation*}
Basic calculus confirms that this is minimized at $\delta = \sqrt{AB}$ if $AB > 0$ or otherwise in the limit as $\delta \to 0$. In both cases, the infimum value is $(\sqrt{A} + \sqrt{B})^2$, as claimed.

The last identity follows because any $x$ with both distances at most $r$ must satisfy $\|x_0-x_1\|\leq 2r$, while the midpoint achieves radius $\|x_0-x_1\|/2$.
\end{proof}

\begin{lemma}[Kronecker trace identity]
\label{lem:kronecker-fact}
For matrices $\bA,\bB,\bC$ with dimensions such that the following products are well-defined,
\begin{equation*}
    \tr(\bA\bC\bB\bC^\top)
    =
    \vec(\bC)^\top(\bB^\top\otimes\bA)\vec(\bC).
\end{equation*}
\end{lemma}

\begin{proof}
This follows from the matrix vectorization identities $\vec(\bA\bC\bB)=(\bB^\top\otimes\bA)\vec(\bC)$ and $\tr(\bU\bV^\top)=\vec(\bV)^\top\vec(\bU)$; see, e.g., \citealp[Section 10.2]{petersen2008matrix}.
\end{proof}

\begin{lemma}[Variational matrix formulas]
\label{lem:variational-matrix-formulas}
For any matrix $\bA$,
\begin{equation*}
    \|\bA\|_\Frob
    =
    \sup_{\|\bB\|_\Frob\leq1}\langle \bA,\bB\rangle_\Frob.
\end{equation*}
For any $\bH\in\Sym^d$ and $k\in\{1,\ldots,d\}$, we have the Ky Fan variational formula
\begin{equation*}
    \tr_k(\bH)
    =
    \max_{\substack{\bPi^2=\bPi=\bPi^\top\\ \tr(\bPi)=k}}
    \tr(\bH\bPi),
\end{equation*}
and equivalently
\begin{equation*}
    \tr_k(\bH)
    =
    \min_{\tau\in\R,\ \bY\succeq\bZero,\ \bY\succeq \bH-\tau\bI_d}
    k\tau+\tr(\bY).
\end{equation*}
\end{lemma}

\begin{proof}
The Frobenius formula simply is Hilbert space duality.  The projection formula is Ky Fan's maximum principle; see, e.g., \citealp[Section 3.4]{Horn_1991_Matrix}. For the epigraph formula, diagonalize $\bH = \sum_{i=1}^d \lambda_i u_i u_i^\top$, for an orthonormal eigenbasis $\{u_i\}_{i=1}^d$.
For fixed $\tau$, any feasible $\bY$ satisfies $u_i^\top\bY u_i \geq \max\{\lambda_i-\tau,0\}$ for all $i$. Hence,
\[
    \tr(\bY)
    \geq
    \sum_{i=1}^d(\lambda_i-\tau)_+.
\]
Equality is attained by
$\bY=(\bH-\tau\bI_d)_+$.
Thus the inner minimum equals
\[
    k\tau+\sum_{i=1}^d(\lambda_i-\tau)_+,
\]
whose minimum over $\tau$ is the sum of the $k$ largest
eigenvalues of $\bH$.
\end{proof}

\section{Bures--Wasserstein Geometry and Regression}
\label{app:Bures-geometry}

This appendix develops the second-moment formulation of least-squares regression in Bures space and collects some basic facts of Bures--Wasserstein geometry used throughout the proofs.

\subsection{Common Block Decompositions}
\label{app:block-decomp}

Throughout the proofs, we frequently apply two types of matrix block decompositions. Recall that $d=p+m$. If $\bSigma\in\Sym^d$ is symmetric, we write
\begin{equation*}
    \bSigma = \begin{pmatrix}
        \bSigma_X & \bSigma_{XY}\\
        \bSigma_{YX} & \bSigma_Y
    \end{pmatrix},
    \qquad
    \bSigma_X \in \Sym^p,
    \quad
    \bSigma_{XY} = \bSigma_{YX}^\top\in\R^{p\times m},
    \quad
    \bSigma_Y \in \Sym^m.
\end{equation*}
This matches our existing notation for second moments when $\bSigma = \bSigma_P$. We employ this decomposition when $\bSigma$ is a second-moment matrix, a nearby symmetric matrix, or a tangent direction to the positive semidefinite cone.

If $\bS \in \R^{d\times k}$ is introduced as a square-root factor for a matrix $\bSigma \in \PSD^d$, i.e., $\bSigma = \bS \bS^\top$, or as the difference between two such factors, then we often decompose
\begin{equation*}
    \bS=\begin{pmatrix}\bS_X\\ \bS_Y\end{pmatrix},
    \qquad
    \bS_X\in\R^{p \times k},
    \quad
    \bS_Y\in\R^{m \times k}.
\end{equation*}
This decomposition arises naturally for Bures distance computations. For intuition when working with such blocks, note that if $Z = (X,Y) = (\bS_X W, \bS_Y W)$ for $W \sim \cN(0,\bI_k)$, then $\E[ZZ^\top] = \bS \bS^\top$.

\subsection{Regression in Bures Space}
\label{app:Bures-regression}

Since all risks considered in the paper depend only on second moments, we extend the basic definitions from the main body to Bures space.  For $\bSigma\in\PSD^d$ and $\bA \in \cA = \R^{m \times p}$, define
\begin{align*}
    \bA_\bSigma &\defeq \bSigma_{YX}\bSigma_X^\dagger,\\
    L(\bA,\bSigma)&\defeq
    \tr(\bA^\top\bA\bSigma_X)-2\tr(\bA\bSigma_{XY})+\tr(\bSigma_Y),\\
    R(\bA,\bSigma)&\defeq L(\bA,\bSigma)-L(\bA_\bSigma,\bSigma),\\
    \bSigma_E&\defeq
    \begin{pmatrix}\bA_\bSigma&-\bI_m\end{pmatrix}
    \,\bSigma\,
    \begin{pmatrix}\bA_\bSigma&-\bI_m\end{pmatrix}^\top.
\end{align*}

For $P \in \cP^*$, these agree with the distributional definitions in the main body when $\bSigma=\bSigma_P$. We note that the residual second moment can be simplified as follows:
\begin{align*}
    \bSigma_E
    &=
    \begin{pmatrix}\bA_\bSigma&-\bI_m\end{pmatrix}
    \begin{pmatrix}
        \bSigma_X & \bSigma_{XY}\\
        \bSigma_{YX} & \bSigma_Y
    \end{pmatrix}
    \begin{pmatrix}\bA_\bSigma&-\bI_m\end{pmatrix}^\top\\
    &= \bA_\bSigma\bSigma_X\bA_\bSigma^\top
    -\bA_\bSigma\bSigma_{XY}
    -\bSigma_{YX}\bA_\bSigma^\top
    +\bSigma_Y\\
    &=
    \bSigma_{YX}\bSigma_X^\dagger
    \bSigma_X\bSigma_X^\dagger\bSigma_{XY}
    -2\bSigma_{YX}\bSigma_X^\dagger\bSigma_{XY}
    +\bSigma_Y\\
    &=
    \bSigma_Y-\bSigma_{YX}\bSigma_X^\dagger\bSigma_{XY},
\end{align*}
where the last equality uses that $\bSigma_X^\dagger\bSigma_X\bSigma_X^\dagger = \bSigma_X^\dagger$.

The next lemma records several useful identities for these Bures-space regression quantities in terms of a square-root factor for $\bSigma$.

\begin{lemma}[Regression factor identities and regret formula]
\label{lem:regret-projection-formula}
Let $\bSigma\in\PSD^d$, and fix any $\bS\in\R^{d\times r}$ such that $\bSigma=\bS\bS^\top$.  Let $\bPi_X=\bS_X^\dagger\bS_X$ be the orthogonal projection onto the row space of $\bS_X$.  Then
\begin{equation*}
    \bA_\bSigma
    =
    \bSigma_{YX}\bSigma_X^\dagger
    =
    \bS_Y\bS_X^\dagger.
\end{equation*}
For the chosen factor $\bS$, the corresponding residual factor
\[
    \bE
    \defeq
    \bS_Y-\bA_\bSigma\bS_X
    =
    \bS_Y(\bI-\bPi_X)
\]
satisfies
\begin{equation*}
    \bE \bS_X^\top=\bZero,
    \qquad
    \bE \bPi_X=\bZero,
    \qquad
    \bE \bE^\top=\bSigma_E.
\end{equation*}
Finally, for every $\bA \in \cA$, we have $L(\bA,\bSigma)=\|\bA\bS_X-\bS_Y\|_\Frob^2$ and
\begin{equation*}
    R(\bA,\bSigma)
    =
    \|(\bS_Y-\bA\bS_X)\bPi_X\|_\Frob^2
    =
    \|(\bA-\bA_\bSigma)\bS_X\|_\Frob^2.
\end{equation*}
\end{lemma}

\begin{proof}
By \cref{lem:pseudoinverse-projection-facts}, $\bS_X^\dagger=\bS_X^\top(\bS_X\bS_X^\top)^\dagger$, and so
\begin{equation*}
    \bS_Y\bS_X^\dagger
    =\bS_Y\bS_X^\top(\bS_X\bS_X^\top)^\dagger
    =\bSigma_{YX}\bSigma_X^\dagger
    =\bA_\bSigma.
\end{equation*}
Thus, $\bA_\bSigma\bS_X = \bS_Y \bS_X^\dagger \bS_X = \bS_Y\bPi_X$, and hence $\bE = \bS_Y(\bI-\bPi_X)$. Since $\bPi_X\bS_X^\top=\bS_X^\top$ and
$(\bI-\bPi_X)\bPi_X=\bZero$, we obtain
\begin{equation*}
    \bE \bS_X^\top=\bZero,
    \qquad
    \bE \bPi_X=\bZero.
\end{equation*}
Moreover, $\begin{pmatrix}\bA_\bSigma&-\bI_m\end{pmatrix}\,\bS = -\bE$, and so
\begin{equation*}
    \bE \bE^\top
    =
    \begin{pmatrix}\bA_\bSigma&-\bI_m\end{pmatrix}
    \bSigma
    \begin{pmatrix}\bA_\bSigma&-\bI_m\end{pmatrix}^\top
    =
    \bSigma_E.
\end{equation*}
Finally,
\begin{equation*}
    L(\bA,\bSigma)
    =\tr((\bA\bS_X-\bS_Y)(\bA\bS_X-\bS_Y)^\top)
    =\|\bA\bS_X-\bS_Y\|_\Frob^2.
\end{equation*}
We decompose this norm using $\bPi_X$ and $\bI-\bPi_X$:
\begin{align*}
    \|\bA\bS_X-\bS_Y\|_\Frob^2
    &=
    \|(\bA\bS_X-\bS_Y)\bPi_X\|_\Frob^2
    +
    \|\bS_Y(\bI-\bPi_X)\|_\Frob^2.
\end{align*}
The second term is independent of $\bA$ and the first is zero for $\bA=\bA_\bSigma$. Thus,
\begin{equation*}
    R(\bA,\bSigma)
    =
    \|(\bA\bS_X-\bS_Y)\bPi_X\|_\Frob^2.
\end{equation*}
Finally, since $\bS_X\bPi_X=\bS_X$ and
$\bA_\bSigma\bS_X=\bS_Y\bPi_X$,
\begin{equation*}
    (\bA\bS_X-\bS_Y)\bPi_X
    =
    (\bA-\bA_\bSigma)\bS_X,
\end{equation*}
which gives the final equality.
\end{proof}

Next, set
\begin{equation*}
    \cC^*
    \defeq
    \{\bSigma_P:P\in\cP^*\}
    =
    \{\bSigma\in\PSD^d:\bSigma_X\succ\bZero\}.
\end{equation*}
Using the second-moment formulation of the risk together with
\cref{lem:W2-Bures-equivalence}, the condition number and our main-body estimators admit equivalent Bures-space formulations. For $\bSigma\in\cC^*$, define
\begin{equation}
\label{eq:condition-number-cov-def}
    R^\star_\eps(\bSigma)
    \defeq
    \inf_{\bA\in\cA}
    \sup_{\bSigma'\in\cB_\eps(\bSigma)}
    R(\bA,\bSigma'),
    \qquad
    \kappa(\bSigma)
    \defeq
    \limsup_{\eps\downarrow0}
    \eps^{-2}R^\star_\eps(\bSigma).
\end{equation}
Then, by \eqref{eq:local-minimax} and
\cref{lem:W2-Bures-equivalence}, $ \kappa_P=\kappa(\bSigma_P)$ for $P \in \cP^*$. Likewise, the estimators can be written as
\begin{align*}
    \ERM[\bSigma]
    &\in
    \argmin_{\bA\in\cA}
    L(\bA,\bSigma),\\
    \DRO_\eps[\bSigma]
    &\in
    \argmin_{\bA\in\cA}
    \sup_{\bSigma'\in\cB_\eps(\bSigma)}
    L(\bA,\bSigma'),\\
    \MRO_\eps[\bSigma]
    &\in
    \argmin_{\bA\in\cA}
    \sup_{\bSigma'\in\cB_\eps(\bSigma)}
    R(\bA,\bSigma'),\\
    \SNR_\eps[\bSigma]
    &\in
    \argmin_{\bA\in\cA}
        L(\bA,\bSigma)
        +\eps^2\|\bA\|_\op^2.
\end{align*}

\subsection{Bures--Wasserstein Geometry}
\label{app:Bures-proofs}

We now turn to proofs of the \cref{ssec:prelims} lemmas on the Bures metric.

\begin{proof}[Proof of \cref{lem:Bures-Procrustes}]
Given $\bA,\bB \in \PSD^d$, the identity
\begin{align}
     \sB(\bA,\bB) &= \min_{\bU \in \OO^d} \|\bA^{\frac{1}{2}} - \bB^{\frac{1}{2}}\bU\|_\Frob \label{eq:Bures-Procrustes-1}\\
     &= \min_{\substack{\bS,\bT \in \R^{d \times d}\\ \bA = \bS \bS^\top\!\!, \,\bB = \bT \bT^\top}} \|\bS - \bT\|_\Frob.\label{eq:Bures-Procrustes-2}
\end{align}
is well-known; see, e.g., \citet[Theorem 1]{bhatia2019bures}. 

Suppose $\sB(\bA,\bB) \leq \eps$. Fix $\bU \in \OO^d$ minimizing \eqref{eq:Bures-Procrustes-1}, so that $\|\bA^{1/2} - \bB^{1/2}\, \bU\|_\Frob \leq \eps$. Setting $\bDelta \defeq \bA^{1/2}\,\bU^\top - \bB^{1/2}$, we then have $\|\bDelta\|_\Frob = \|\bDelta \bU\|_\Frob \leq \eps$ and
\begin{equation*}
    (\bB^{\frac{1}{2}} + \bDelta)(\bB^{\frac{1}{2}} + \bDelta)^\top = \bA^{\frac{1}{2}} \bU^\top \bU \bA^{\frac{1}{2}} = \bA,
\end{equation*}
as desired. On the other hand, if such $\bDelta$ exists, then $\bS \defeq \bB^{\frac{1}{2}} + \bDelta$ and $\bT \defeq \bB^{\frac{1}{2}}$ are feasible square-root factors for \eqref{eq:Bures-Procrustes-2}, and so
\begin{align*}
    \sB(\bA,\bB) \leq \|\bS - \bT\|_\Frob = \|\bDelta\|_\Frob \leq \eps.
\end{align*}

If $\bA$ and $\bB$ are simultaneously diagonalizable, then $\bA^{1/2}$ and $\bB^{1/2}$ commute. In this case, we have
\begin{align*}
    \sB(\bA,\bB)^2 &= \tr(\bA) + \tr(\bB) - 2\tr\bigl((\bA^{\frac{1}{2}}\bB\bA^{\frac{1}{2}})^{\frac{1}{2}}\bigr)\\
    &= \tr(\bA) + \tr(\bB) - 2\tr(\bA^{\frac{1}{2}}\bB^{\frac{1}{2}})\\
    &= \|\bA^{\frac{1}{2}} - \bB^{\frac{1}{2}}\|_\Frob^2,
\end{align*}
implying that the minimum in \eqref{eq:Bures-Procrustes-1} is attained by $\bU = \bI_d$, as claimed.
\end{proof}

\begin{proof}[Proof of \cref{lem:W2-Bures-equivalence}]
The second equality follows directly from \cref{lem:Bures-Procrustes}. We now prove the first.

Proposition 1 of \cite{nguyen2021mean} establishes a related claim for the augmented Gelbrich distance over mean--covariance pairs, assuming non-singular covariance matrices. Here we give a direct argument for the Bures case with uncentered and potentially singular second-moment matrices.

\textbf{$\bm{\subseteq}$ direction.}
Fix $Q \in \cB_\eps(P)$ and any coupling of random variables $(Z,Z')$ with $Z \sim P$ and $Z' \sim Q$.
Setting $\bC \defeq \E[Z(Z')^\top]$, the block second-moment matrix
\[
    \begin{pmatrix}
        \bSigma_P & \bC\\
        \bC^\top & \bSigma_Q
    \end{pmatrix}
\]
is PSD. Hence, there exists a contraction $\bK \in \R^{d \times d}$ with
$\|\bK\|_\op \leq 1$ such that $\bC = \bSigma_P^{1/2} \bK \bSigma_Q^{1/2}$ (see, e.g., \citealp[Proposition 1.3.2]{bhatia2009positive}). Therefore, by duality between the operator norm $\|\cdot\|_\op$ and nuclear norm $\|\cdot\|_*$ (the sum of singular values), we have
\begin{align*}
    \tr(\bC) = \tr\bigl(\bK \bSigma_Q^{1/2} \bSigma_P^{1/2}\bigr)
    &\leq
    \bigl\|\bSigma_Q^{1/2}\bSigma_P^{1/2}\bigr\|_* =
    \tr\left(
        (\bSigma_P^{1/2}
        \bSigma_Q
        \bSigma_P^{1/2})^{\frac{1}{2}}
    \right).
\end{align*}
It then follows from definition of $\sB$ that
\begin{align*}
    \E\|Z-Z'\|^2 =
    \tr(\bSigma_P)+\tr(\bSigma_Q)-2\tr(\bC) \geq
    \sB(\bSigma_P,\bSigma_Q)^2.
\end{align*}
Infimizing over couplings gives
$\Wtwo(P,Q)\geq\sB(\bSigma_P,\bSigma_Q)$.
Thus $\bSigma_Q\in\cB_\eps(\bSigma_P)$.

\textbf{$\bm{\supseteq}$ direction.} Fix $\bSigma \in \cB_\eps(\bSigma_P)$.  It suffices to identify $Q \in \cB_\eps(P)$ such that $\bSigma = \bSigma_Q$. Abbreviate $\bB_P \defeq \bSigma_P^{1/2}$. By the Bures bound and \cref{lem:Bures-Procrustes}, there exists a factor $\bB \in \R^{d \times d}$ such that $\|\bB_P - \bB\|_\Frob \leq \eps$ and $\bB \bB^\top = \bSigma$. Let $\bPi \in \R^{d \times d}$ be the orthogonal projection onto $\range(\bSigma_P) = \range(\bB_P)$. By \cref{lem:pseudoinverse-projection-facts}, we have $\bPi = \bB_P \bB_P^\dagger = \bB_P^\dagger \bSigma_P \bB_P^\dagger$, where $\dagger$ denotes the Moore--Penrose pseudoinverse. Let $Z \sim P$ and $W \sim \cN(0,\bI_d - \bPi)$ be independent. Now consider the random variable $U \defeq \bB_P^\dagger Z + W$, constructed so that
\begin{align*}
    \E\bigl[UU^\top\bigr] &= \bB_P^\dagger \bSigma_P \bB_P^\dagger + \bI_d - \bPi = \bPi + \bI_d - \bPi = \bI_d\\
    \bB_P U &= \bPi Z + \bB_P W = Z.
\end{align*}
Indeed, $Z$ is supported on the range of $\bSigma_P$ and $W$ is supported on the null space of $\bB_P$.
We take $Q$ as the law of $Z' \defeq \bB U$. Indeed, since $U$ has identity second moment, we have $\bSigma_Q = \bB \bB^\top = \bSigma$. Moreover, the coupling $(Z,Z')$ certifies that
\begin{align*}
    \Wtwo(P,Q)^2 \leq \E[\|Z - Z'\|^2] = \E[\|(\bB_P - \bB)U\|^2] = \|\bB_P - \bB\|_\Frob^2 \leq \eps^2.
\end{align*}
Thus, $Q \in \cB_\eps(P)$, as desired.
\end{proof}

Lastly, we apply \cref{lem:Bures-Procrustes} to prove a simple upper bound on the Bures distance.

\begin{lemma}[Bures distance upper bound]
\label{lem:Bures-quadratic-upper}
For $\bA,\bB \in \PSD^d$ with
$\mathrm{range}(\bB) \subseteq \mathrm{range}(\bA)$, we have
\begin{equation*}
    \sB(\bA,\bB)^2 \leq \tr\bigl((\bB-\bA)\bA^\dagger(\bB-\bA)\bigr).
\end{equation*}
\end{lemma}

\begin{proof}
We assume without loss of generality that $\bA \succ \bZero$. Indeed, if $\bA = \bZero$, then the range constraint forces $\bB = \bZero$ and the claim holds trivially. Otherwise, since $\range(\bB)\subseteq\range(\bA)$, both matrices
vanish on $\range(\bA)^\perp$. We may therefore restrict to
$\range(\bA)$, where $\bA$ is positive definite and $\bA^\dagger$
agrees with its inverse.

Set $\bC \defeq \bA^{-1/2}\bB\bA^{-1/2} \succeq \bZero$. By construction, we have
\begin{equation*}
    \bB =
    \bA^{1/2}\bC\bA^{1/2}
    = \bigl(\bA^{1/2}\bC^{1/2}\bigr)
    \bigl(\bA^{1/2}\bC^{1/2}\bigr)^\top.
\end{equation*}
Using the square-root characterization of the Bures distance (\cref{lem:Bures-Procrustes}) with the particular square-root factors $\bA^{1/2}$ and $\bA^{1/2}\bC^{1/2}$, we obtain
\begin{equation*}
    \sB(\bA,\bB)^2 \leq
    \bigl\|\bA^{1/2}(\bC^{1/2}-\bI_d)\bigr\|_\Frob^2.
\end{equation*}
Since $\bC^{1/2} - \bI_d = (\bC - \bI_d)(\bC^{1/2} + \bI_d)^{-1}$ and $\|(\bC^{1/2} + \bI_d)^{-1}\|_\op \leq 1$, we further bound
\begin{align*}
    \bigl\|\bA^{1/2}(\bC^{1/2}-\bI_d)\bigr\|_\Frob &\leq \bigl\|\bA^{1/2}(\bC-\bI_d)\bigr\|_\Frob \|(\bC^{1/2} + \bI_d)^{-1}\|_\op \tag{\cref{lem:basic-matrix-norms}}\\
    &\leq \bigl\|\bA^{1/2}(\bC-\bI_d)\bigr\|_\Frob.
\end{align*}
By our choice of $\bC$, we have $\bA^{1/2}(\bC-\bI_d) = (\bB - \bA)\bA^{-1/2}$, and thus
\begin{equation*}
    \bigl\|\bA^{1/2}(\bC-\bI_d)\bigr\|_\Frob^2
    = \tr\bigl((\bB-\bA)\bA^{-1}(\bB-\bA)\bigr).
\end{equation*}
Combining the above gives the claimed bound on $\sB(\bA,\bB)$.
\end{proof}

Finally, we establish Lipschitz continuity of the square root
of the smallest feature eigenvalue.

\begin{lemma}[Bures continuity of the smallest feature eigenvalue]
\label{lem:Bures-lambda-min-continuity}
For $\bSigma,\bar{\bSigma}\in\PSD^d$, we have
\begin{equation*}
    \left|
        \sqrt{\lambda_{\min}(\smash{\bar{\bSigma}_X)}}
        -
        \sqrt{\lambda_{\min}(\bSigma_X)}
    \right|
    \leq
    \sB(\bar{\bSigma},\bSigma).
\end{equation*}
In particular, $\lambda_{\min}(\bSigma_X)$ is continuous with respect to the Bures metric.
\end{lemma}

\begin{proof}
By \cref{lem:Bures-Procrustes}, there exist square-root factors
$\bS,\bar{\bS}\in\R^{d\times d}$ of $\bSigma$ and
$\bar{\bSigma}$ such that
\begin{equation*}
    \|\bar{\bS}-\bS\|_\Frob
    =
    \sB(\bar{\bSigma},\bSigma).
\end{equation*}
Using the \cref{app:block-decomp} block decompositions, we have
$\bSigma_X=\bS_X\bS_X^\top$ and
$\bar{\bSigma}_X=\bar{\bS}_X\bar{\bS}_X^\top$. Hence,
by \cref{lem:factor-perturbation},
\begin{align*}
    \left|
        \sqrt{\lambda_{\min}(\smash{\bar{\bSigma}_X)}}
        -
        \sqrt{\lambda_{\min}(\bSigma_X)}
    \right|
    &=
    \left|
        \sigma_{\min}(\bar{\bS}_X)
        -
        \sigma_{\min}(\bS_X)
    \right|\\
    &\leq
    \|\bar{\bS}_X-\bS_X\|_\op\\
    &\leq
    \|\bar{\bS}-\bS\|_\Frob\\
    &=
    \sB(\bar{\bSigma},\bSigma).
\end{align*}
The continuity claim follows immediately.
\end{proof}

\section{DRO Duality and Symmetry Reductions}

This appendix establishes a dual formulation for the inner DRO
maximization problem and an averaging principle for DRO, MRO,
and SNR under orthogonal symmetries. We use $\sup \varnothing = -\infty$.

\subsection{Strong Duality for the Inner DRO Problem}

We first establish a helper lemma, which will be applied both for the inner DRO problem below and for the fixed-projection problem in \cref{prf:MRO-convex-reformulation}. Very similar arguments appear in \cite{shafieezadeh2018wasserstein,nguyen2023bridging}, but we provide a self-contained proof to account for the case where $r < d$ (untreated in those works).

\begin{lemma}[Quadratic-ball duality]
\label{lem:quadratic-ball-duality}
Let $\bM\in\R^{k\times d}$ and
$\tilde{\bW}\in\R^{d\times r}$, where $k,r\geq1$.
Set $b\defeq\|\bM\|_\op^2$.
For every $\eta>b$, we have
\begin{equation}
\label{eq:quadratic-ball-envelope}
    \sup_{\bW\in\R^{d\times r}}
    \left\{
        \|\bM\bW\|_\Frob^2
        -\eta\|\bW-\tilde{\bW}\|_\Frob^2
    \right\}
    =
    \eta\tr\left(
        (\eta\bI_k-\bM\bM^\top)^{-1}
        \bM\tilde{\bW}\tilde{\bW}^\top\bM^\top
    \right).
\end{equation}
Moreover, for every $t\in\R$, with
$\sup\varnothing=-\infty$, we have
\begin{equation}
\label{eq:quadratic-ball-duality}
\begin{aligned}
    \sup_{\substack{
        \bW\in\R^{d\times r}\\
        \|\bW-\tilde{\bW}\|_\Frob^2\leq t
    }}
    \|\bM\bW\|_\Frob^2 &=
    \sup_{\bW\in\R^{d\times r}}
    \inf_{\eta\geq0}
    \left\{
        \|\bM\bW\|_\Frob^2
        +\eta\bigl(t-\|\bW-\tilde{\bW}\|_\Frob^2\bigr)
    \right\}\\
    &=
    \inf_{\eta\geq0}
    \sup_{\bW\in\R^{d\times r}}
    \left\{
        \|\bM\bW\|_\Frob^2
        +\eta\bigl(t-\|\bW-\tilde{\bW}\|_\Frob^2\bigr)
    \right\}\\
    &=
    \inf_{\eta>b}
    \eta\left(
        t+
        \tr\left(
            (\eta\bI_k-\bM\bM^\top)^{-1}
            \bM\tilde{\bW}\tilde{\bW}^\top\bM^\top
        \right)
    \right).
\end{aligned}
\end{equation}
\end{lemma}

\begin{proof}
Fix $t\in\R$. For $\eta\geq0$, define
\[
    \Phi_t(\eta)
    \defeq
    \sup_{\bW\in\R^{d\times r}}
    \left\{
        \|\bM\bW\|_\Frob^2
        +\eta\bigl(t-\|\bW-\tilde{\bW}\|_\Frob^2\bigr)
    \right\}.
\]
We first compute this supremum for $\eta>b$.
Expanding the quadratic expression gives
\begin{align*}
    &\|\bM\bW\|_\Frob^2
    -\eta\|\bW-\tilde{\bW}\|_\Frob^2 =
    -\tr\bigl(
        \bW^\top(\eta\bI_d-\bM^\top\bM)\bW
    \bigr)
    +2\eta\langle\bW,\tilde{\bW}\rangle_\Frob
    -\eta\|\tilde{\bW}\|_\Frob^2.
\end{align*}
Since $\eta\bI_d-\bM^\top\bM\succ\bZero$,
this expression is strictly concave in $\bW$.
Its unique maximizer satisfies
\[
    (\eta\bI_d-\bM^\top\bM)\bW
    =\eta\tilde{\bW},
\]
and therefore equals $\eta(\eta\bI_d-\bM^\top\bM)^{-1}\tilde{\bW}$. Substitution yields
\begin{align}
    \Phi_t(\eta)
    &=
    \eta t
    +\eta^2\tr\left(
        \tilde{\bW}^\top
        (\eta\bI_d-\bM^\top\bM)^{-1}
        \tilde{\bW}
    \right)
    -\eta\|\tilde{\bW}\|_\Frob^2\notag\\
    &=
    \eta t
    +\eta\tr\left(
        \tilde{\bW}^\top
        \left[
            \eta(\eta\bI_d-\bM^\top\bM)^{-1}-\bI_d
        \right]
        \tilde{\bW}
    \right)\notag\\
    &=
    \eta t
    +\eta\tr\left(
        (\eta\bI_k-\bM\bM^\top)^{-1}
        \bM\tilde{\bW}\tilde{\bW}^\top\bM^\top
    \right),\label{eq:quadratic-ball-Phi-t-formula}
\end{align}
where the last equality uses
\cref{lem:rectangular-resolvent-identities}
and cyclicity of trace.
Subtracting $\eta t$ proves
\eqref{eq:quadratic-ball-envelope}.

We next prove the final equality in \eqref{eq:quadratic-ball-duality}. Given \eqref{eq:quadratic-ball-Phi-t-formula}, it suffices to show
\[
    \inf_{\eta\geq0}\Phi_t(\eta)
    =
    \inf_{\eta>b}\Phi_t(\eta).
\]
For $0\leq\eta<b$, varying one column of $\bW$
along a top eigenvector of $\bM^\top\bM$ makes
the expression defining $\Phi_t(\eta)$ unbounded above,
since its leading quadratic coefficient is $b-\eta>0$.
Thus $\Phi_t(\eta)=+\infty$.
At the endpoint $\eta=b$, observe that as
$\eta\downarrow b$, the functions $ \bW\mapsto\|\bM\bW\|_\Frob^2-\eta\|\bW-\tilde{\bW}\|_\Frob^2$ increase pointwise to their values at $b$, so their suprema converge to the supremum at $b$.
Further, $\eta \mapsto \eta t$ is continuous, so we obtain
\[
    \lim_{\eta\downarrow b}\Phi_t(\eta)
    =\Phi_t(b),
\]
allowing $+\infty$.
Hence excluding all $\eta\leq b$ does not change
the infimum.

We now return to the initial constrained supremum in \eqref{eq:quadratic-ball-duality}.
For each fixed $\bW$, we have
\[
    \inf_{\eta\geq0}
    \left\{
        \|\bM\bW\|_\Frob^2
        +\eta\bigl(t-\|\bW-\tilde{\bW}\|_\Frob^2\bigr)
    \right\}
    =
    \begin{cases}
        \|\bM\bW\|_\Frob^2,
        &\|\bW-\tilde{\bW}\|_\Frob^2\leq t,\\
        -\infty,
        &\text{otherwise}.
    \end{cases}
\]
This proves the first equality in
\eqref{eq:quadratic-ball-duality}.

For the remaining equality, suppose first that $t>0$ and set
\[
    \gamma
    \defeq
    \max_{\substack{
        \bW\in\R^{d\times r}\\
        \|\bW-\tilde{\bW}\|_\Frob^2\leq t
    }}
    \|\bM\bW\|_\Frob^2.
\]
This maximum exists and is finite because the feasible
set is compact and the objective is continuous.
By definition of $\gamma$, the quadratic implication
\[
    t-\|\bW-\tilde{\bW}\|_\Frob^2\geq0
    \quad\Longrightarrow\quad
    \gamma-\|\bM\bW\|_\Frob^2\geq0
\]
holds for every $\bW$.
The first inequality is strictly satisfied at
$\bW=\tilde{\bW}$.
After vectorizing $\bW$, the S-lemma
\citep[Theorem~2.2 and Section~8.1]{polik2007survey}
thus gives some $\eta\geq0$ such that
\[
    \gamma-\|\bM\bW\|_\Frob^2
    -\eta\bigl(t-\|\bW-\tilde{\bW}\|_\Frob^2\bigr)
    \geq0
    \qquad\forall\,\bW\in\R^{d\times r}.
\]
Equivalently, $\Phi_t(\eta)\leq\gamma$.
Conversely, $\Phi_t(\eta)\geq\gamma$ for every
$\eta\geq0$, since the additional term in the definition of $\Phi_t$ is nonnegative
on the feasible set.
Consequently,
\[
    \gamma=\inf_{\eta\geq0}\Phi_t(\eta).
\]
This yields the second equality in \eqref{eq:quadratic-ball-duality} when $t > 0$.

If $t=0$, the only feasible point is
$\bW=\tilde{\bW}$, so the constrained supremum is
$\|\bM\tilde{\bW}\|_\Frob^2$.
Moreover, the formula \eqref{eq:quadratic-ball-Phi-t-formula} established for $\Phi_t(\eta)$ implies that
\[
    \inf_{\eta \geq 0} \Phi_0(\eta) \leq \lim_{\eta\to\infty}\Phi_0(\eta)
    =
    \|\bM\tilde{\bW}\|_\Frob^2,
\]
using that $\eta(\eta\bI_k-\bM\bM^\top)^{-1}\to\bI_k$.
Since $\Phi_0(\eta)$ is at least the constrained
supremum for every $\eta\geq0$, equality follows.

Finally, if $t<0$, the constrained feasible set is
empty and its supremum equals $-\infty$.
Meanwhile, the formula \eqref{eq:quadratic-ball-Phi-t-formula} for $\Phi_t$ gives
\[
    \Phi_t(\eta)=\eta t+O(1) \to -\infty
    \qquad\text{as }\eta\to\infty,
\]
so $\inf_{\eta\geq0}\Phi_t(\eta)=-\infty$ as well.
\end{proof}

We now apply this to the inner maximization problem in the DRO
formulation. For $\bA\in\cA$, write
\[
    \bM_\bA
    \defeq
    \begin{pmatrix}\bA&-\bI_m\end{pmatrix}
    \in\R^{m\times d},
\]
so that $L(\bA,P)=\E_P\|\bM_\bA Z\|^2$.
Direct multiplication gives
\[
    \bM_\bA\bM_\bA^\top=\bA\bA^\top+\bI_m,
    \qquad
    \|\bM_\bA\|_\op^2=1+\|\bA\|_\op^2.
\]
We state the dual result for a generic quadratic loss before specializing to regression.

\begin{lemma}[Inner DRO strong duality]
\label{lem:inner-DRO-strong-duality}
Let $P\in\cP_2(\R^d)$, $\eps > 0$, and
$\bM\in\R^{k\times d}$ for $1\leq k\leq d$. Then
\[
    \sup_{Q\in\cB_\eps(P)}\E_Q\|\bM Z\|^2
    =
    \inf_{\eta>\|\bM\|_\op^2}
    \eta\left(
        \eps^2+
        \tr\left(
            (\eta\bI_k-\bM\bM^\top)^{-1}
            \bM\bSigma_P\bM^\top
        \right)
    \right).
\]
For regression, taking $k=m$ and $\bM=\bM_\bA$ gives
\[
    \sup_{Q\in\cB_\eps(P)}L(\bA,Q)
    =
    \inf_{\eta>1+\|\bA\|_\op^2}
    \eta\left(
        \eps^2+
        \tr\left(
            ((\eta-1)\bI_m-\bA\bA^\top)^{-1}
            \bM_\bA\bSigma_P\bM_\bA^\top
        \right)
    \right).
\]
\end{lemma}
\begin{proof}
By \cref{lem:W2-Bures-equivalence}, the second moments
of laws in $\cB_\eps(P)$ are precisely the matrices
$\bW\bW^\top$, where $\bW\in\R^{d\times d}$ satisfies
$\|\bW-\bSigma_P^{1/2}\|_\Frob^2\leq\eps^2$.
Since $\E_Q\|\bM Z\|^2
    =\tr(\bM\bSigma_Q\bM^\top)$, we obtain
\begin{align*}
    \sup_{Q\in\cB_\eps(P)}\E_Q\|\bM Z\|^2
    &=
    \sup_{\substack{
        \bW\in\R^{d\times d}\\
        \|\bW-\bSigma_P^{1/2}\|_\Frob^2\leq\eps^2
    }}
    \|\bM\bW\|_\Frob^2\\
    &=
    \inf_{\eta>\|\bM\|_\op^2}
    \eta\left(
        \eps^2+
        \tr\left(
            (\eta\bI_k-\bM\bM^\top)^{-1}
            \bM\bSigma_P\bM^\top
        \right)
    \right),
\end{align*}
where the second equality applies
\cref{lem:quadratic-ball-duality}
with $r=d$, $\tilde{\bW}=\bSigma_P^{1/2}$,
and $t=\eps^2$.

Finally, the regression specialization follows by taking
$\bM=\bM_\bA$ and using
$\bM_\bA\bM_\bA^\top=\bA\bA^\top+\bI_m$
together with $L(\bA,Q)=\E_Q\|\bM_\bA Z\|^2$.
\end{proof}

We note that the infimum over the open dual domain need not be attained;
its value may be approached at the left endpoint. We next specialize to the case where $\bA \bA^\top$ is diagonal.

\begin{corollary}[Diagonal case]
\label{cor:DRO-strong-duality-diagonal}
Let $P\in\cP_2(\R^d)$, $\eps > 0$, and $\bA\in\cA$ satisfy
\[
    \bA\bA^\top=\diag(a_1^2,\ldots,a_m^2).
\]
For $i=1,\ldots,m$, define the $i$th coordinate residual second moment
\[
    r_i^2
    \defeq
    \E_P\bigl[(e_i^\top(\bA X-Y))^2\bigr]
    =
    e_i^\top\bM_\bA\bSigma_P\bM_\bA^\top e_i.
\]
We then have
\[
    \sup_{Q\in\cB_\eps(P)}L(\bA,Q)
    =
    \inf_{\eta>1+\max_i a_i^2}
    \eta\left(
        \eps^2+
        \sum_{i=1}^m
        \frac{r_i^2}{\eta-1-a_i^2}
    \right).
\]
\end{corollary}

\begin{proof}
In the regression formula of
\cref{lem:inner-DRO-strong-duality},
\[
    ((\eta-1)\bI_m-\bA\bA^\top)^{-1}
    =
    \diag\left(
        \frac{1}{\eta-1-a_1^2},\ldots,
        \frac{1}{\eta-1-a_m^2}
    \right).
\]
Taking the trace of its product with
$\bM_\bA\bSigma_P\bM_\bA^\top$ gives the stated sum.
\end{proof}

\subsection{Averaging over Symmetries}

All symmetry reductions used below preserve the feature and
response subspaces. We therefore formulate the averaging
principle directly for block-diagonal orthogonal transformations. Recall that for a measurable map $T$, $T_\sharp P$ denotes the pushforward law of
$T(Z)$ when $Z\sim P$.

\begin{lemma}[Averaging over symmetries]
\label{lem:DRO-symmetry}
Let $U$ be a compact subgroup of
\[
    \left\{
        \diag(\bU_X,\bU_Y):
        \bU_X\in\OO^p,\ \bU_Y\in\OO^m
    \right\}.
\]
Suppose $P,\tilde{P} \in\cP_2(\R^d)$ satisfy
$\bU_\sharp P=P$ and $\bU_\sharp \tilde{P}=\tilde{P}$
for every $\bU\in U$.
Let $\eps > 0$, and fix any minimizer $\hat{\bA} \in \cA$ for the
$\DRO_\eps[\tilde{P}]$, $\MRO_\eps[\tilde{P}]$, or
$\SNR_\eps[\tilde{P}]$ optimization problem.

Let $\mu$ be normalized Haar measure on $U$ and define
\[
    \bar{\bA}
    \defeq
    \int_U\bU_Y^\top\hat{\bA}\bU_X\,\dd\mu(\bU).
\]
Then $\bar{\bA}$ is a minimizer of the same optimization problem,
is invariant in the sense that
\[
    \bU_Y^\top\bar{\bA}\bU_X=\bar{\bA}
    \qquad\forall\,\bU\in U,
\]
and satisfies $R(\bar{\bA},P)\leq R(\hat{\bA},P)$. Consequently, restricting any of these optimization problems to invariant coefficient matrices preserves its optimal value.
Moreover, to lower bound the target regret of every optimizer,
it suffices to lower bound the target regret of every
invariant optimizer.
\end{lemma}

\begin{proof}
For $\bU=\diag(\bU_X,\bU_Y)\in U$, define the linear
coefficient transformation
\[
    T_\bU(\bA)\defeq\bU_Y^\top\bA\bU_X.
\]
Orthogonality gives
\begin{align*}
    L(\bA,\bU_\sharp Q)
    &=
    \E_Q\|\bA\bU_X X-\bU_Y Y\|^2\\
    &=
    \E_Q\|\bU_Y^\top\bA\bU_X X-Y\|^2\\
    &=
    L(T_\bU(\bA),Q).
\end{align*}
The map $T_\bU$ is a bijection from $\cA$ to itself, so
\[
    \inf_{\bB\in\cA}L(\bB,\bU_\sharp Q)
    =
    \inf_{\bB\in\cA}L(\bB,Q).
\]
Subtracting this identity from the preceding risk identity yields
\[
    R(\bA,\bU_\sharp Q)=R(T_\bU(\bA),Q).
\]
Also, left and right multiplication by orthogonal matrices gives
$\|T_\bU(\bA)\|_\op=\|\bA\|_\op$.

The source ambiguity set is invariant under each $\bU\in U$. Indeed,
\[
    \Wtwo(\bU_\sharp Q,\tilde P)
    =
    \Wtwo(\bU_\sharp Q,\bU_\sharp\tilde P)
    =
    \Wtwo(Q,\tilde P).
\]
Applying this also to $\bU^{-1}$ proves
$\bU_\sharp\cB_\eps(\tilde P)=\cB_\eps(\tilde P)$.
It follows that each of the three objectives
\[
    \sup_{Q\in\cB_\eps(\tilde P)}L(\bA,Q),
    \qquad
    \sup_{Q\in\cB_\eps(\tilde P)}R(\bA,Q),
    \qquad
    L(\bA,\tilde P)+\eps^2\|\bA\|_\op^2
\]
is invariant under $T_\bU$.
Each objective is also convex in $\bA$.

The integral defining $\bar{\bA}$ is well-defined because its
integrand is continuous on the compact group $U$.
For any fixed $\bV=\diag(\bV_X,\bV_Y)\in U$,
\begin{align*}
    \bV_Y^\top\bar{\bA}\bV_X
    &=
    \int_U
        (\bU_Y\bV_Y)^\top\hat{\bA}
        (\bU_X\bV_X)\,\dd\mu(\bU)\\
    &=
    \bar{\bA},
\end{align*}
where the last equality uses right invariance of Haar measure
under $\bU\mapsto\bU\bV$.
Thus $\bar{\bA}$ is invariant.

Let $J$ denote whichever of the three objectives is minimized
by $\hat{\bA}$.
By convexity, invariance, and Jensen's inequality,
\[
    J(\bar{\bA})
    \leq
    \int_U J(T_\bU(\hat{\bA}))\,\dd\mu(\bU)
    =
    J(\hat{\bA}).
\]
Since $\hat{\bA}$ is a minimizer, $\bar{\bA}$ is also a minimizer.

Finally, invariance of $P$ gives
$R(T_\bU(\hat{\bA}),P)=R(\hat{\bA},P)$.
The regret is convex in its coefficient argument, so another
application of Jensen's inequality yields
\[
    R(\bar{\bA},P)
    \leq
    \int_U R(T_\bU(\hat{\bA}),P)\,\dd\mu(\bU)
    =
    R(\hat{\bA},P).
\]
This proves all claims.
\end{proof}

\section{Remaining Proofs for \cref{sec:intro}}
\label{prf:intro}

\subsection{Existence of Minimizers (Footnote~\ref{footnote:minimizer-existence})}
\label{app:minimizer-existence}

We treat the estimators individually.

\textbf{ERM.} Solving the normal equation $\nabla_\bA L(\bA,P) = \bZero$ reveals that $\bSigma_{YX}^P(\bSigma_X^P)^{\dagger}$ is always a minimizer of $L(\cdot,P)$.

\textbf{DRO.} Since $L(\cdot,Q)$ is continuous and convex, the DRO objective $F(\bA) = \sup_{Q \in \cB_\eps(P)} L(\bA,Q)$ is convex and lower semicontinuous as the supremum of continuous convex functions. It thus suffices to show that $F$ is coercive. To prove this, consider the feasible measure $Q = P * \cN(0,\eps^2 \diag(\bI_p,\bZero_{m \times m})/p)$, which certifies that
\begin{equation*}
    F(\bA) \geq L(\bA,Q) = L(\bA,P) + \frac{\eps^2}{p} \E_{W \sim \cN(0,\bI_p)} \|\bA W\|^2 \geq \frac{\eps^2}{p} \|\bA\|_\Frob^2.
\end{equation*}
Thus $\|\bA\|_\Frob \to \infty$ implies $F(\bA) \to \infty$, and so $F$ is coercive, as desired.

\textbf{MRO.} Similar to the above, continuity and convexity of $R(\cdot,Q)$ imply that the MRO objective $G(\bA) = \sup_{Q \in \cB_\eps(P)} R(\bA,Q)$ is convex and lower semicontinuous. It thus remains to show that $G$ is coercive. To prove this, consider the same feasible measure $Q$ as for DRO. Since $Q$ is non-degenerate, we have
\begin{align*}
    G(\bA) \geq R(\bA,Q) = \bigl\|(\bA - \bA_Q)(\bSigma_X^Q)^{1/2}\bigr\|_\Frob^2 \geq \frac{\eps^2}{p} \|\bA - \bA_Q\|_\Frob^2.
\end{align*}

Thus $\|\bA\|_\Frob \to \infty$ implies $G(\bA) \to \infty$, and so $G$ is coercive, as desired.

\textbf{SNR.} The regularized objective $H(\bA) = L(\bA,P) + \eps^2 \|\bA\|_\op^2$ is clearly convex and lower semicontinuous. Moreover, since the risk is non-negative, it is also coercive.

\section{Remaining Proofs for \cref{sec:condition-number}}
\label{app:condition-number}

\subsection{Condition Number Formula (Proof of \cref{thm:condition-number-formula})}
\label{prf:condition-number-formula}

We formulate the proof in Bures space, adopting the conventions of Appendix~\ref{app:Bures-geometry}. Initially, we find it simpler to analyze the alternative condition number $\tilde{\kappa} : \cC^* \to [0,\infty]$, defined for $\bSigma \in \cC^*$ by
\begin{equation}
\label{eq:alt-condition-number-def}
    \tilde{\kappa}(\bSigma)
    \defeq
    \limsup_{\eps\downarrow0}
    \eps^{-2}
    \sup_{\bSigma'\in\cB_\eps(\bSigma)}
    R(\bA_{\bSigma'},\bSigma),
\end{equation}
which we will later show coincides with $\kappa$.

Our first step is to differentiate the parameter map $g(\bSigma) \defeq \bSigma_{YX} \bSigma_X^{-1}$. To avoid boundary issues, we extend this map to the open domain $\cC \defeq \{ \bSigma \in \Sym^d: \bSigma_X \succ 0 \}$.

\begin{lemma}[Parameter map derivative]
\label{lem:parameter-map-derivative}
The map $g:\cC \to \cA$ is continuously differentiable, with
\begin{equation*}
    D g(\bSigma)[\bH] = (\bH_{YX} - g(\bSigma) \bH_X) \bSigma_X^{-1} \quad \forall\, \bH \in \Sym^d.
\end{equation*}
\end{lemma}
\begin{proof}
The maps $\bSigma\mapsto\bSigma_{YX}$ and
$\bSigma\mapsto\bSigma_X$ are linear, while matrix inversion is smooth
on the positive definite cone. Hence $g$ is continuously
differentiable on $\cC$. By the product rule and the standard derivative
of the inverse map,
\begin{align*}
    Dg(\bSigma)[\bH]
    &=
    \bH_{YX}\bSigma_X^{-1}
    -\bSigma_{YX}\bSigma_X^{-1}
        \bH_X\bSigma_X^{-1}\\
    &=
    \bigl(\bH_{YX}-g(\bSigma)\bH_X\bigr)\bSigma_X^{-1},
\end{align*}
as claimed.
\end{proof}

We now characterize $\tilde{\kappa}$ in terms of this derivative. We prove a slightly strengthened, uniform claim.

\begin{lemma}[Derivative expression for $\tilde{\kappa}$]
\label{lem:alt-condition-number-formula}
Fixing $\bSigma \in \cC^*$ with square root $\bS \defeq \bSigma^{1/2}$, we have
\begin{equation}
\label{eq:alt-condition-number-formula}
    \tilde{\kappa}(\bSigma) = \sup_{\bDelta \in \R^{d \times d}: \|\bDelta\|_\Frob \leq 1} \bigl\|D g(\bSigma)[\bS \bDelta^\top + \bDelta \bS] \bSigma_X^{1/2}\bigr\|_{\Frob}^2.
\end{equation}
Moreover, the expansion
\begin{equation}
\label{eq:alt-condition-number-taylor-expansion}
     \sup_{\bSigma' \in \cB_\eps(\bSigma)} R(g(\bSigma'),\bSigma) = \tilde{\kappa}(\bSigma) \eps^2 + o(\eps^2) \quad \eps \to 0
\end{equation}
holds uniformly as $\bSigma$ varies in any compact subset of $\cC^*$.
\end{lemma}

\begin{proof}
Write $\bSigma(\bDelta) \defeq (\bS + \bDelta)(\bS + \bDelta)^\top = \bSigma + \bS \bDelta^\top + \bDelta \bS + \bDelta \bDelta^\top$, and let $K$ be any compact subset of $\cC^*$. By \cref{lem:parameter-map-derivative}, $g$ is continuously
differentiable on $\cC \supset K$, and so
\begin{equation*}
    g(\bSigma+\bH)-g(\bSigma)
    =
    Dg(\bSigma)[\bH]+o(\|\bH\|_\Frob)
\end{equation*}
as $\bH\to\bZero$, uniformly over $\bSigma\in K$. Further,
\begin{equation*}
    \|Dg(\bSigma)\|_\op \defeq \sup_{\|\bH\|_\Frob \leq 1} \|Dg(\bSigma)[\bH]\|_\Frob
\end{equation*}
is bounded for $\bSigma \in K$. Further
\begin{equation*}
    \|\bSigma(\bDelta) - \bSigma\|_\Frob \leq 2\|\bS\|_\Frob \|\bDelta\|_\Frob + \|\bDelta\|_\Frob^2 = 2\sqrt{\tr(\bSigma)} \|\bDelta\|_\Frob + \|\bDelta\|_\Frob^2 = O(\|\bDelta\|_\Frob)
\end{equation*}
as $\bDelta \to \bZero$, uniformly for $\bSigma \in K$. Combining gives
\begin{align*}
    g(\bSigma(\bDelta)) - g(\bSigma) &= Dg(\bSigma)[\bS \bDelta^\top + \bDelta \bS + \bDelta \bDelta^\top] + o(\|\bDelta\|_\Frob)\\
    &= Dg(\bSigma)[\bS \bDelta^\top + \bDelta \bS] + (1 + \|Dg(\bSigma)\|_\op)o(\|\bDelta\|_\Frob)\\
    &= Dg(\bSigma)[\bS \bDelta^\top + \bDelta \bS] + o(\|\bDelta\|_\Frob)
\end{align*}
as $\bDelta \to \bZero$, uniformly over $\bSigma \in K$. Hence,
\begin{align*}
    \left\|(g(\bSigma(\bDelta)) - g(\bSigma))\bSigma_X^{1/2}\right\|_\Frob^2 &= \left\|\bigl(Dg(\bSigma)[\bS \bDelta^\top + \bDelta \bS ] + o(\|\bDelta\|_\Frob)\bigr)\bSigma_X^{1/2}\right\|_\Frob^2\\
    &= \left\|Dg(\bSigma)[\bS \bDelta^\top + \bDelta \bS ]\bSigma_X^{1/2}\right\|_\Frob^2 + E(\bSigma,\bDelta)
\end{align*}
where, by Cauchy--Schwarz and submultiplicativity of the Frobenius norm
\begin{align*}
    |E(\bSigma,\bDelta)| &\leq 2\|D g(\bSigma)\|_\op \|\bS \bDelta^\top + \bDelta \bS\|_\Frob o(\|\bDelta\|_\Frob) \|\bSigma_X\|_\Frob + o(\|\bDelta\|_\Frob^2) \|\bSigma_X\|_\Frob\\
    &\leq 4\|D g(\bSigma)\|_\op \sqrt{\tr(\bSigma)}\|\bDelta\|_\Frob o(\|\bDelta\|_\Frob) \|\bSigma_X\|_\Frob + o(\|\bDelta\|_\Frob^2) \|\bSigma_X\|_\Frob\\
    &= o(\|\bDelta\|_\Frob^2)
\end{align*}
as $\bDelta \to \bZero$, uniformly over $\bSigma \in K$. Combining this with \cref{lem:Bures-Procrustes} gives
\begin{align*}
    \sup_{\bSigma' \in \cB_\eps(\bSigma)} R(g(\bSigma'),\bSigma) &= \sup_{\bDelta \in \R^{d \times d}: \|\bDelta\|_\Frob \leq \eps} R\bigl(g(\bSigma(\bDelta)), \bSigma\bigr)\\
    &= \sup_{\bDelta \in \R^{d \times d}: \|\bDelta\|_\Frob \leq \eps} \left\|\left(g(\bSigma(\bDelta)) - g(\bSigma)\right) \bSigma_X^{1/2}\right\|_\Frob^2\\
    &= \sup_{\bDelta \in \R^{d \times d}: \|\bDelta\|_\Frob \leq \eps} \left\|\left(Dg(\bSigma)[\bS \bDelta^\top + \bDelta \bS ] + o(\|\bDelta\|_\Frob)\right) \bSigma_X^{1/2}\right\|_\Frob^2\\
    &= \sup_{\bDelta \in \R^{d \times d}: \|\bDelta\|_\Frob \leq 1} \left\|Dg(\bSigma)[\bS \bDelta^\top + \bDelta \bS ] \bSigma_X^{1/2}\right\|_\Frob^2 \,\eps^2 + o(\eps^2),
\end{align*}
as $\eps \to 0$, uniformly over $\bSigma \in K$. The lemma then follows by the definition of $\tilde{\kappa}$.
\end{proof}

Next, we substitute the derivative formula from \cref{lem:parameter-map-derivative} into \eqref{eq:alt-condition-number-formula} and simplify. We use the variational form of the Frobenius norm $\|\bX\|_\Frob = \sup_{\bY:\|\bY\|_\Frob \leq 1} \langle \bX,\bY \rangle_\Frob$ to streamline calculations.

\begin{lemma}[Closed form for $\tilde{\kappa}$]
\label{lem:alt-condition-number-closed-form}
For $\bSigma \in \cC^*$, we have
\begin{equation*}
    \tilde{\kappa}(\bSigma) = 1 + \lambda_{\max}\left(\bA_\bSigma \bA_\bSigma^\top + \lambda_{\min}(\bSigma_X)^{-1}\bSigma_E\right).
\end{equation*}
In particular, we have
\begin{equation*}
     1 + \max\left\{\|\bA_\bSigma\|_\op^2, \frac{\lambda_{\max}(\bSigma_E)}{\lambda_{\min}(\bSigma_X)}\right\}  \leq \tilde{\kappa}(\bSigma) \leq  1 + \|\bA_\bSigma\|_\op^2 + \frac{\lambda_{\max}(\bSigma_E)}{\lambda_{\min}(\bSigma_X)}.
\end{equation*}
\end{lemma}
\begin{proof}
Abbreviate $\bA \defeq g(\bSigma) = \bSigma_{YX}\bSigma_X^{-1}$ and $\bS \defeq \bSigma^{1/2}$. Combining \cref{lem:parameter-map-derivative,lem:alt-condition-number-formula} gives
\begin{align*}
    \tilde{\kappa}(\bSigma) &= \sup_{\bDelta \in \R^{d \times d}: \|\bDelta\|_\Frob \leq 1}\|D g(\bSigma)[\bS \bDelta^\top + \bDelta \bS] \bSigma_X^{1/2}\|_{\Frob}^2\\
    &= \sup_{\bDelta \in \R^{d \times d}: \|\bDelta\|_\Frob \leq 1}\bigl\|\bigl(\bS_Y \bDelta_X^\top + \bDelta_Y \bS_X^\top - \bA (\bS_X \bDelta_X^\top + \bDelta_X \bS_X^\top)\bigr) \bSigma_X^{-1/2}\bigr\|_{\Frob}^2\\
    &=  \sup_{\bDelta \in \R^{d \times d}: \|\bDelta\|_\Frob \leq 1} \left\|\left((\bDelta_Y - \bA\bDelta_X) \bS_X^\top + (\bS_Y - \bA \bS_X)\bDelta_X^\top\right)\bSigma_X^{-1/2}\right\|_\Frob^2.
\end{align*}

To simplify the objective, write $\bQ \defeq \bS_X^\top \bSigma_X^{-1/2} \in \R^{d \times p}$, noting that $\bQ^\top \bQ = \bI_p$ and that $\bQ \bQ^\top \in \R^{d \times d}$ is the orthogonal projection onto the row space of $\bS_X$. Further define the residual factor $\bE \defeq \bS_Y - \bA \bS_X \in \R^{m \times d}$, which by \cref{lem:regret-projection-formula} satisfies $\bE \bE^\top = \bSigma_E$ and  $\bE \bS_X^\top = \bZero$ (implying $\bE\bQ = \bZero$). Substituting these definitions into the equation above and rearranging, we obtain
\begin{align*}
    \tilde{\kappa}(\bSigma) &=
    \sup_{\|\bDelta\|_\Frob\leq1}
    \bigl\|
            (\bDelta_Y - \bA \bDelta_X)\bQ + \bE \bDelta_X^\top \bSigma_X^{-1/2}
    \bigr\|_\Frob^2.\\
    &=
    \sup_{\|\bDelta\|_\Frob\leq1}
    \Bigl\|
        \underbrace{
            \bE\bDelta_X^\top\bSigma_X^{-1/2}
            -\bA\bDelta_X\bQ
        }_{\text{feature perturbations}}
        +
        \underbrace{
            \bDelta_Y\bQ
        }_{\text{response perturbations}}
    \Bigr\|_\Frob^2.
\end{align*}
The first component depends on the perturbation $\bDelta_X$ to the feature factor $\bS_X$, while the second component depends on the perturbation $\bDelta_Y$ to the response factor $\bS_Y$.

To simplify further, we use the variational form of the Frobenius norm. For $\bB \in \R^{m \times p}$, define
\begin{align*}
    \phi(\bB;\bDelta) &\defeq\bigl\langle \bB,   {\bE\bDelta_X^\top\bSigma_X^{-1/2}
            -\bA\bDelta_X\bQ + \bDelta_Y \bQ }\bigr\rangle_\Frob.
\end{align*}
Since the Frobenius unit ball is symmetric, we have
\[
    \tilde{\kappa}(\bSigma)
    =
    \sup_{\|\bDelta\|_\Frob\leq1}
    \Biggl(\,\sup_{\|\bB\|_\Frob\leq1}
    \phi(\bB;\bDelta)\Biggr)^2 =
    \sup_{\|\bDelta\|_\Frob\leq1}
    \sup_{\|\bB\|_\Frob\leq1}
    \phi(\bB;\bDelta)^2.
\]
We now swap these suprema and maximize over $\bDelta$ for
fixed $\bB$.
Grouping terms according to the perturbed row block and using cyclicity of the trace gives
\begin{align*}
    \phi(\bB;\bDelta)
    &=
    \bigl\langle
        \bSigma_X^{-1/2}\bB^\top\bE
        -\bA^\top\bB\bQ^\top,\,
        \bDelta_X
    \bigr\rangle_\Frob
    +
    \left\langle
        \bB\bQ^\top,\bDelta_Y
    \right\rangle_\Frob.
\end{align*}
Since the Frobenius norm is self-dual,
\begin{align*}
    \sup_{\|\bDelta\|_\Frob\leq1}\phi(\bB;\bDelta)^2
    &=
    \left\|
        \begin{pmatrix}
            \bSigma_X^{-1/2}\bB^\top\bE
            -\bA^\top\bB\bQ^\top\\
            \bB\bQ^\top
        \end{pmatrix}
    \right\|_\Frob^2\\
    &=
    \bigl\|
        \bSigma_X^{-1/2}\bB^\top\bE
        -\bA^\top\bB\bQ^\top
    \bigr\|_\Frob^2
    +\|\bB\bQ^\top\|_\Frob^2\\
    &=
    \underbrace{
        \|\bSigma_X^{-1/2}\bB^\top\bE\|_\Frob^2
        +\|\bA^\top\bB\|_\Frob^2
    }_{\text{feature contribution}}
    +
    \underbrace{
        \|\bB\|_\Frob^2.
    }_{\text{response contribution}}
\end{align*}
The last equality uses $\bE\bQ=\bZero$ to eliminate the
cross term within the feature contribution and applies
$\bQ^\top\bQ=\bI_p$ to simplify the other norms.
For fixed $\bB$, restricting to $\bDelta_Y=\bZero$
leaves exactly the first bracket, while restricting to
$\bDelta_X=\bZero$ leaves the final term.

By homogeneity, the outer supremum may be restricted to
$\|\bB\|_\Frob=1$.
The response contribution is then $1$ for every $\bB$.
Using $\bE\bE^\top=\bSigma_E$ and
\cref{lem:kronecker-fact}, we obtain
\begin{align*}
    \tilde{\kappa}(\bSigma)
    &=
    1+\sup_{\|\bB\|_\Frob=1}
    \left\{
        \tr(\bSigma_X^{-1}\bB^\top\bSigma_E\bB)
        +\tr(\bA\bA^\top\bB\bB^\top)
    \right\}\\
    &=
    1+\sup_{\|\bB\|_\Frob=1}
    \vec(\bB)^\top
    \left(
        \bSigma_X^{-1}\otimes\bSigma_E
        +\bI_p\otimes\bA\bA^\top
    \right)
    \vec(\bB)\\
    &=
    1+\left\|
        \bSigma_X^{-1}\otimes\bSigma_E
        +\bI_p\otimes\bA\bA^\top
    \right\|_\op.
\end{align*}
The remaining operator norm measures worst-case sensitivity to feature perturbations alone.

To compute it, eigendecompose
$\bSigma_X=\bU\diag(s_1,\ldots,s_p)\bU^\top$,
where $\bU\in\OO^p$ and $0<s_1\leq\cdots\leq s_p$.
Conjugating by $\bU\otimes\bI_m$ gives a block-diagonal matrix with blocks $s_i^{-1}\bSigma_E+\bA\bA^\top$. 
Since $s_i^{-1}\leq s_1^{-1}$ and
$\bSigma_E\succeq\bZero$, the first block dominates all
others in Loewner order.
Consequently,
\[
   \tilde{\kappa}(\bSigma)
    =
    \underbrace{1}_{\text{response contribution}}
    +
    \underbrace{
        \lambda_{\max}\left(
            \bA\bA^\top+s_1^{-1}\bSigma_E
        \right)
    }_{\text{feature contribution}}.
\]
This proves the formula since $s_1=\lambda_{\min}(\bSigma_X)$.

Finally, positive semidefiniteness of the two matrices inside
$\lambda_{\max}$ and subadditivity of $\lambda_{\max}$ give
\[
    \max\left\{
        \|\bA\|_\op^2,\,
        s_1^{-1}\lambda_{\max}(\bSigma_E)
    \right\}
    \leq\tilde{\kappa}(\bSigma)-1
    \leq
    \|\bA\|_\op^2+s_1^{-1}\lambda_{\max}(\bSigma_E),
\]
establishing the remaining bounds.
\end{proof}

We next show that $\kappa$ and $\tilde{\kappa}$ coincide. The upper bound is relatively direct in light of \eqref{eq:alt-condition-number-taylor-expansion}.

\begin{lemma}[Condition number comparison, upper bound]
\label{lem:condition-number-comparison-UB}
For all $\bSigma \in \cC^*$, $\kappa(\bSigma) \leq \tilde{\kappa}(\bSigma)$.
\end{lemma}
\begin{proof}

We upper bound $\kappa(\bSigma)$ in \eqref{eq:condition-number-cov-def} by considering the feasible parameter $\bA = \bA_\bSigma = g(\bSigma)$. This gives
\begin{align*}
    \kappa(\bSigma) &\leq \limsup_{\eps \downarrow 0} \eps^{-2} \sup_{\bSigma' \in \cB_\eps(\bSigma)} R(g(\bSigma),\bSigma')\\
    &\leq \limsup_{\eps \downarrow 0} \eps^{-2} \sup_{\bSigma' \in \cB_\eps(\bSigma)} \sup_{\bSigma'' \in \cB_\eps(\bSigma')} R(g(\bSigma''),\bSigma').
\end{align*}
Noting that the convergence in \eqref{eq:alt-condition-number-taylor-expansion} is uniform over $\bSigma' \in \cB_\eps(\bSigma)$, that expansion implies
\begin{align*}
    \kappa(\bSigma) &\leq \limsup_{\eps \downarrow 0} \eps^{-2} \left( \sup_{\bSigma' \in \cB_\eps(\bSigma)} \tilde{\kappa}(\bSigma') \eps^2 + o(\eps^2)\right)\\
    &= \limsup_{\eps \downarrow 0} \sup_{\bSigma' \in \cB_\eps(\bSigma)} \tilde{\kappa}(\bSigma')\\
    &= \tilde{\kappa}(\bSigma).
\end{align*}
The final equality uses continuity of $\tilde{\kappa}$, which is evident from our closed form in \cref{lem:alt-condition-number-closed-form}.
\end{proof}

We prove a slightly stronger result in the other direction (using $\liminf$ instead of $\limsup$ below).

\begin{lemma}[Condition number comparison, lower bound]
\label{lem:condition-number-comparison-LB}
For all $\bSigma \in \cC^*$, we have
\begin{align*}
    \kappa(\bSigma) \geq J_\bSigma \defeq \liminf_{\eps \downarrow 0} \eps^{-2} \inf_{\bA \in \cA} \sup_{\bSigma' \in \cB_\eps(\bSigma)} R(\bA,\bSigma') \geq \tilde{\kappa}(\bSigma).
\end{align*}
\end{lemma}
\begin{proof}
The first inequality is immediate from \eqref{eq:condition-number-cov-def}, since $\limsup\geq\liminf$.

For the second inequality, fix $\bDelta_\star \in \R^{d \times d}$ with $\|\bDelta_\star\|_\Frob = 1$ achieving the supremum in formula \eqref{eq:alt-condition-number-formula} for $\tilde{\kappa}(\bSigma)$, noting that a maximizer exists by compactness of the Frobenius ball and continuity of the objective. Now, for each $t \in \R$, define $\bSigma^t \defeq (\bSigma^{1/2} + t\bDelta_\star)(\bSigma^{1/2} +t \bDelta_\star)^\top$, which satisfies $\sB(\bSigma^t,\bSigma) \leq |t|$ by \cref{lem:Bures-Procrustes}. In particular, $\bSigma^{\pm\eps}\in\cB_\eps(\bSigma)$. We thus have
\begin{align}
    J_\bSigma &\defeq \liminf_{\eps \downarrow 0} \eps^{-2} \inf_{\bA \in \cA} \sup_{\bSigma' \in \cB_\eps(\bSigma)} R(\bA,\bSigma')\notag\\
    &\geq \liminf_{\eps \downarrow 0} \eps^{-2} \inf_{\bA \in \cA} \max\bigl\{ R(\bA,\bSigma^{+\eps}), R(\bA,\bSigma^{-\eps}) \bigr\}\notag\\
    &= \liminf_{\eps \downarrow 0} \eps^{-2} \inf_{\bA \in \cA}\max\left\{\bigl\|(\bA - \bA_{+\eps})(\bSigma^{+\eps}_X)^{1/2}\bigr\|_\Frob^2, \bigl\|(\bA - \bA_{-\eps})(\bSigma^{-\eps}_X)^{1/2}\bigr\|_\Frob^2\right\},
\end{align}
where $\bA_t \defeq g(\bSigma^t)$. Since $\bSigma_X^{t} \to \bSigma_X$ as $t \to 0$, we have for fixed $\lambda \in (0,1)$ that $\bSigma_X^t \succ \lambda \bSigma_X$ for sufficiently small $t$. Consequently, we have
\begin{align*}
    J_\bSigma &\geq \sup_{\lambda \in (0,1)} \lambda \liminf_{\eps \downarrow 0} \eps^{-2} \inf_{\bA \in \cA}\max\left\{\bigl\|(\bA - \bA_{+\eps})\bSigma_X^{1/2}\bigr\|_\Frob^2, \bigl\|(\bA - \bA_{-\eps})\bSigma_X^{1/2}\bigr\|_\Frob^2\right\}\\
    &= \liminf_{\eps \downarrow 0} \eps^{-2} \inf_{\bA \in \cA}\max\left\{\bigl\|(\bA - \bA_{+\eps})\bSigma_X^{1/2}\bigr\|_\Frob^2, \bigl\|(\bA - \bA_{-\eps})\bSigma_X^{1/2}\bigr\|_\Frob^2\right\}.
\end{align*}
By \cref{lem:elementary-scalar-ineqs}, the infimum is attained by the midpoint $\bar{\bA}_\eps = (\bA_{+\eps} + \bA_{-\eps})/2$. This gives,
\begin{align*}
    J_\bSigma &\geq \frac{1}{4} \liminf_{\eps \downarrow 0} \eps^{-2} \bigl\|(\bA_{+\eps} - \bA_{-\eps})\bSigma_X^{1/2}\bigr\|_\Frob^2
\end{align*}
By construction, we have
\begin{align*}
    \bSigma^{+\eps} - \bSigma^{-\eps} = 2\eps \bigl(\bSigma^{1/2} \bDelta_\star^\top + \bDelta_\star \bSigma^{1/2}\bigr),
\end{align*}
which, since $g$ is continuously differentiable, gives
\begin{align*}
    \bA_{+\eps} - \bA_{-\eps} &= g(\bSigma^{+\eps}) - g(\bSigma^{-\eps})\\
    &= Dg(\bSigma)[\bSigma^{+\eps} - \bSigma] - Dg(\bSigma)[\bSigma^{-\eps} - \bSigma] + o(\eps)\\
    &= Dg(\bSigma)[\bSigma^{+\eps} - \bSigma^{-\eps}] + o(\eps)\\
    &= 2\eps Dg(\bSigma)\bigl[\bSigma^{1/2} \bDelta_\star^\top + \bDelta_\star \bSigma^{1/2}\bigr] + o(\eps)
\end{align*}
as $\eps \to 0$. Substituting this expansion into the preceding lower bound for $J_\bSigma$ and using optimality of $\bDelta_\star$, we obtain
\begin{align*}
    J_\bSigma &\geq \bigl\|Dg(\bSigma)\bigl[\bSigma^{1/2} \bDelta_\star^\top + \bDelta_\star \bSigma^{1/2}\bigr]\bSigma_X^{1/2}\bigr\|_\Frob^2\\
    &=  \sup_{\bDelta \in \R^{d \times d}: \|\bDelta\|_\Frob \leq 1}\|D g(\bSigma)[\bSigma^{1/2}\bDelta^\top + \bDelta \bSigma^{1/2}] \bSigma_X^{1/2}\|_{\Frob}^2\\
    &= \tilde{\kappa}(\bSigma),
\end{align*}
as desired.
\end{proof}

Combining \cref{lem:condition-number-comparison-UB} and \cref{lem:condition-number-comparison-LB} with \eqref{eq:condition-number-cov-def} gives
\begin{align*}
    \kappa(\bSigma) = \tilde{\kappa}(\bSigma) = \lim_{\eps \downarrow 0} \eps^{-2} \inf_{\bA \in \cA} \sup_{\bSigma' \in \cB_\eps(\bSigma)} R(\bA,\bSigma') = \lim_{\eps \downarrow 0} \eps^{-2} R^\star_\eps(\bSigma)
\end{align*}
for $\bSigma \in \cC^*$. The theorem then follows by the \cref{lem:alt-condition-number-closed-form} formula and inequalities.\qed

\section{Remaining Proofs for \cref{sec:non-asymptotic}}
\label{prf:non-asymptotic}

\subsection{Regret Transfer (Proof of \cref{lem:regret-stability})}
\label{prf:regret-stability}

We first establish a refined bound by comparing projected residual factors. The argument separates the direct effect of the perturbation
from the change in the feature row space, and it allows the perturbed
feature second-moment matrix to be singular.

\begin{lemma}[Regret stability, refined]
\label{lem:regret-stability-refined}
For all $\eps\geq0$, $\bA\in\cA$, $P\in\cP^*$, and
$Q\in\cB_\eps(P)$, we have
\begin{align*}
    \left|
        \sqrt{R(\bA,P)}-\sqrt{R(\bA,Q)}
    \right| \leq
    \left(
        \max\left\{
            \|\bA-\bA_P\|_\op,\,
            \sqrt{1+\|\bA\|_\op^2}\,
        \right\}
        +
        \sqrt{
            \frac{\lambda_{\max}(\bSigma_E^P)}
                 {\lambda_{\min}(\bSigma_X^P)}
        }
    \,\right)\eps.
\end{align*}
\end{lemma}

\begin{proof}
We adopt the Bures-space conventions and block decompositions of
\cref{app:Bures-geometry}. Set
$\bSigma\defeq\bSigma_P$ and
$\bar{\bSigma}\defeq\bSigma_Q$.
By \cref{lem:W2-Bures-equivalence,lem:Bures-Procrustes}, there are square-root factors $\bS,\bar{\bS}\in\R^{d\times d}$ such that
\[
    \bSigma=\bS\bS^\top,
    \qquad
    \bar{\bSigma}=\bar{\bS}\bar{\bS}^\top,
    \qquad
    \bDelta\defeq\bar{\bS}-\bS,
    \qquad
    \|\bDelta\|_\Frob\leq\eps.
\]
Let $\bPi_X$ and $\bar{\bPi}_X$ denote the orthogonal projections
onto the row spaces of $\bS_X$ and $\bar{\bS}_X$, respectively, and define the residual factors
\[
    \bE_\bA\defeq\bS_Y-\bA\bS_X,
    \qquad
    \bar{\bE}_\bA\defeq\bar{\bS}_Y-\bA\bar{\bS}_X,
    \qquad
    \bE\defeq\bE_{\bA_\bSigma}.
\]
By \cref{lem:regret-projection-formula}, we have
\[
    \sqrt{R(\bA,\bSigma)}
    =\|\bE_\bA\bPi_X\|_\Frob,
    \qquad
    \sqrt{R(\bA,\bar{\bSigma})}
    =\|\bar{\bE}_\bA\bar{\bPi}_X\|_\Frob,
\]
and the optimal residual factor $\bE$ satisfies
$\bE\bPi_X=\bZero$ and $\bE\bE^\top=\bSigma_E$.

We compare these projected residuals directly. Since
$\bS_X\bPi_X=\bS_X$ and
$\bar{\bS}_X\bar{\bPi}_X=\bar{\bS}_X$, we have
\begin{align*}
    \bE_\bA\bPi_X-\bar{\bE}_\bA\bar{\bPi}_X &= (\bS_Y - \bA \bS_X) \bPi_X - (\bar{\bS}_Y - \bA \bar{\bS}_X) \bar{\bPi}_X\\
    &= \bS_Y(\bPi_X-\bar{\bPi}_X) -\bDelta_Y\bar{\bPi}_X
    +\bA\bDelta_X.
\end{align*}
Since our desired bound does not depend on $\bS_Y$ or $\bSigma_Y$ directly, we substitute $\bS_Y = \bA_\bSigma\bS_X+\bE$ for the first term to obtain
\begin{align*}
    \bS_Y(\bPi_X-\bar{\bPi}_X) &= (\bA_\bSigma\bS_X+\bE)(\bPi_X-\bar{\bPi}_X)\\
    &= \bA_\bSigma\bS_X -\bA_\bSigma\bS_X \bar{\bPi}_X - \bE \bar{\bPi}_X \tag{$\bS_X \bPi_X = \bS_X$, $\bE \bPi_X = \bZero$}\\
    &=\bA_\bSigma\bS_X (\bI_d - \bar{\bPi}_X ) - \bE \bar{\bPi}_X\\
    &= \bA_\bSigma\bar{\bS}_X (\bI_d - \bar{\bPi}_X ) - \bA_\bSigma\bDelta_X (\bI_d - \bar{\bPi}_X ) - \bE \bar{\bPi}_X\\
     &= - \bA_\bSigma\bDelta_X (\bI_d - \bar{\bPi}_X ) - \bE \bar{\bPi}_X,
\end{align*}
where the last equality uses $\bar{\bS}_X(\bI_d - \bar{\bPi}_X) = \bZero$. Substituting this into the previous equation gives
\begin{align*}
   \bE_\bA\bPi_X-\bar{\bE}_\bA\bar{\bPi}_X &= - \bA_\bSigma\bDelta_X (\bI_d - \bar{\bPi}_X ) - \bE \bar{\bPi}_X -\bDelta_Y\bar{\bPi}_X
    +\bA\bDelta_X\\
    &=
    (\bA-\bA_\bSigma)\bDelta_X(\bI_d-\bar{\bPi}_X)
    +(\bA\bDelta_X-\bDelta_Y)\bar{\bPi}_X
    -\bE\bar{\bPi}_X.
\end{align*}
Let $\bT$ denote the sum of the first two terms in the last line,
and write
\[
    c\defeq
    \max\left\{
        \|\bA-\bA_\bSigma\|_\op,\,
        \sqrt{1+\|\bA\|_\op^2}
    \right\}.
\]
Note that
$1+\|\bA\|_\op^2 = \lambda_{\max}(\bI_m + \bA \bA^\top) = \|\begin{pmatrix}\bA& -\bI_m\end{pmatrix}\|_\op^2$. Moreover, the two terms of $\bT$ have orthogonal right projections, so we obtain
\begin{align*}
    \|\bT\|_\Frob^2
    &=
    \|(\bA-\bA_\bSigma)
        \bDelta_X(\bI_d-\bar{\bPi}_X)\|_\Frob^2
    +
    \|(\bA\bDelta_X-\bDelta_Y)\bar{\bPi}_X\|_\Frob^2\\
    &\leq
    \|\bA-\bA_\bSigma\|_\op^2\|\bDelta_X(\bI_d-\bar{\bPi}_X)\|_\Frob^2
    +
    \|\begin{pmatrix} \bA & -\bI_m \end{pmatrix}\|_\op^2 \|\bDelta \bar{\bPi}_X\|_\Frob^2 \tag{\cref{lem:basic-matrix-norms}}\\
    &\leq
    c^2\left(
        \|\bDelta_X(\bI_d-\bar{\bPi}_X)\|_\Frob^2
        +\|\bDelta_X\bar{\bPi}_X\|_\Frob^2
        +\|\bDelta_Y\bar{\bPi}_X\|_\Frob^2
    \right)\\
    &\leq
    c^2\left(
        \|\bDelta_X\|_\Frob^2+\|\bDelta_Y\|_\Frob^2
    \right)
    =
    c^2\|\bDelta\|_\Frob^2.
\end{align*}
The last inequality uses the orthogonal decomposition of
$\bDelta_X$ and the contraction property of $\bar{\bPi}_X$.

It remains to bound $\bE\bar{\bPi}_X$, the component of the original
optimal residual in the perturbed feature row space.
Since $\bE\bPi_X=\bZero$, the row-space perturbation bound
(\cref{lem:sin-thm-variant}) gives
\begin{align*}
    \|\bE\bar{\bPi}_X\|_\Frob
    &=
    \|\bE(\bI_d-\bPi_X)\bar{\bPi}_X\|_\Frob\\
    &\leq
    \|\bE\|_\op
    \|(\bI_d-\bPi_X)\bar{\bPi}_X\|_\Frob\\
    &\leq
    \frac{\|\bE\|_\op}{\sigma_{\min}(\bS_X)}
    \|\bDelta_X\|_\Frob\\
    &\leq
    \sqrt{
        \frac{\lambda_{\max}(\bSigma_E)}
             {\lambda_{\min}(\bSigma_X)}
    }\,\eps.
\end{align*}
Here $\bS_X$ has full row rank because $\bSigma_X\succ\bZero$;
no rank assumption on $\bar{\bS}_X$ is needed.

Finally, the reverse triangle inequality yields
\begin{align*}
    \left|
        \sqrt{R(\bA,\bSigma)}
        -
        \sqrt{R(\bA,\bar{\bSigma})}
    \right|
    &\leq
    \|\bE_\bA\bPi_X-\bar{\bE}_\bA\bar{\bPi}_X\|_\Frob\\
    &\leq
    \|\bT\|_\Frob+\|\bE\bar{\bPi}_X\|_\Frob\\
    &\leq
    \left(
        c+
        \sqrt{
            \frac{\lambda_{\max}(\bSigma_E)}
                 {\lambda_{\min}(\bSigma_X)}
        }
    \right)\eps.
\end{align*}
Translating back to $P$ and $Q$ proves the claim.
\end{proof}

\begin{proof}[Proof of \cref{lem:regret-stability}]
By
\cref{thm:condition-number-formula},
\[
    \sqrt{\kappa_P} \geq \max\left\{\sqrt{1+\|\bA_P\|_\op^2},
    \sqrt{
        \frac{\lambda_{\max}(\bSigma_E^P)}
             {\lambda_{\min}(\bSigma_X^P)}
    }
   \right\}.
\]
Moreover, the triangle inequality for $\|\cdot\|_\op$ and subadditivity of the square root imply
\begin{align*}
    \max\left\{
        \|\bA-\bA_P\|_\op,\,
        \sqrt{1+\|\bA\|_\op^2}
    \right\}
    &\leq
    \|\bA\|_\op+\max\{\|\bA_P\|_\op,1\}\\
    &\leq
    \|\bA\|_\op+\sqrt{\kappa_P}.
\end{align*}
Applying \cref{lem:regret-stability-refined} with $\eps\defeq\Wtwo(P,Q)$ thus gives
\[
    \left|
        \sqrt{R(\bA,P)}-\sqrt{R(\bA,Q)}
    \right|
    \leq
    \bigl(\|\bA\|_\op+2\sqrt{\kappa_P}\bigr)\Wtwo(P,Q).
\]

For $\bA=\bA_P$, we have $R(\bA_P,P)=0$, and so \cref{lem:regret-stability-refined} gives
\begin{align*}
    R(\bA_P,Q)
    &\leq
    \left(
        \sqrt{1+\|\bA_P\|_\op^2}
        +
        \sqrt{
            \frac{\lambda_{\max}(\bSigma_E^P)}
                 {\lambda_{\min}(\bSigma_X^P)}
        }\,
    \right)^2\Wtwo(P,Q)^2\\
    &\leq
    4\kappa_P\Wtwo(P,Q)^2,
\end{align*}
proving the second claim.
\end{proof}

\subsection{Stability Implications for ERM and DRO (Remaining Proof for \cref{lem:stability-implications})}
\label{prf:stability-implications-ERM-DRO}
We employ a few helper lemmas, adopting the notation and block decompositions from Appendix~\ref{app:Bures-geometry}.

\begin{lemma}[Perturbed condition number]
\label{lem:perturbed-condition-number}
Fix $\bSigma\!\in\!\cC^*$ and $\bar{\bSigma}\!\in\!\PSD^d$ with
$\sB(\bar{\bSigma},\bSigma)
\!\leq\!\sqrt{\lambda_{\min}(\bSigma_X)}/2$.
Then $\bar{\bSigma}\in\cC^*$ and
$\kappa(\bar{\bSigma})\leq 19\kappa(\bSigma)$.
\end{lemma}
\begin{proof}
By \cref{thm:condition-number-formula}, the condition number is determined up to constants by the operator norm of the optimal coefficient matrix, the maximum eigenvalue of the residual second-moment matrix, and the minimum eigenvalue of the feature second-moment matrix. We now control each of these quantities for $\bar{\bSigma}$ in terms of those for $\bSigma$, beginning with the minimum feature eigenvalue.

Set $\eps \defeq \sB(\bar{\bSigma},\bSigma)$, $\lambda \defeq\lambda_{\min}(\bSigma_X) > 0$ and $\bar{\lambda}\defeq\lambda_{\min}(\bar{\bSigma}_X)$. By \cref{lem:Bures-lambda-min-continuity},
\[
    \sqrt{\bar{\lambda}}
    \geq\sqrt{\lambda}-\eps
    \geq\sqrt{\lambda}/2,
\]
since $\eps \leq \sqrt{\lambda}/2$. Thus
$\bar{\lambda}\geq\lambda/4$ and $\bar{\bSigma} \in \cC^*$.

Next, let $\bA\defeq\bA_\bSigma$,
$\bar{\bA}\defeq\bA_{\bar{\bSigma}}$, and $k \defeq \kappa(\bSigma)$. We then bound
\begin{align*}
    \sqrt{\bar{\lambda}}\,
    \|\bar{\bA}-\bA\|_\op &\leq \sqrt{\bar{\lambda}}\,
    \|\bar{\bA}-\bA\|_\Frob \tag{\cref{lem:basic-matrix-norms}}\\
    &\leq \bigl\|(\bar{\bA}-\bA)\bar{\bSigma}_X^{1/2}\bigr\|_\Frob \\
    &=\sqrt{R(\bA,\bar{\bSigma})} \tag{regret formula}\\
    &\leq 2\sqrt{k}\,\eps \tag{\cref{lem:regret-stability}}.
\end{align*}
Since $\eps/\sqrt{\bar{\lambda}}\leq 1$ and
$\|\bA\|_\op\leq\sqrt{k}$ by
\cref{thm:condition-number-formula}, it follows that
\[
    \|\bar{\bA}\|_\op
    \leq\|\bA\|_\op
    +\frac{2\sqrt{k}\eps}{\sqrt{\bar{\lambda}}}
    \leq 3\sqrt{k}.
\]

We now bound the residual relative to the smallest feature eigenvalue.
By \cref{lem:Bures-Procrustes}, there are square-root factors
$\bS,\bar{\bS}\in\R^{d\times d}$ of $\bSigma,\bar{\bSigma}$ with
$\bDelta\defeq\bar{\bS}-\bS$ satisfying
$\|\bDelta\|_\Frob\leq\eps$.
Define the corresponding residual factors and projection
\[
    \bE\defeq\bS_Y-\bA\bS_X,
    \qquad
    \bar{\bE}\defeq\bar{\bS}_Y-\bar{\bA}\bar{\bS}_X,
    \qquad
    \bar{\bPi}_X\defeq\bar{\bS}_X^\dagger\bar{\bS}_X.
\]
By \cref{lem:regret-projection-formula},
$\bE\bE^\top=\bSigma_E$,
$\bar{\bE}\bar{\bE}^\top=\bar{\bSigma}_E$, $\bar{\bE} =\bar{\bS}_Y(\bI_d-\bar{\bPi}_X)$, and $\bar{\bS}_X(\bI_d-\bar{\bPi}_X)=\bZero$. Further,
\begin{align*}
    \bar{\bS}_Y = \bS_Y + \bDelta_Y = \bE + \bA \bS_X + \bDelta_Y = \bE + \bA \bar{\bS}_X - \bA \bDelta_X + \bDelta_Y.
\end{align*}
We thus compute
\begin{align*}
    \bar{\bE} &= \bar{\bS}_Y(\bI_d-\bar{\bPi}_X)\\
    &= (\bE + \bA \bar{\bS}_X - \bA \bDelta_X + \bDelta_Y)(\bI_d-\bar{\bPi}_X)\\
    &= (\bE+\bDelta_Y-\bA\bDelta_X)(\bI_d-\bar{\bPi}_X).
\end{align*}
Since the right-hand projection is a contraction, we have
\begin{align*}
    \|\bar{\bE}\|_\op
    &\leq
    \|\bE\|_\op+\|\bDelta_Y-\bA\bDelta_X\|_\op\\
    &\leq
    \sqrt{\|\bSigma_E\|_\op}
    +\sqrt{1+\|\bA\|_\op^2}\,\eps\\
    &\leq
    \sqrt{k}\,(\sqrt{\lambda}+\eps).
\end{align*}
Here the second inequality uses
$\|\begin{pmatrix}-\bA & \bI_m\end{pmatrix}\|_\op=\sqrt{1+\|\bA\|_\op^2}$,
and the last uses \cref{thm:condition-number-formula}.
Therefore, we bound
\[
    \frac{\|\bar{\bSigma}_E\|_\op}{\bar{\lambda}}
    =
    \frac{\|\bar{\bE}\|_\op^2}{\bar{\lambda}}
    \leq
    k\frac{(\sqrt{\lambda}+\eps)^2}
            {(\sqrt{\lambda}-\eps)^2}
    \leq 9k,
\]
where the last inequality uses $\eps\leq\sqrt{\lambda}/2$.

Finally, \cref{thm:condition-number-formula} gives
\[
    \kappa(\bar{\bSigma})
    \leq
    1+\|\bar{\bA}\|_\op^2
    +\frac{\|\bar{\bSigma}_E\|_\op}{\bar{\lambda}}
    \leq 1+18 k
    \leq 19 k
\]
using $k \geq1$.
\end{proof}

\begin{lemma}[DRO strong duality]
\label{lem:DRO-minimax}
Let $\eps \geq 0$. For all $P \in \cP^*$ and $\hat{P} \in \cB_\eps(P)$, we have
\begin{equation*}
    \inf_{\bA \in \cA} \sup_{Q \in \cB_\eps(\hat{P})} L(\bA,Q) = \max_{Q \in \cB_\eps(\hat{P})} \inf_{\bA \in \cA} L(\bA,Q).
\end{equation*}
In particular, a maximizer exists for the right-hand side.
\end{lemma}
\begin{proof}
By \cref{lem:W2-Bures-equivalence}, it suffices to prove
\begin{equation}
\label{eq:DRO-strong-duality}
    \inf_{\bA \in \cA} \max_{\bSigma \in K} L(\bA,\bSigma) = \max_{\bSigma \in K} \inf_{\bA \in \cA} L(\bA,\bSigma),
\end{equation}
where $K \defeq \cB_\eps(\bSigma_{\hat{P}})$.
The factor representation in \cref{lem:W2-Bures-equivalence}
shows that $K$ is the continuous image of a compact Frobenius
ball and hence compact. It is convex because Wasserstein balls are convex under mixtures, as follows by mixing transport couplings, and the second-moment map is affine.
Since $L$ is continuous, convex in $\bA$, and affine in
$\bSigma$, Sion's minimax theorem \citep{sion1958general}
gives equality of the infimum--supremum values.
Finally, $\bSigma\mapsto\inf_{\bA\in\cA}L(\bA,\bSigma)$
is upper semicontinuous as an infimum of continuous functions,
so its supremum on $K$ is attained.
\end{proof}

We now prove the remaining excess risk bounds for ERM and DRO.

\begin{lemma}[Stability implications, ERM]
\label{lem:stability-implications-ERM}
Let $P\in\cP_2(\R^d)$ and $\eps > 0$. We then have
\begin{equation*}
    R_\eps(\ERM,P)
    \leq4\eps^2\sup_{Q\in\cB_\eps(P)}\kappa_Q.
\end{equation*}
If additionally $\eps^2\leq\lambda_{\min}(\bSigma_X^P)/4$,
then $R_\eps(\ERM,P)\leq 76\kappa_P\eps^2$.
\end{lemma}

\begin{proof}
The first bound is trivial if
$\cB_\eps(P)$ contains a law outside of $\cP^*$, since that law
has infinite condition number. We may therefore assume
$\cB_\eps(P)\subseteq\cP^*$.

For any $Q\in\cB_\eps(P)$, the ERM solutions at $P$ and $Q$
are unique. Applying \cref{lem:regret-stability} then gives
\begin{align*}
    R(\ERM[P],Q)
    &=R(\bA_P,Q)
    \leq4\kappa_P\eps^2,\\
    R(\ERM[Q],P)
    &=R(\bA_Q,P)
    \leq4\kappa_Q\eps^2.
\end{align*}
Since $P\in\cB_\eps(P)$, both bounds are at most
$4\eps^2\sup_{Q'\in\cB_\eps(P)}\kappa_{Q'}$.
Taking the maximum over the two orientations and then the
supremum over $Q\in\cB_\eps(P)$ proves the first bound.

If $\eps^2\leq\lambda_{\min}(\bSigma_X^P)/4$, then
\cref{lem:W2-Bures-equivalence,lem:perturbed-condition-number}
give $\kappa_Q\leq 19\kappa_P$ for every $Q\in\cB_\eps(P)$.
Substitution yields the second bound.
\end{proof}

\begin{lemma}[Stability implications, DRO]
\label{lem:stability-implications-DRO}
Let $P\in\cP_2(\R^d)$ and $\eps > 0$. We then have
\begin{equation*}
    R_\eps(\DRO_\eps,P)
    \leq16\eps^2\sup_{Q\in\cB_{2\eps}(P)}\kappa_Q.
\end{equation*}
If $\eps^2\leq\lambda_{\min}(\bSigma_X^P)/16$, this implies that
$R_\eps(\DRO_\eps,P)\leq 304\kappa_P\eps^2$.
\end{lemma}

\begin{proof}
The first bound is trivial if
$\cB_{2\eps}(P)$ contains a law outside of $\cP^*$, since that law
has infinite condition number. We may therefore assume
$\cB_{2\eps}(P)\subseteq\cP^*$.

Fix $Q\in\cB_\eps(P)$ and let
$(P_{\mathrm{s}},P_{\mathrm{t}})$ be either $(P,Q)$ or $(Q,P)$.
Set $\hat{\bA}\defeq\DRO_\eps[P_{\mathrm{s}}]$.
We first identify $\hat{\bA}$ as the ERM solution at a least-favorable law,
then transfer its zero regret there to the target.

By \cref{lem:DRO-minimax}, there exists a maximizer $\hat Q$ of
\[
    \max_{Q'\in\cB_\eps(P_{\mathrm{s}})}
    \inf_{\bA\in\cA}L(\bA,Q').
\]
DRO optimality and the minimax identity (\cref{lem:DRO-minimax}) give
\begin{align*}
    L(\hat{\bA},\hat Q)
    &\leq
    \sup_{Q'\in\cB_\eps(P_{\mathrm{s}})}
        L(\hat{\bA},Q')\\
    &=
    \inf_{\bA\in\cA}
    \sup_{Q'\in\cB_\eps(P_{\mathrm{s}})}
        L(\bA,Q')\\
    &=
    \max_{Q'\in\cB_\eps(P_{\mathrm{s}})}
    \inf_{\bA\in\cA}L(\bA,Q')\\
    &=
    \inf_{\bA\in\cA}L(\bA,\hat Q).
\end{align*}
The reverse inequality is automatic, so $\hat{\bA}$ minimizes
$L(\cdot,\hat Q)$.
Moreover,
$\hat Q\in\cB_\eps(P_{\mathrm{s}})
\subseteq\cB_{2\eps}(P)\subseteq\cP^*$.
Its risk minimizer is therefore unique, implying $\hat{\bA}=\bA_{\hat Q}$.

Since both $\hat Q$ and $P_{\mathrm{t}}$
belong to $\cB_\eps(P_{\mathrm{s}})$, we have
$\Wtwo(\hat Q,P_{\mathrm{t}})\leq2\eps$.
Applying \cref{lem:regret-stability} with reference law $\hat Q$
therefore gives
\begin{align*}
    R(\hat{\bA},P_{\mathrm{t}})
    &=R(\bA_{\hat Q},P_{\mathrm{t}}) \leq4\kappa_{\hat Q}\Wtwo(\hat Q,P_{\mathrm{t}})^2 \leq16\eps^2
        \sup_{Q'\in\cB_{2\eps}(P)}\kappa_{Q'}.
\end{align*}
Supremizing over source--target pairs $Q\in\cB_\eps(P)$ yields the claimed bound for $R_\eps(\DRO_\eps,P)$.

Finally, if $\eps^2\leq\lambda_{\min}(\bSigma_X^P)/16$, then
$2\eps\leq\sqrt{\lambda_{\min}(\bSigma_X^P)}/2$.
By \cref{lem:W2-Bures-equivalence,lem:perturbed-condition-number},
$\kappa_{Q'}\leq19\kappa_P$ for all $Q' \in \cB_{2\eps}(P)$.
Substitution then gives $R_\eps(\DRO_\eps,P)\leq 304\kappa_P\eps^2$.
\end{proof}

\subsection{Rank-Adaptive DRO Bound
(Proof of \cref{lem:DRO-rank-adaptive})}
\label{prf:DRO-rank-adaptive}

To start, we establish an additional robust regret bound for DRO by more directly exploiting its min--max problem formulation. Similar bounds exist in the literature, typically stated for more generic loss functions \citep{gao2024wasserstein,lee2018minimax}.

\begin{lemma}[DRO regularizer bound]
\label{lem:DRO-regularizer}
For $P\in\cP^*$, define
\[
    \Gamma_P
    \defeq
    \sqrt{\tr\bigl(
        (\bI_m+\bA_P\bA_P^\top)\bSigma_E^P
    \bigr)}.
\]
Then, for every $\eps\geq0$,
\begin{equation}
\label{eq:DRO-regularizer-intermediate}
    R_\eps(\DRO_\eps,P)
    \leq4\Gamma_P\eps+5\kappa_P\eps^2.
\end{equation}
\end{lemma}

\begin{proof}
Fix $P\in\cP^*$, and abbreviate
\[
    \bA\defeq\bA_P,
    \qquad
    \bS\defeq\bSigma_P^{1/2},
    \qquad
    \bM\defeq\begin{pmatrix}\bA&-\bI_m\end{pmatrix},
    \qquad
    \bE\defeq\bS_Y-\bA\bS_X.
\]
By \cref{lem:regret-projection-formula},
$\bE\bE^\top=\bSigma_E^P$. Hence
\[
    \|\bM^\top\bE\|_\Frob^2
    =
    \tr\bigl(
        (\bI_m+\bA\bA^\top)\bSigma_E^P
    \bigr)
    =
    \Gamma_P^2.
\]
Also,
$\|\bM\|_\op^2=1+\|\bA\|_\op^2\leq\kappa_P$
by \cref{thm:condition-number-formula}.

We first bound changes in the risk of the fixed predictor $\bA$ in the neighborhood of $P$.
For any $\delta\geq0$ and $Q\in\cB_\delta(P)$,
\cref{lem:W2-Bures-equivalence} provides
$\bDelta\in\R^{d\times d}$ such that
\[
    \bSigma_Q=(\bS+\bDelta)(\bS+\bDelta)^\top,
    \qquad
    \|\bDelta\|_\Frob\leq\delta.
\]
Since the residual factor at $Q$ for $\bA$ is
$\bE-\bM\bDelta$, we have by \cref{lem:regret-projection-formula} that
\begin{align*}
    L(\bA,Q)-L(\bA,P)
    &=
    \|\bE-\bM\bDelta\|_\Frob^2-\|\bE\|_\Frob^2\\
    &=
    -2\langle\bM^\top\bE,\bDelta\rangle_\Frob
    +\|\bM\bDelta\|_\Frob^2.
\end{align*}
Cauchy--Schwarz and
$\|\bM\bDelta\|_\Frob^2 \leq \|\bM\|_\op^2 \|\bDelta\|_\Frob^2 \leq\kappa_P\delta^2$
then give the two-sided bound
\begin{equation}
\label{eq:DRO-comparator-risk-variation}
    -2\Gamma_P\delta
    \leq
    L(\bA,Q)-L(\bA,P)
    \leq
    2\Gamma_P\delta+\kappa_P\delta^2.
\end{equation}
For the lower bound, we have simply discarded the nonnegative
quadratic term.

\textbf{Target role.} We now bound regret when $P$ is the target law. Fix $Q\in\cB_\eps(P)$ and set
$\hat{\bA}\defeq\DRO_\eps[Q]$.
Since $P\in\cB_\eps(Q)$, DRO optimality and a comparison with
$\bA$ imply that
\begin{align*}
    L(\hat{\bA},P)
    &\leq
    \sup_{Q'\in\cB_\eps(Q)}L(\hat{\bA},Q')\\
    &\leq
    \sup_{Q'\in\cB_\eps(Q)}L(\bA,Q')\\
    &\leq
    \sup_{Q'\in\cB_{2\eps}(P)}L(\bA,Q')\\
    &\leq
    L(\bA,P)+4\Gamma_P\eps+4\kappa_P\eps^2.
\end{align*}
The third inequality uses
$\cB_\eps(Q)\subseteq\cB_{2\eps}(P)$,
and the last applies
\eqref{eq:DRO-comparator-risk-variation} with $\delta=2\eps$.
Subtracting $L(\bA,P)=\inf_{\bB\in\cA}L(\bB,P)$ gives
\[
    R(\DRO_\eps[Q],P)
    \leq4\Gamma_P\eps+4\kappa_P\eps^2.
\]
We note that, using the terminology of \cite{gao2024wasserstein}, the RHS is an upper bound on the Wasserstein regularizer
of the loss $(x,y)\mapsto\|\bA_Px-y\|^2$ at $P$, evaluated
at radius $2\eps$.

\textbf{Source role.}
Now set $\hat{\bA}\defeq\DRO_\eps[P]$.
Comparing with $\bA$, we bound regret at $Q \in \cB_\eps(P)$ by
\begin{align*}
    R(\hat{\bA},Q)
    &=
    L(\hat{\bA},Q)-L(\bA,Q)+R(\bA,Q)\\
    &\leq
    \sup_{Q'\in\cB_\eps(P)}L(\bA,Q')
    -L(\bA,Q)+R(\bA,Q)\\
    &\leq
    4\Gamma_P\eps+\kappa_P\eps^2+R(\bA,Q)\\
    &\leq
    4\Gamma_P\eps+5\kappa_P\eps^2.
\end{align*}
The first inequality uses feasibility of $Q$ and DRO optimality.
The second applies the upper bound in
\eqref{eq:DRO-comparator-risk-variation} to $Q'$ and its lower
bound to $Q$, both with $\delta=\eps$.
The last uses
$R(\bA_P,Q)\leq4\kappa_P\Wtwo(P,Q)^2$
from \cref{lem:regret-stability}.

Taking the maximum of these two role-specific bounds and then
the supremum over $Q\in\cB_\eps(P)$ gives
\eqref{eq:DRO-regularizer-intermediate}.
\end{proof}

Next, we combine \cref{lem:DRO-regularizer} with \cref{lem:stability-implications-DRO} to obtain the claimed result. 

\begin{proof}[Proof of \cref{lem:DRO-rank-adaptive}]
Fixing $P\in\cP^*$, we abbreviate
\[
    \lambda\defeq\lambda_{\min}(\bSigma_X^P),
    \qquad
    \mu\defeq \lambda_{\max}(\bSigma_E^P), \qquad r \defeq \reff(\bSigma_E^P),
    \qquad
    s\defeq\kappa_P-1.
\]
The condition number formula (\cref{thm:condition-number-formula}) gives
\[
    \|\bA_P\|_\op^2\leq s,
    \qquad
    \mu\leq s\lambda.
\]
Also $\tr(\bSigma_E^P)=r\mu$, including when
$\bSigma_E^P=\bZero$ under the convention $r=0$.

If $\eps^2\leq\lambda/16$, the small-radius bound in
\cref{lem:stability-implications-DRO} gives
\[
    R_\eps(\DRO_\eps,P)\lesssim\kappa_P\eps^2,
\]
which suffices. We therefore assume $\eps^2>\lambda/16$ below.

To bound $\Gamma_P$, we estimate its coefficient-dependent
trace term in two ways:
\begin{align*}
    \tr(\bA_P\bA_P^\top\bSigma_E^P)
    &\leq
    \|\bA_P\|_\op^2\tr(\bSigma_E^P)
    =r\mu\|\bA_P\|_\op^2,\\
    \tr(\bA_P\bA_P^\top\bSigma_E^P)
    &\leq
    \mu\|\bA_P\|_\Frob^2
    \leq p\mu\|\bA_P\|_\op^2,
\end{align*}
with the inequalities using \cref{lem:basic-matrix-norms}.
Taking the smaller bound and using
$\|\bA_P\|_\op^2\leq s$ yields
\begin{align*}
    \Gamma_P^2
    &=
    \tr(\bSigma_E^P)
    +\tr(\bA_P\bA_P^\top\bSigma_E^P)\\
    &\leq
    \mu\bigl(r+s\min\{p,r\}\bigr)\\
    &\leq
    \lambda\bigl(rs+s^2\min\{p,r\}\bigr).
\end{align*}
Using $\sqrt{a+b}\leq\sqrt a+\sqrt b$ for $a,b\geq0$, we further bound
\[
    \Gamma_P
    \leq
    \sqrt{\lambda}
    \left(\sqrt{rs}+s\sqrt{\min\{p,r\}}\right).
\]
Since $\sqrt{\lambda}<4\eps$, \cref{lem:DRO-regularizer} now gives
\begin{align*}
    R_\eps(\DRO_\eps,P)
    &\leq
    4\Gamma_P\eps+5\kappa_P\eps^2\\
    &\leq
    \left(
        16s\sqrt{\min\{p,r\}}
        +16\sqrt{rs}
        +5\kappa_P
    \right)\eps^2\\
    &\lesssim
    \left(
        \sqrt{\min\{p,r\}}\,(\kappa_P-1)
        +\sqrt{r(\kappa_P-1)}
        +\kappa_P
    \right)\eps^2,
\end{align*}
as desired.
\end{proof}

\subsection{Main Population-Limit Regret Bounds (Proof of \cref{thm:pop-limit-risk-bds})}
\label{prf:pop-limit-risk-bds}

We start with upper bounds, which follow from our existing results. Indeed, \cref{lem:stability-implications} implies that $R_\eps(\SNR_\eps,\cG), R_\eps(\MRO_\eps,\cG) \lesssim \kappa \eps^2$. Moreover, \cref{lem:DRO-rank-adaptive} and the basic effective rank bound $\reff(\bSigma_E^P) \leq m$ imply that
\begin{equation*}
    R_\eps(\DRO_\eps,\cG) \lesssim \left(
        \sqrt{\min\{p,m\}}\,(\kappa-1)
        +\sqrt{m(\kappa-1)}
        +\kappa
    \right)\eps^2,
\end{equation*}
as desired. Of course, no upper bound was claimed for ERM.

For the lower bounds, we scale up the local minimax result to obtain the minimax lower bound for $\cG$. Then, we give an explicit construction establishing unbounded regret for ERM. Two further constructions, whose analysis is more subtle, yield the dimension-dependent terms in the DRO rate.

We start with a helper lemma that demonstrates a certain invariance to rescaling.

\begin{lemma}[Rescaling invariance]
\label{lem:rescaling-invariance}
Let $P,Q\in\cP_2(\R^d)$ and
$\bA\in\cA$. For $\lambda>0$, let $P_\lambda$ and $Q_\lambda$ denote the pushforwards of $P$ and $Q$, respectively, 
under $z\mapsto\lambda z$. We then have
\begin{align*}
    \Wtwo(P_\lambda,Q_\lambda)
    &=\lambda\Wtwo(P,Q),\\
    L(\bA,P_\lambda)&=\lambda^2L(\bA,P),\\
    R(\bA,P_\lambda)&=\lambda^2R(\bA,P),\\
    \kappa_{P_\lambda}&=\kappa_P.
\end{align*}
Moreover, for $\eps\geq0$, $R^\star_{\lambda\eps}(P_\lambda)
    =\lambda^2R^\star_\eps(P)$, and the optimization problems defining
$\DRO_{\lambda\eps}[P_\lambda]$ and $\DRO_\eps[P]$
have the same minimizer set.
\end{lemma}

\begin{proof}
Scaling both coordinates of a transport coupling multiplies its expected
squared cost by $\lambda^2$. Applying the same argument to the
inverse scaling proves the Wasserstein identity.
The risk and regret identities follow from
$\bSigma_{P_\lambda}=\lambda^2\bSigma_P$.
The condition number formula is invariant under this scaling,
as is membership in $\cP^*$.

The Wasserstein identity implies that $\cB_{\lambda\eps}(P_\lambda)
    =\{Q_\lambda:Q\in\cB_\eps(P)\}$.
Combining this with regret homogeneity and
\eqref{eq:local-minimax} proves the local minimax identity.
Likewise,
\[
    \sup_{Q'\in\cB_{\lambda\eps}(P_\lambda)}L(\bA,Q')
    =
    \lambda^2\sup_{Q\in\cB_\eps(P)}L(\bA,Q),
\]
so the DRO minimizer sets coincide.
\end{proof}

We now turn to the minimax lower bound for $\cG(\kappa)$.

\begin{lemma}[Minimax lower bound]
\label{lem:minimax-lb}
For every $\eps > 0$ and $\kappa \geq 1$, we have $R^\star_\eps(\cG(\kappa))\geq\kappa\eps^2$.
\end{lemma}

\begin{proof}
Let $P \in \cP_2(\R^d)$ be the law of $(X,Y)$ where
\[
    X\sim\cN(0,\bI_p),
    \qquad
    Y=\sqrt{\kappa-1}\,X_1e_1,
    \qquad e_1\in\R^m.
\]
By construction, $\bA_P$ has only one potentially nonzero entry,
$\sqrt{\kappa-1}$ in position $(1,1)$, and the optimal residual
is zero. Thus, \cref{thm:condition-number-formula} gives
$\kappa_P=\kappa$.

For $t>0$, let $P_t$ be the pushforward of $P$ under $z\mapsto tz$.
By \cref{lem:rescaling-invariance}, $P_t\in\cG(\kappa)$ and
\[
    R^\star_\eps(\cG(\kappa))
    \geq R^\star_\eps(P_t)
    =t^2R^\star_{\eps/t}(P).
\]
Further, the local expansion in \cref{thm:condition-number-formula} gives
\[
    \lim_{t\to\infty}t^2R^\star_{\eps/t}(P)
    =\kappa\eps^2.
\]
Since the preceding lower bound holds for every $t>0$,
the claim follows.
\end{proof}

Next, we prove the lower bound for ERM.

\begin{lemma}[ERM lower bound]
For every $\eps>0$ and $\kappa\geq1$,
$R_\eps(\ERM,\cG(\kappa))=\infty$, independently of the
choice of ERM minimizer.
\end{lemma}

\begin{proof}
Consider target $P \in \cP_2(\R^d)$ defined as the law of $(\eps W/2,0_m)$ where $W\sim\cN(0,\bI_p)$.
Then $\bSigma_X^P=(\eps^2/4)\bI_p$,
$\bA_P=\bZero$, and $\bSigma_E^P=\bZero$.
Hence $\kappa_P=1$ and $P\in\cG(\kappa)$.

For $\delta\in(0,1)$, consider source $Q_\delta \in \cP_2(\R^d)$ taken as the law of
\[
    \tilde X=\frac{\eps}{2}(\delta W_1,W_2,\ldots,W_p),
    \qquad
    \tilde Y=\frac{\eps}{2}W_1e_1,
    \qquad e_1\in\R^m.
\]
The coupling using the same $W$ gives
\[
    \Wtwo(P,Q_\delta)^2
    \leq\frac{\eps^2}{4}\bigl((1-\delta)^2+1\bigr)
    \leq\eps^2.
\]
The source feature second-moment matrix is positive definite, and its
unique ERM solution has entry $1/\delta$ in position $(1,1)$
and zeros elsewhere. Therefore,
\[
    R_\eps(\ERM,\cG(\kappa))
    \geq R(\ERM[Q_\delta],P)
    = \frac{\eps^2}{4\delta^2}.
\]
Letting $\delta\downarrow0$ proves the claim.
\end{proof}

For the DRO result, we combine two lower bounds in addition to the minimax lower bound. Although DRO solutions are generally difficult to characterize, we are able to streamline analysis in both cases by exploiting rotational symmetry. We fix $\eps = 1$ for now to simplify presentation.

\begin{lemma}[DRO lower bound, A]
\label{lem:DRO-LB-A}
For every $\kappa\geq1$,
\[
    R_1(\DRO_1,\cG(\kappa))
    \gtrsim\sqrt{\min\{p,m\}}\,(\kappa-1),
\]
independently of the choice of DRO minimizer.
\end{lemma}
\begin{proof}
The claim is immediate for $\kappa=1$.
Write $q \defeq \min\{p,m\}$.
If $q<65$, then \cref{lem:minimax-lb} gives
\[
    R_1(\DRO_1,\cG(\kappa))
    \geq R_1^\star(\cG(\kappa))
    \gtrsim\kappa
    \gtrsim\sqrt q\,(\kappa-1).
\]
We may therefore assume $\kappa>1$ and $q \geq 65$.

\textbf{Construction.} Set $b\defeq\sqrt{(\kappa-1)/2} > 0$.
Let $X\sim\cN(0,\bI_p)$ and
\begin{equation*}
    E \sim \cN\bigl(0,\diag(0,b^2\bI_{q-1},\bZero_{m-q})\bigr)
\end{equation*}
be independent. Take target $P$ as the law of $(X,Y)$ where
\begin{equation*}
    Y = \begin{pmatrix}bX_{1:q}\\0_{m-q}\end{pmatrix}+ E
\end{equation*}
Here and below, empty blocks of size zero should be omitted. By this construction,
\[
    \bA_P=
    \begin{pmatrix}
        b\bI_q&\bZero_{q\times(p-q)}\\
        \bZero_{(m-q)\times q}&\bZero_{(m-q)\times(p-q)}
    \end{pmatrix},
    \qquad
    \bSigma_X^P=\bI_p,
\]
and $\bSigma_E^P=\diag(0,b^2\bI_{q-1},\bZero_{m-q})$. Thus,
\[
    \bA_P\bA_P^\top
    +\lambda_{\min}(\bSigma_X^P)^{-1}\bSigma_E^P
    =
    \diag(b^2,2b^2\bI_{q-1},\bZero_{m-q}),
\]
and \cref{thm:condition-number-formula} gives $\kappa_P=1+2b^2=\kappa$.

For the perturbed source, take $\tilde{P}$ as the law of $(\tilde{X},Y)$, where $\tilde{X}_1 = c X_1$ and $\tilde{X}_{2:p} = X_{2:p}$, for $c \in (0,1]$ to be tuned. This construction ensures $\Wtwo(\tilde{P},P)^2 \leq \E[\|X - \tilde{X}\|^2] \leq 1$.

\textbf{Average over symmetries.} Consider the compact subgroup
\[
    U\defeq
    \left\{
        \diag(1,\bH,s\bI_{p-q},1,\bH,t\bI_{m-q}):
        \bH\in\OO^{q-1},\ s,t\in\{-1,1\}
    \right\}
    \subseteq\OO^d.
\]
Both $P$ and $\tilde P$ are invariant under pushforwards by each $\bU \in U$.
Moreover, the invariant coefficient matrices (in the sense of \cref{lem:DRO-symmetry}) are exactly
\[
    \bA(\alpha,\beta)
    \defeq
    \begin{pmatrix}
        \diag(\alpha,\beta\bI_{q-1})&\bZero_{q\times(p-q)}\\
        \bZero_{(m-q)\times q}&\bZero_{(m-q)\times(p-q)}
    \end{pmatrix},
    \qquad \alpha,\beta\in\R.
\]
Indeed, varying $s$ and $t$ eliminates the extra feature columns
and response rows. Within the leading $q\times q$ block,
taking $\bH=-\bI_{q-1}$ eliminates the off-diagonal blocks,
and commuting with every $\bH\in\OO^{q-1}$ forces the remaining
$(q-1)\times(q-1)$ block to be a scalar multiple of the identity.
By \cref{lem:DRO-symmetry}, it therefore suffices to lower bound
the target regret of invariant DRO optimizers.

\textbf{Construct candidate DRO optimizer.} Define the invariant
objective by
\[
    J(\alpha,\beta)
    \defeq
    \sup_{Q\in\cB_1(\tilde P)}
    L(\bA(\alpha,\beta),Q).
\]
We have
\[
    \bA(\alpha,\beta)\bA(\alpha,\beta)^\top
    =
    \diag(\alpha^2,\beta^2\bI_{q-1},\bZero_{m-q}),
\]
and the coordinate residual second moments at the source are
\[
    \E\bigl[(\bA(\alpha,\beta)\tilde X-Y)_i^2\bigr]
    =
    \begin{cases}
        (\alpha c-b)^2, & i=1,\\
        (\beta-b)^2+b^2, & 2\leq i\leq q,\\
        0, & q<i\leq m.
    \end{cases}
\]
Thus \cref{cor:DRO-strong-duality-diagonal} with $\eps=1$ gives
\begin{equation}
\label{eq:J-strong-duality}
    J(\alpha,\beta)
    =
    \inf_{\eta>1+\max\{\alpha^2,\beta^2\}}
    \eta\left(
        1+\frac{(\alpha c-b)^2}{\eta-1-\alpha^2}
        +\frac{(q-1)[(\beta-b)^2+b^2]}{\eta-1-\beta^2}
    \right).
\end{equation}

Now consider the related problem associated with coordinates
$2,\ldots,q$:
\[
    V(\beta)
    \defeq
    \inf_{\eta>1+\beta^2}
    \eta\left(
        1+\frac{(q-1)[(\beta-b)^2+b^2]}{\eta-1-\beta^2}
    \right).
\]
By \cref{lem:elementary-scalar-ineqs}, with
$\delta=\eta-(1+\beta^2)$, this infimum is attained at
\[
    \eta(\beta)
    \defeq
    1+\beta^2+
    \sqrt{(1+\beta^2)(q-1)[(\beta-b)^2+b^2]},
\]
with value
\[
    V(\beta)
    =
    \left(
        \sqrt{1+\beta^2}
        +\sqrt{(q-1)[(\beta-b)^2+b^2]}
    \right)^2.
\]
This function is strictly convex and coercive.
Let $\beta_\star$ denote its unique minimizer, set
$\eta_\star\defeq\eta(\beta_\star)$, and let
$\alpha_\star\defeq\sqrt{\eta_\star-1} > 0$.
We set the free scaling parameter $c$ to $b/\alpha_\star > 0$.
Below we shall verify that $c \leq 1$, ensuring that
$\tilde P\in\cB_1(P)$.

\textbf{Certify optimality.} We next show that $J$ is uniquely minimized at $(\alpha_\star,\beta_\star)$.
For every feasible $\eta$ for \eqref{eq:J-strong-duality},
$\eta/(\eta-1-\alpha^2)\geq1$.
The remaining terms are bounded from below by $V(\beta)$, so
\[
    J(\alpha,\beta)
    \geq V(\beta)+(\alpha c-b)^2
    \geq V(\beta_\star).
\]
At $(\alpha_\star,\beta_\star)$, the first-coordinate residual
vanishes because $\alpha_\star c=b$.
Also $\alpha_\star^2>\beta_\star^2$, so
\[
    J(\alpha_\star,\beta_\star)
    =
    \inf_{\eta>\eta_\star}
    \eta\left(
        1+\frac{(q-1)[(\beta_\star-b)^2+b^2]}
                 {\eta-1-\beta_\star^2}
    \right).
\]
When minimizing instead over the larger range $\eta > 1 + \beta_\star^2$, our construction ensures that the minimum is attained at $\eta = \eta_\star$. Although $\eta_\star$ is excluded from the current feasible range, the objective is continuous there, so the infimum is approached as $\eta\downarrow\eta_\star$ and equals
$V(\beta_\star)$.
The preceding lower bound then forces every minimizing pair $(\alpha,\beta)$ to satisfy $\beta=\beta_\star$ and $\alpha c=b$. Since $c = b/\alpha_\star > 0$, this implies that $(\alpha_\star,\beta_\star)$ is the unique minimizing pair.

\textbf{Bound $\bm{\alpha_\star}$ and $\bm{\beta_\star}$ from below, verify $\bm{c}$ is admissible.}
Define
\[
    h(\beta)\defeq\sqrt{V(\beta)}
    =
    \sqrt{1+\beta^2}
    +\sqrt{q-1}\sqrt{(\beta-b)^2+b^2},
\]
so that $\beta_\star\in\argmin_\beta h(\beta)$.
This function is strictly convex and coercive.
For $\beta\leq b/2$,
\[
    h'(\beta)
    =
    \frac{\beta}{\sqrt{1+\beta^2}}
    +\sqrt{q-1}\,
      \frac{\beta-b}{\sqrt{(\beta-b)^2+b^2}}
    \leq1-\sqrt{\frac{q-1}{5}}<0.
\]
The inequality uses $\beta-b\leq-b/2$.
Hence $\beta_\star>b/2$.
It follows that
\begin{align}
    \alpha_\star^2
    &=\eta_\star-1\notag\\
    &\geq
    \sqrt{(1+\beta_\star^2)(q-1)
        [(\beta_\star-b)^2+b^2]}\notag\\
    &\geq
    \frac{b^2}{2}\sqrt{q-1}.\label{eq:alpha_star_lb}
\end{align}
Since $q\geq65$, this implies $\alpha_\star\geq2b$ and hence
$c=b/\alpha_\star \leq 1/2$.

\textbf{Lower bound regret.}
Since $c \leq 1$, $\tilde{P} \in \cB_1(P)$, and its unique invariant
DRO optimizer satisfies
\begin{align*}
    R(\bA(\alpha_\star,\beta_\star),P)
    &=
    (\alpha_\star-b)^2+(q-1)(\beta_\star-b)^2\\
    &\geq(\alpha_\star-b)^2\\
    &\geq\frac14\alpha_\star^2\\
    &\geq\frac{b^2}{8}\sqrt{q-1}
    \gtrsim\sqrt q\,(\kappa-1).
\end{align*}
The first equality uses $\bSigma_X^P=\bI_p$ and the regret
formula, the second inequality uses that $\alpha_\star \geq 2b$, the third inequality uses \eqref{eq:alpha_star_lb}, and the last inequality uses $b^2=(\kappa-1)/2$.

By \cref{lem:DRO-symmetry}, every DRO optimizer at $\tilde P$
has target regret at least that of
$\bA(\alpha_\star,\beta_\star)$.
Since $P\in\cG(\kappa)$ and $\tilde P\in\cB_1(P)$,
\[
    R_1(\DRO_1,\cG(\kappa))
    \geq R(\DRO_1[\tilde P],P)
    \geq R(\bA(\alpha_\star,\beta_\star),P)
    \gtrsim\sqrt q\,(\kappa-1).
\]
This proves the claim for every choice of DRO minimizer.
\end{proof}

\begin{lemma}[DRO lower bound, B]
\label{lem:DRO-LB-B}
For every $\kappa\geq1$,
\[
    R_1(\DRO_1,\cG(\kappa))
    \gtrsim\sqrt{m(\kappa-1)},
\]
independently of the choice of DRO minimizer.
\end{lemma}

\begin{proof}
The claim is immediate for $\kappa=1$.
If $m=1$, then \cref{lem:minimax-lb} gives
\[
    R_1(\DRO_1,\cG(\kappa))
    \geq R_1^\star(\cG(\kappa))
    \gtrsim\kappa
    \geq\sqrt{\kappa-1}.
\]
We may therefore assume $\kappa>1$ and $m\geq2$.

\textbf{Construction.} Set
\[
    s\defeq\kappa-1>0,
    \qquad
    N\defeq(m-1)s>0,
    \qquad
    a \defeq \frac12\min\bigl\{\sqrt{s},N^{1/4}\bigr\}>0.
\]
Let $X\sim\cN(0,\bI_p)$ and
$E\sim\cN(0,\diag(0,s\bI_{m-1}))$ be independent. Fix target $P \in \cP_2(\R^d)$ as the law of $(X,Y)$, where
\[
    Y=aX_1e_1+E,
    \qquad e_1\in\R^m.
\]
By this construction,
\[
    \bA_P=
    \begin{pmatrix}
        ae_1&\bZero_{m\times(p-1)}
    \end{pmatrix},
    \qquad
    \bSigma_X^P=\bI_p,
    \qquad
    \bSigma_E^P=\diag(0,s\bI_{m-1}).
\]
Thus
\[
    \bA_P\bA_P^\top
    +\lambda_{\min}(\bSigma_X^P)^{-1}\bSigma_E^P
    =
    \diag(a^2,s\bI_{m-1}).
\]
Since $a^2\leq s$ and $m\geq2$,
\cref{thm:condition-number-formula} gives
$\kappa_P=1+s=\kappa$.

For the perturbed source, take $\tilde P$ as the law of
$(\tilde X,Y)$, where
$\tilde X_1=cX_1$ and $\tilde X_{2:p}=X_{2:p}$,
for $c\in(0,1]$ to be specified.
Coupling the same $X,E$ gives
\[
    \Wtwo(\tilde P,P)^2
    \leq\E\|X-\tilde X\|^2
    =(1-c)^2\leq1.
\]

\textbf{Average over symmetries.} Consider the compact subgroup
\[
    U\defeq
    \left\{
        \diag(1,t\bI_{p-1},1,\bH):
        t\in\{-1,1\},\ \bH\in\OO^{m-1}
    \right\}
    \subseteq\OO^d.
\]
Both $P$ and $\tilde P$ are invariant under pushforwards by
each $\bU\in U$.
The invariant coefficient matrices (in the sense of \cref{lem:DRO-symmetry}) are exactly
\[
    \bA(\alpha)
    \defeq
    \begin{pmatrix}
        \alpha e_1&\bZero_{m\times(p-1)}
    \end{pmatrix},
    \qquad \alpha\in\R.
\]
Indeed, varying $t$ eliminates columns $2,\ldots,p$, and
taking $\bH=-\bI_{m-1}$ eliminates rows $2,\ldots,m$.
By \cref{lem:DRO-symmetry}, it therefore suffices to lower bound
the target regret of invariant DRO optimizers.

\textbf{Construct candidate DRO optimizer.} Define the invariant
objective by
\[
    J(\alpha)
    \defeq
    \sup_{Q\in\cB_1(\tilde P)}L(\bA(\alpha),Q).
\]
We have
\[
    \bA(\alpha)\bA(\alpha)^\top
    =\diag(\alpha^2,\bZero_{m-1}),
\]
and the coordinate residual second moments at the source are
\[
    \E\bigl[(\bA(\alpha)\tilde X-Y)_i^2\bigr]
    =
    \begin{cases}
        (\alpha c-a)^2, & i=1,\\
        s, & 2\leq i\leq m.
    \end{cases}
\]
Thus \cref{cor:DRO-strong-duality-diagonal} with $\eps=1$ gives
\begin{equation}
\label{eq:DRO-LB-B-dual}
    J(\alpha)
    =
    \inf_{\eta>1+\alpha^2}
    \eta\left(
        1+\frac{(\alpha c-a)^2}{\eta-1-\alpha^2}
        +\frac{N}{\eta-1}
    \right).
\end{equation}

Now consider the related problem associated with coordinates
$2,\ldots,m$:
\[
    V
    \defeq
    \inf_{\eta>1}\eta\left(1+\frac{N}{\eta-1}\right).
\]
By \cref{lem:elementary-scalar-ineqs}, with $\delta=\eta-1$,
this infimum is attained at $\eta_\star\defeq1+\sqrt N$ and $V=(1+\sqrt N)^2$. We now set $\alpha_\star\defeq\sqrt{\eta_\star-1}=N^{1/4}$ and $c\defeq a/\alpha_\star$.
Our choice of $a$ gives $0<a\leq\alpha_\star/2$ and hence
$0<c\leq1/2$. Thus the source construction is admissible.

\textbf{Certify optimality.} We next show that $J$ is uniquely
minimized at $\alpha_\star$.
For every feasible $\eta$ in \eqref{eq:DRO-LB-B-dual},
$\eta/(\eta-1-\alpha^2)\geq1$.
The remaining terms are bounded from below by $V$, so
\[
    J(\alpha)\geq V+(\alpha c-a)^2\geq V.
\]
At $\alpha=\alpha_\star$, the first-coordinate residual vanishes
because $\alpha_\star c=a$.
Since $1+\alpha_\star^2=\eta_\star$, we obtain
\[
    J(\alpha_\star)
    =
    \inf_{\eta>\eta_\star}
    \eta\left(1+\frac{N}{\eta-1}\right).
\]
When minimizing instead over the larger range $\eta>1$,
the bulk minimum is attained at $\eta = \eta_\star$.
Although $\eta_\star$ is excluded from the restricted range,
the objective is continuous there, so the infimum is
approached as $\eta\downarrow\eta_\star$ and equals $V$.
Thus $J(\alpha_\star)=V$.

If $\alpha$ is any minimizer of $J$, the preceding lower bound gives $0=J(\alpha)-V\geq(\alpha c-a)^2$. Hence, $\alpha c=a$ and, since $c>0$,
$\alpha=a/c=\alpha_\star$.
Thus, $\bA(\alpha_\star)$ is the unique invariant
DRO optimizer.

\textbf{Lower bound regret.}
Since $\bSigma_X^P=\bI_p$, the regret formula and our choice of $c = a/\alpha_\star$ give
\begin{align*}
    R(\bA(\alpha_\star),P)
    &=
    \|\bA(\alpha_\star)-\bA_P\|_\Frob^2\\
    &=
    (\alpha_\star-a)^2\\
    &=
    \alpha_\star^2(1-c)^2\\
    &\geq
    \frac14\alpha_\star^2
    =
    \frac14\sqrt{(m-1)(\kappa-1)}
    \gtrsim\sqrt{m(\kappa-1)}.
\end{align*}
The inequality uses $c\leq1/2$, and the final comparison uses
$m-1\geq m/2$ for $m\geq2$.

By \cref{lem:DRO-symmetry}, every DRO optimizer at $\tilde P$
has target regret at least that of $\bA(\alpha_\star)$.
Since $P\in\cG(\kappa)$ and $\tilde P\in\cB_1(P)$,
\[
    R_1(\DRO_1,\cG(\kappa))
    \geq R(\DRO_1[\tilde P],P)
    \geq R(\bA(\alpha_\star),P)
    \gtrsim\sqrt{m(\kappa-1)}.
\]
This proves the claim for every choice of DRO minimizer.
\end{proof}

Combining the results above, we obtain the following unified lower bound.

\begin{lemma}[DRO lower bound]
For all $\eps > 0$ and $\kappa \geq 1$, we have
\begin{equation*}
    R_\eps(\DRO_\eps,\cG(\kappa)) \gtrsim \bigl(\sqrt{\min\{p,m\}} (\kappa - 1) + \sqrt{m(\kappa - 1)} + \kappa\,\bigr) \eps^2,
\end{equation*}
no matter how ties are broken to select DRO solutions.
\end{lemma}
\begin{proof}
Scale the hard source--target pairs from
\cref{lem:DRO-LB-A,lem:DRO-LB-B} by $\eps$.
By \cref{lem:rescaling-invariance}, their condition numbers
are unchanged, their Wasserstein distances are multiplied
by $\eps$, and the DRO minimizer sets are unchanged when
the supplied radius changes from $1$ to $\eps$.
Their target regrets are multiplied by $\eps^2$.
Combining these bounds with \cref{lem:minimax-lb} and using
$\max\{a,b,c\}\geq(a+b+c)/3$ gives the claim.
\end{proof}

\section{Remaining Proofs for \cref{sec:computation}}
\label{prf:computation}

\subsection{Convex Reformulation for MRO
(Proof of \cref{thm:MRO-convex-reformulation})}
\label{prf:MRO-convex-reformulation}

Throughout, we use the notation from
\cref{thm:MRO-convex-reformulation} and the block decompositions
from \cref{app:block-decomp}.
The observed matrix $\tilde{\bSigma}$, its square-root factor
$\tilde{\bS}$, and
the radius $\eps>0$ are fixed, and their dependence is suppressed.
We use $\sup\varnothing=-\infty$.
Recalling $\tilde{\bE}_\bA=\tilde{\bS}_Y-\bA\tilde{\bS}_X$,
our goal is to prove
\[
    F(\bA)
    =
    \inf_{\eta>b_\bA}
    \eta\left(
        \eps^2-\|\tilde{\bS}_X\|_\Frob^2
        +\tr_p(\bH_{\bA,\eta})
    \right)
    \qquad\forall\,\bA\in\cA,
\]
where $b_\bA\defeq1+\|\bA\|_\op^2$ and, for $\eta>b_\bA$,
\[
    \bH_{\bA,\eta}\defeq
    \tilde{\bS}_X^\top\tilde{\bS}_X
    +\tilde{\bE}_\bA^\top
      \bigl((\eta-1)\bI_m-\bA\bA^\top\bigr)^{-1}
      \tilde{\bE}_\bA.
\]
We must also show that the displayed objective, namely
$D(\bA,\eta)$, is jointly convex in $(\bA,\eta)$ on its convex
domain $\eta>b_\bA$.
Infimizing over $\bA$ then gives the claimed convex
reformulation of MRO.

By \cref{lem:regret-projection-formula,lem:W2-Bures-equivalence},
the worst-case regret in the $\eps$-neighborhood of
$\tilde{\bSigma}$ is
\begin{align*}
    F(\bA)
    &=
    \sup_{\bSigma\in\cB_\eps(\tilde{\bSigma})}
        R(\bA,\bSigma)
     =
    \sup_{\substack{
        \bS\in\R^{d\times d}\\
        \|\bS-\tilde{\bS}\|_\Frob\leq\eps
    }}
    \|(\bS_Y-\bA\bS_X)\bPi_{\bS_X}\|_\Frob^2,
\end{align*}
where $\bPi_{\bS_X}\in\R^{d\times d}$ denotes the orthogonal
projection onto the row space of $\bS_X\in\R^{p\times d}$.
This projection has rank at most $p$ and depends on $\bS_X$.
We will relax this dependence by introducing an independent
projection from the family
\begin{equation}
\label{eq:MRO-Ep-def}
    \Pi_p\defeq
    \{\bPi\in\PSD^d:\bPi^2=\bPi,\ \tr(\bPi)=p\},
\end{equation}
while retaining the constraint $\bS_X=\bS_X\bPi$.
This means that the row space of $\bS_X$ is contained in
$\range(\bPi)$, a condition satisfied by $\bPi_{\bS_X}$
by definition.

For $\bA\in\cA$, a squared perturbation budget $t>0$, and
$\bPi\in\Pi_p$, define
\begin{align}
    f_\bPi(\bA,t)
    &\defeq
    \sup_{\substack{
        \bS\in\R^{d\times d}\\
        \bS_X=\bS_X\bPi\\
        \|\bS-\tilde{\bS}\|_\Frob^2\leq t
    }}
    \|(\bS_Y-\bA\bS_X)\bPi\|_\Frob^2,\label{eq:MRO-f-Pi-def}\\
    V_\bA(t)
    &\defeq
    \sup_{\bPi\in\Pi_p}f_\bPi(\bA,t).\label{eq:MRO-V-def}
\end{align}
Every $\bPi_{\bS_X}$ is dominated in Loewner order by some
$\bPi\in\Pi_p$.
For such an extension, $\bS_X=\bS_X\bPi$, and the projected
squared residual can only increase.
Consequently,
\[
    F(\bA)\leq V_\bA(\eps^2).
\]

The remainder of the proof proceeds in three steps.
\begin{enumerate}
    \item \textbf{Fixed-projection duality.}
    Consider the Lagrangian reformulation
    \begin{equation}
        f_\bPi(\bA,t)
        =
        \sup_{\substack{
            \bS\in\R^{d\times d}\\
            \bS_X=\bS_X\bPi
        }}
        \inf_{\eta\geq0}
        \left\{
            \|(\bS_Y-\bA\bS_X)\bPi\|_\Frob^2
            +\eta\bigl(t-\|\bS-\tilde{\bS}\|_\Frob^2\bigr)
        \right\}.
    \end{equation}
    We first prove that the supremum and infimum can be exchanged.
    Specifically, for $\eta\geq0$, define the fixed-projection
    Lagrangian envelope
    \begin{equation}
    \label{eq:MRO-relaxed-envelope-definition}
        \Lambda_\bPi(\bA,t,\eta)
        \defeq
        \sup_{\substack{
            \bS\in\R^{d\times d}\\
            \bS_X=\bS_X\bPi
        }}
        \left\{
            \|(\bS_Y-\bA\bS_X)\bPi\|_\Frob^2
            +\eta\bigl(t-\|\bS-\tilde{\bS}\|_\Frob^2\bigr)
        \right\}.
    \end{equation}
    We show that
    \[
        f_\bPi(\bA,t)
        =
        \inf_{\eta>b_\bA}\Lambda_\bPi(\bA,t,\eta)
    \]
    via a change of coordinates and \cref{lem:quadratic-ball-duality}. The same lemma also gives the closed form 
    \begin{equation}
    \label{eq:MRO-relaxed-envelope-formula}
        \Lambda_\bPi(\bA,t,\eta)
        =
        \eta\left(
            t-\|\tilde{\bS}_X\|_\Frob^2
            +\tr(\bH_{\bA,\eta}\bPi)
        \right),
        \qquad \eta>b_\bA.
    \end{equation}

    \item \textbf{Projection--multiplier minimax exchange.}
    Applying the non-convex minimax theorem
    \citet[Theorem~1.1]{ricceri2018minimax}, we obtain
    \begin{align*}
        V_\bA(t)
        &=\sup_{\bPi\in\Pi_p}\inf_{\eta>b_\bA}
          \Lambda_\bPi(\bA,t,\eta)\\
        &=\inf_{\eta>b_\bA}\sup_{\bPi\in\Pi_p}
          \Lambda_\bPi(\bA,t,\eta)\\
        &=\inf_{\eta>b_\bA}
          \eta\left(
              t-\|\tilde{\bS}_X\|_\Frob^2
              +\tr_p(\bH_{\bA,\eta})
          \right).
    \end{align*}
    The last equality uses the Ky Fan variational formula
    from \cref{lem:variational-matrix-formulas}.     At $t=\eps^2$, this matches our claimed expression for
    $F(\bA)$. It remains to show that
    $F(\bA)\geq V_\bA(\eps^2)$.

    \item \textbf{Exactness and joint convexity.}
    For fixed $\bA$, the final expression above reveals that $V_\bA$ is concave
    in the squared budget $t$, and its definition implies that it
    is finite for every $t>0$.
    Hence, $V_\bA$ is continuous on $(0,\infty)$.
    For $0<s<\eps^2$, fix $\bPi\in\Pi_p$ and any $\bS$
    feasible for the supremum in \eqref{eq:MRO-f-Pi-def} defining $f_\bPi(\bA,s)$.
    Using the spare perturbation budget, we approximate $\bS$
    by factors within distance $\eps$ of $\tilde{\bS}$ whose
    feature row-space projection is exactly $\bPi$.
    Their actual regrets approach
    $\|(\bS_Y-\bA\bS_X)\bPi\|_\Frob^2$.
    Supremizing over $\bS$ and then $\bPi$ gives
    \[
        V_\bA(s)\leq F(\bA)\leq V_\bA(\eps^2).
    \]
    Letting $s\uparrow\eps^2$ and using continuity proves
    $F(\bA)=V_\bA(\eps^2)$. Joint convexity follows by expressing $D$ as the supremum over $\bPi \in \Pi_p$ of the envelopes $\Lambda_\bPi(\bA,\eps^2,\eta)$, each jointly convex in $(\bA,\eta)$.
\end{enumerate}

We now establish these three steps in order.

\begin{lemma}[Fixed-projection duality]
\label{lem:MRO-fixed-projection-duality}
For every $\bA\in\cA$, $t>0$, and $\bPi\in\Pi_p$, the Lagrangian envelope $\Lambda_\bPi$
satisfies \eqref{eq:MRO-relaxed-envelope-formula} whenever
$\eta>b_\bA$. Moreover,
\begin{equation}
\label{eq:MRO-fixed-projection-duality-new}
    f_\bPi(\bA,t)
    =\inf_{\eta\geq0}\Lambda_\bPi(\bA,t,\eta)
    =\inf_{\eta>b_\bA}\Lambda_\bPi(\bA,t,\eta).
\end{equation}
\end{lemma}

\begin{proof}
Fix $\bA\in\cA$, $t>0$, and $\bPi\in\Pi_p$, and write
$\bM\defeq \begin{pmatrix}\bA&-\bI_m\end{pmatrix}$
so that $\|\bM\|_\op^2=b_\bA$.

We first reparameterize the supremum in \eqref{eq:MRO-f-Pi-def} defining $f_\bPi$ using reduced coordinates which eliminate the row-space constraint. We then apply \cref{lem:quadratic-ball-duality} to establish strong duality and derive \eqref{eq:MRO-relaxed-envelope-formula}.

\textbf{Step 1: Reduced coordinates.} 
For any $\bS\in\R^{d\times d}$ satisfying
$\bS_X=\bS_X\bPi$, feasible for the supremum in \eqref{eq:MRO-f-Pi-def}, consider substituting $\bS_Y$ with $\bS_Y \bPi + \tilde{\bS}_Y(\bI_d - \bPi)$. This preserves the objective and cannot increase $\|\bS - \tilde{\bS}\|_\Frob^2$, so it preserves feasibility. Thus, we may freely restrict the feasible set to those $\bS$ which further satisfy $\bS_Y(\bI_d - \bPi) = \tilde{\bS}_Y(\bI_d - \bPi)$. Now fix any $\bU\in\R^{d\times p}$ whose columns form an orthonormal
basis of $\range(\bPi)$, so that
$\bU^\top\bU=\bI_p$ and $\bU\bU^\top=\bPi$. Define the reduced coordinates of feasible $\bS$ by
\[
    \bW_X\defeq\bS_X\bU\in\R^{p\times p},
    \qquad
    \bW_Y\defeq\bS_Y\bU\in\R^{m\times p},
    \qquad
    \bW\defeq\begin{pmatrix}\bW_X\\\bW_Y\end{pmatrix}
        =\bS\bU\in\R^{d\times p}.
\]
On the restricted feasible set, the correspondence between $\bS$ and $\bW$ is bijective, since $\bS_X = \bS_X \bPi = \bW_X \bU^\top$, $\bS_Y \bPi = \bW_Y \bU^\top$, and $\bS_Y(\bI_d - \bPi)$ is fixed. Further define
\[
    \tilde{\bW}_X\defeq\tilde{\bS}_X\bU,
    \qquad
    \tilde{\bW}_Y\defeq\tilde{\bS}_Y\bU,
    \qquad
    \tilde{\bW}\defeq
    \begin{pmatrix}\tilde{\bW}_X\\\tilde{\bW}_Y\end{pmatrix}
    =\tilde{\bS}\bU.
\]
We then have
\begin{align*}
    \|\bS - \tilde{\bS}\|_\Frob^2 &= \|(\bS_X - \tilde{\bS}_X)\bPi\|_\Frob^2 + \|(\bS_X - \tilde{\bS}_X)(\bI_d - \bPi)\|_\Frob^2 + \|(\bS_Y - \tilde{\bS}_Y)\bPi\|_\Frob^2\\
    &= \|(\bW_X - \tilde{\bW}_X)\bU^\top\|_\Frob^2 + \|\tilde{\bS}_X(\bI_d - \bPi)\|_\Frob^2 + \|(\bW_Y - \tilde{\bW}_Y)\bU^\top\|_\Frob^2\\
    &= \|\bW - \tilde{\bW}\|_\Frob^2 + \underbrace{\|\tilde{\bS}_X(\bI_d - \bPi)\|_\Frob^2}_{\defeq c_\bPi}\,,
\end{align*}
using $\bU^\top \bU = \bI_p$. Returning to \eqref{eq:MRO-f-Pi-def}, this yields
\begin{align}
    f_\bPi(\bA,t) &= \sup_{\substack{
        \bS\in\R^{d\times d}\\
        \bS_X=\bS_X\bPi\\
        \bS_Y(\bI_d - \bPi) = \tilde{\bS}_Y(\bI_d - \bPi)\\
        \|\bS - \tilde{\bS}\|_\Frob^2 \leq t
    }} 
        \|(\bS_Y-\bA\bS_X)\bPi\|_\Frob^2
    \notag\\
    &= \sup_{\substack{
        \bW \in \R^{d \times p}\\
        \|\bW - \tilde{\bW}\|_\Frob^2 \leq t - c_\bPi
    }} 
        \|(\bW_Y-\bA\bW_X)\bU^\top\|_\Frob^2
    \notag\\
    &= \sup_{\substack{
        \bW \in \R^{d \times p}\\
        \|\bW - \tilde{\bW}\|_\Frob^2 \leq t - c_\bPi
    }} 
        \|\bW_Y-\bA\bW_X\|_\Frob^2.
    \label{eq:MRO-reduced-f-Pi}
\end{align}
Likewise, for all $\eta \geq 0$, an analogous argument gives
\begin{align}
\label{eq:MRO-reduced-Lagrangian-envelope}
    \Lambda_\bPi(\bA,t,\eta) = 
        \sup_{\bW \in \R^{d \times p}}
    \left\{
        \|\bM \bW\|_\Frob^2
        +\eta\bigl(t-c_\bPi - \|\bW - \tilde{\bW}\|_\Frob^2\bigr)
    \right\}.
\end{align}

\textbf{Step 2: Apply quadratic-ball duality.}
Since $\bW_Y-\bA\bW_X=-\bM\bW$, the reduced formulations in
\eqref{eq:MRO-reduced-f-Pi} and \eqref{eq:MRO-reduced-Lagrangian-envelope}
match the constrained and dual problems in
\cref{lem:quadratic-ball-duality}, with $k=m$, $r=p$,
center $\tilde{\bW}$, and budget parameter $t-c_\bPi$.
Since $\|\bM\|_\op^2=b_\bA$, that lemma gives
\[
    f_\bPi(\bA,t)
    =\inf_{\eta\geq0}\Lambda_\bPi(\bA,t,\eta)
    =\inf_{\eta>b_\bA}\Lambda_\bPi(\bA,t,\eta).
\]
For $\eta>b_\bA$, \eqref{eq:quadratic-ball-envelope} also gives
\[
    \Lambda_\bPi(\bA,t,\eta)
    =\eta(t-c_\bPi)
    +\eta\tr\left(
        (\eta\bI_m-\bM\bM^\top)^{-1}
        \bM\tilde{\bW}\tilde{\bW}^\top\bM^\top
    \right).
\]
Recall that
\[
    \bM\bM^\top=\bA\bA^\top+\bI_m,
    \qquad
    \bM\tilde{\bW}
    =(\bA\tilde{\bS}_X-\tilde{\bS}_Y)\bU
    =-\tilde{\bE}_\bA\bU.
\]
Since $\bU\bU^\top=\bPi$, substitution yields
\begin{equation}
\label{eq:MRO-envelope-residual-form}
    \Lambda_\bPi(\bA,t,\eta)
    =\eta(t-c_\bPi)
    +\eta\tr\left(
        ((\eta-1)\bI_m-\bA\bA^\top)^{-1}
        \tilde{\bE}_\bA\bPi\tilde{\bE}_\bA^\top
    \right),
    \qquad \eta>b_\bA.
\end{equation}
Combining \eqref{eq:MRO-envelope-residual-form} with the identity $c_\bPi=\|\tilde{\bS}_X\|_\Frob^2 -\tr(\tilde{\bS}_X^\top\tilde{\bS}_X\bPi)$, cyclicity of the trace, and the definition of $\bH_{\bA,\eta}$, we obtain \eqref{eq:MRO-relaxed-envelope-formula}.
\end{proof}

\begin{lemma}[Projection--multiplier minimax exchange]
\label{lem:MRO-projection-relaxed-duality}
For every $\bA\in\cA$ and $t>0$,
\begin{equation}
\label{eq:MRO-relaxed-dual}
\begin{aligned}
    V_\bA(t)
    &=\inf_{\eta>b_\bA}
       \sup_{\bPi\in\Pi_p}\Lambda_\bPi(\bA,t,\eta)\\
    &=\inf_{\eta>b_\bA}
      \eta\left(
          t-\|\tilde{\bS}_X\|_\Frob^2
          +\tr_p(\bH_{\bA,\eta})
      \right).
\end{aligned}
\end{equation}
\end{lemma}

\begin{proof}
Fix $\bA\in\cA$ and $t>0$, and write
\[
    I\defeq(b_\bA,\infty),
    \qquad
    I_j\defeq[b_\bA+1/j,b_\bA+j],
    \quad j\geq2.
\]
These nested compact intervals exhaust $I$.
By \cref{lem:MRO-fixed-projection-duality} and the definition of
$V_\bA$,
\[
    V_\bA(t)
    =\sup_{\bPi\in\Pi_p}\inf_{\eta\in I}
       \Lambda_\bPi(\bA,t,\eta).
\]
We now apply the minimax theorem of
\citet[Theorem~1.1]{ricceri2018minimax} to the joint objective 
$g(\bPi,\eta)\defeq-\Lambda_\bPi(\bA,t,\eta)$, with exhaustion sequence $\{I_j\}_{j \geq 2}$. This requires checking the following assumptions.

\textbf{Compactness, continuity, and convexity.} The set $\Pi_p$ is closed, since its defining constraints in \eqref{eq:MRO-Ep-def} are continuous. Moreover, $\|\bPi\|_\Frob^2=\tr(\bPi)=p$ for each $\bPi\in\Pi_p$,
so it is in fact compact. This implies that each $\bPi\mapsto\inf_{\eta\in I_j}g(\bPi,\eta)$ is
relatively inf-compact (see the definition in \citealp{ricceri2018minimax}) because the domain $\Pi_p$ is compact.
Equation \eqref{eq:MRO-relaxed-envelope-formula}, established in
\cref{lem:MRO-fixed-projection-duality}, shows that
$(\bPi,\eta)\mapsto\Lambda_\bPi(\bA,t,\eta)$ is jointly
continuous on $\Pi_p\times I$.
Finally, for fixed $\bPi$, $\eta \mapsto \Lambda_\bPi(\bA,t,\eta)$ is convex
as a supremum of affine functions in
\eqref{eq:MRO-relaxed-envelope-definition}. 

\textbf{Connectedness of best-response sets.} Fixing $\eta\in I$, we next prove that the set of projections $\bPi \in \Pi_p$
maximizing $\Lambda_\bPi(\bA,t,\eta)$ is connected. 
Since $\eta>0$, \eqref{eq:MRO-relaxed-envelope-formula} gives
\[
    \argmax_{\bPi\in\Pi_p}\Lambda_\bPi(\bA,t,\eta)
    =\argmax_{\bPi\in\Pi_p}\tr(\bH_{\bA,\eta}\bPi).
\]
List the eigenvalues of $\bH_{\bA,\eta} \in \PSD^d$ in decreasing order, counted with
multiplicity, and let $\alpha$ be the $p$th eigenvalue.
Let $\bP_>$ project onto the direct sum of eigenspaces with
eigenvalues strictly greater than $\alpha$, set
$r_>\defeq\rank(\bP_>)<p$, and write
$\mathcal V\defeq\vnull(\bH_{\bA,\eta}-\alpha\bI_d)$.
The equality conditions in the Ky Fan principle
(\cref{lem:variational-matrix-formulas}) show that every
maximizing projection is $\bP_>+\bQ$, where $\bQ$ projects
onto an $r$-dimensional subspace of $\mathcal V$, with
$r\defeq p-r_>$.

Set $n\defeq\dim\mathcal V$.
If $r=n$, the maximizing set is a singleton.
Otherwise, choose $\bV\in\R^{d\times n}$ whose columns form an
orthonormal basis of $\mathcal V$, and set
$\bQ_0\defeq\diag(\bI_r,\bZero_{n-r})$.
The maximizing set is then
\[
    \left\{
        \bP_>+\bV\bO\bQ_0\bO^\top\bV^\top:
        \bO\in\OO^n,\ \det(\bO)=1
    \right\}.
\]
Indeed, allowing all $\bO\in\OO^n$ generates every
projection onto an $r$-dimensional subspace of $\mathcal V$.
The determinant restriction loses none of them, since negating one
column of $\bO$ changes its determinant sign but leaves
$\bO\bQ_0\bO^\top$ unchanged because $\bQ_0$ is diagonal.
The displayed set is therefore a continuous image of the
special orthogonal group in dimension $n$, which is path-connected.
Hence the maximizing set is connected.

Lastly, fixing $\bPi \in \Pi_p$ and $j \geq 2$, we must show that the set of $\eta \in I_j$ minimizing $\Lambda_\bPi(\bA,t,\eta)$ is nonempty and connected. This follows directly by continuity and convexity of the objective in $\eta$.\medskip

To conclude, Ricceri's theorem then gives
\[
    V_\bA(t)
    =\inf_{\eta>b_\bA}\sup_{\bPi\in\Pi_p}
       \Lambda_\bPi(\bA,t,\eta).
\]
Substituting \eqref{eq:MRO-relaxed-envelope-formula} and applying
\cref{lem:variational-matrix-formulas} gives
\[
    \sup_{\bPi\in\Pi_p}\Lambda_\bPi(\bA,t,\eta)
    =\eta\left(
        t-\|\tilde{\bS}_X\|_\Frob^2
        +\tr_p(\bH_{\bA,\eta})
    \right),
\]
which proves the second equality in \eqref{eq:MRO-relaxed-dual}.
\end{proof}

\begin{lemma}[Exactness and joint convexity]
\label{lem:MRO-exactness-convexity}
For each $\bA\in\cA$, the function $t \mapsto V_\bA(t)$ is
finite, concave, and continuous on $(0,\infty)$, and
\[
    F(\bA)=V_\bA(\eps^2).
\]
Moreover, $D$, extended by $+\infty$ outside
$\eta>b_\bA$, is jointly convex on $\cA\times\R$.
\end{lemma}

\begin{proof}
We prove the claims separately.

\textbf{Finiteness, concavity, and continuity.}
Fixing $\bA\in\cA$, the supremum in \eqref{eq:MRO-V-def} defining $V_\bA(t)$ is feasible for each $t>0$. Indeed, we may take $\bS=\tilde{\bS}$ and let $\bPi\in\Pi_p$ project onto a
$p$-dimensional subspace containing the row space of
$\tilde{\bS}_X$. Moreover, since each $\bPi \in \Pi_p$ is dominated in Loewner order by $\bI_d$, we have
\begin{align*}
    0 &\leq V_\bA(t)\\
    &\leq \sup_{\substack{\bS \in \R^{d \times d}\\
    \|\bS - \tilde{\bS}\|_\Frob^2 \leq t}} \|\bS_Y - \bA \bS_X\|_\Frob^2\\
    &\leq b_\bA \sup_{\substack{\bS \in \R^{d \times d}\\
    \|\bS - \tilde{\bS}\|_\Frob^2 \leq t}} \|\bS\|_\Frob^2\tag{\cref{lem:basic-matrix-norms}}\\
    &\leq b_\bA\bigl(\|\tilde{\bS}\|_\Frob+\sqrt t\bigr)^2<\infty.
\end{align*}
By \cref{lem:MRO-projection-relaxed-duality},
$t\mapsto V_\bA(t)$ is the infimum of affine functions of $t$,
hence concave.
Being finite and concave on the open interval $(0,\infty)$, this function is
continuous there.

\textbf{Exactness of the projection relaxation.}
The inequality $F(\bA)\leq V_\bA(\eps^2)$ was established above.
For the reverse inequality, fix $0<s<\eps^2$ and a pair
$(\bPi,\bS) \in \Pi_p \times \R^{d \times d}$ satisfying
\[
    \bS_X=\bS_X\bPi,\qquad
    \|\bS-\tilde{\bS}\|_\Frob^2\leq s.
\]
Now choose $\bU\in\R^{d\times p}$ with
$\bU^\top\bU=\bI_p$ and $\bPi=\bU\bU^\top$.
Since $\bS_X=\bS_X\bPi$, we may write
$\bS_X=\bC\bU^\top$ with $\bC\defeq\bS_X\bU\in\R^{p\times p}$.
Then, for $\tau>0$, define
\[
    \bS_X^{(\tau)} \defeq (\bC+\tau\bI_p)\bU^\top,
    \qquad
    \bS^{(\tau)} \defeq
    \begin{pmatrix}\bS_X^{(\tau)}\\\bS_Y\end{pmatrix}.
\]
The polynomial $\det(\bC+\tau\bI_p)$ has leading coefficient $1$,
so it is nonzero for sufficiently small $\tau > 0$.
For such $\tau$, $\bS_X^{(\tau)}$ has full row rank and
$\bPi_{\bS_X^{(\tau)}}=\bPi$.
Moreover,
\[
    \|\bS^{(\tau)}-\tilde{\bS}\|_\Frob
    \leq\|\bS-\tilde{\bS}\|_\Frob
       +\|\bS^{(\tau)}-\bS\|_\Frob
    \leq\sqrt s+\tau\sqrt p<\eps
\]
for sufficiently small $\tau$.
By \cref{lem:W2-Bures-equivalence},
$\bSigma^{(\tau)}\defeq\bS^{(\tau)}(\bS^{(\tau)})^\top$ thus
belongs to $\cB_\eps(\tilde{\bSigma})$ for sufficiently small $\tau$. Hence, \cref{lem:regret-projection-formula} gives
\begin{align*}
    F(\bA)
    &\geq \lim_{\tau \downarrow 0} R(\bA,\bSigma^{(\tau)}) =  \lim_{\tau \downarrow 0}\|(\bS_Y-\bA\bS_X^{(\tau)})\bPi\|_\Frob^2 =
      \|(\bS_Y-\bA\bS_X)\bPi\|_\Frob^2.
\end{align*}
Taking the supremum over $(\bPi,\bS)$ gives
\[
    V_\bA(s)\leq F(\bA)\leq V_\bA(\eps^2).
\]
Letting $s\uparrow\eps^2$ and using continuity proves
$F(\bA)=V_\bA(\eps^2)$.

\textbf{Joint convexity.}
For fixed $\bS,\bPi$, the expression
\[
    \|(\bS_Y-\bA\bS_X)\bPi\|_\Frob^2
    +\eta\bigl(\eps^2-\|\bS-\tilde{\bS}\|_\Frob^2\bigr)
\]
is convex in $\bA$ and affine in $\eta$, with no mixed term.
It is therefore jointly convex in $(\bA,\eta)$.
The constraint $\bS_X=\bS_X\bPi$ is independent of these variables,
so taking the supremum in
\eqref{eq:MRO-relaxed-envelope-definition} preserves joint convexity.
Thus $\Lambda_\bPi(\cdot,\eps^2,\cdot)$ is jointly convex
on $\cA\times[0,\infty)$ for every fixed $\bPi$.
By \eqref{eq:MRO-relaxed-envelope-formula} and the Ky Fan formula,
\[
    D(\bA,\eta)
    =\sup_{\bPi\in\Pi_p}\Lambda_\bPi(\bA,\eps^2,\eta)
    \qquad\text{when }\eta>b_\bA.
\]
Hence $D$ is jointly convex on this domain.
The domain itself is convex, being the strict epigraph of
$\bA\mapsto1+\|\bA\|_\op^2$.
Extending $D$ by $+\infty$ outside this domain preserves convexity.
\end{proof}

\begin{proof}[Proof of \cref{thm:MRO-convex-reformulation}]
By \cref{lem:MRO-projection-relaxed-duality,lem:MRO-exactness-convexity},
\[
    F(\bA)=V_\bA(\eps^2)
    =\inf_{\eta>b_\bA}D(\bA,\eta)
    \qquad\forall\,\bA\in\cA,
\]
and $D$ is jointly convex as an extended-value function.
Taking the infimum over $\bA$ gives
\[
    \inf_{\bA\in\cA}F(\bA)
    =\inf_{\bA\in\cA,\ \eta\in\R}D(\bA,\eta),
\]
as claimed.
\end{proof}

\subsection{Subgradient Method Convergence for SNR (Proof of \cref{prop:SNR-subgradient-convergence})}
\label{prf:SNR-subgradient-convergence}

If $\tilde{\bSigma}_Y = \bZero$, then $G(\bZero) = 0$. Since $G$ is always non-negative, $\bZero$ is a minimizer, as claimed.

Otherwise, recall that
\begin{align*}
    B &\defeq (d \tr \tilde{\bSigma}_Y)^{1/2}/\eps\\
    M &\defeq 2(B \|\tilde{\bSigma}_X\|_\op +  \|\tilde{\bSigma}_{YX}\|_\Frob +  \eps^2 B)\\
    T &\defeq \lceil M^2 B^2 \delta^{-2} \rceil.
\end{align*}

First, we note a simple bound on the norm of a minimizer of $G(\bA) = L(\bA,\tilde{\bSigma}) + \eps^2\|\bA\|_\op^2$.

\begin{lemma}[SNR solution bound]
\label{lem:SNR-solution-bd}
Each minimizer $\bA_\star$ of $G$ satisfies $\|\bA_\star\|_\Frob \leq B$.
\end{lemma}
\begin{proof}
By optimality, we have
\begin{equation*}
    \eps^2 \|\bA_\star\|_\op^2 \leq G(\bA_\star) \leq G(\bZero) = \tr(\tilde{\bSigma}_Y),
\end{equation*}
and so $\|\bA_\star\|_\Frob \leq \sqrt{d}\|\bA_\star\|_\op \leq (d\tr(\tilde{\bSigma}_Y))^{1/2}/\eps = B$.
\end{proof}

Thus, writing $\cK$ for the Frobenius ball of radius $B$ centered at $\bA_1 = \bZero$, we have $G(\bA_\star) = \min_{\bA \in \cK} G(\bA)$. Moreover, the projection operator appearing in the proposition is exactly the Frobenius projection onto $\cK$.

Next, we claim that $G$ is $M$-Lipschitz under the Frobenius norm over the domain $\cK$. Indeed, we can write the subdifferential set $\partial G(\bA)$ as the convex hull
\begin{equation*}
    \conv \left\{2\bA \tilde{\bSigma}_X - 2 \tilde{\bSigma}_{YX} + 2 \eps^2 \|\bA\|_\op uv^\top : u \in \R^m, v \in \R^p, \|u\| = \|v\| = 1, \bA v = \|\bA\|_\op u \right\},
\end{equation*}
Here, the operator-norm subgradient is handled via the characterization of the spectral-norm subdifferential set in Example 2.8 of \citet{lewis1995convex}. Thus the Lipschitz constant of $G$ over $\cK$ is at most
\begin{align*}
    \sup_{\substack{\bA \in \cK\\ u \in \R^m, \|u\| \leq 1\\
    v \in \R^p, \|v\| \leq 1}} \left\|2\bA \tilde{\bSigma}_X - 2 \tilde{\bSigma}_{YX} + 2 \eps^2 \|\bA\|_\op u v^\top\right\|_\Frob \leq M,
\end{align*}
using the triangle inequality and \cref{lem:basic-matrix-norms}.

Noting that the Frobenius geometry is Euclidean, \cref{prop:SNR-subgradient-convergence} then follows by standard guarantees for the projected subgradient method (e.g., Theorem 3.2 of \citealp{Bubeck_2015_Convex}).\qed

\subsection{FISTA Convergence for SNR (Proof of \cref{prop:SNR-FISTA-convergence})}
\label{prf:SNR-FISTA-convergence}

First, we observe that \cref{alg:prox-spec-sq}, given input $\bB$ and parameter $\alpha$, evaluates the proximal operator of $\alpha \|\cdot\|_\op^2$ at $\bB$. In other words, it solves the minimization problem
\begin{equation*}
    \min_{\bC \in \R^{m \times p}} \alpha\|\bC\|_\op^2 + \frac{1}{2}\|\bC - \bB\|_\Frob^2.
\end{equation*}

Indeed, by the discussion of orthogonally invariant matrix functions in \citet[Sections~6.7.2--6.7.3]{Parikh_2014_Proximal}, the solution to the proximal problem above can be obtained by first computing an SVD of $\bB$ and then applying the $\ell_\infty^2$ proximal operator to the vector of its singular values. The clipping procedure of \cref{alg:prox-spec-sq} is the squared-$\ell_\infty$ proximal operator derived in \citet[Theorem~4]{Jacobsson_2017_Precoding}.

Given that \cref{alg:prox-spec-sq} correctly implements the proximal operator for $\alpha \|\cdot\|_\op^2$, \cref{alg:SNR-FISTA} is simply a standard application of the FISTA algorithm \citep{beck2009fista}, applied to the objective $G$, decomposed as the sum of the smooth term $L(\cdot,\tilde{\bSigma})$ and the proximable term $\eps^2\|\cdot\|_\op^2$. In particular, letting $\bA_\star$ denote any minimizer of $G$, Theorem~4.4 of \citet{beck2009fista} gives
\begin{align*}
    G(\bA_T) - G(\bA_\star) \leq \frac{2M \|\bA_\star\|_\Frob^2}{(T+1)^2},
\end{align*}
using that $L(\cdot,\tilde{\bSigma})$ is $M = 2\lambda_{\max}(\tilde{\bSigma}_X)$ smooth. By \cref{lem:SNR-solution-bd}, $\|\bA_\star\|_\Frob^2 \leq d\tr(\tilde{\bSigma}_Y)/\eps^2$, and so $\bA_T$ has additive approximation error $\delta$ once $T \geq (2M d \tr(\tilde{\bSigma}_Y))^{1/2}\eps^{-1} \cdot \delta^{-1/2}$, as claimed.

\section{Remaining Proofs for \cref{sec:finite-sample}}
\label{prf:finite-sample}

\subsection{Properties of the Bures Variance Proxy
(Proof of \cref{prop:Bures-variance-properties})}
\label{prf:Bures-variance-properties}

Let $Z \sim P$ and abbreviate $\bSigma \defeq \bSigma_P$. Since $\sV_\sB(P)$ is the expectation of a quartic in the entries of $Z$, it is finite for $P \in \cP_4(\R^d)$. By non-negativity of the Frobenius norm, $\sV_\sB(P)$ vanishes if and only if 
\[
(ZZ^\top - \bSigma)\bSigma_P^{\dagger/2} = 0 \iff ZZ^\top = \bSigma
\]
almost surely. The equivalence holds because $Z$ lies in the range of $\bSigma$ almost surely.

Expanding the squared norm in the definition of $\sV_\sB$ and using
$\bSigma\bSigma^\dagger Z=Z$ almost surely gives
\[
    \left\|
        (ZZ^\top-\bSigma)\bSigma^{\dagger/2}
    \right\|_\Frob^2
    =
    \|Z\|^2Z^\top\bSigma^\dagger Z
    -2\|Z\|^2+\tr(\bSigma).
\]
Taking expectations and using
$\E_P\|Z\|^2=\tr(\bSigma)$ proves the equality in
\eqref{eq:Bures-variance-proxy}.

For $\bSigma\neq\bZero$, let $\lambda_{\min}^+(\bSigma)$
denote its smallest positive eigenvalue. The fourth-moment bound follows from the inequality
\begin{equation*}
    Z^\top\bSigma^\dagger Z \leq \frac{\|Z\|^2}{\lambda_{\min}^+(\bSigma)}
\end{equation*}
on \(\mathrm{range}(\bSigma)\), implying that
\begin{equation*}
    \sV_{\sB}(P) \leq \E_P\bigl[\|Z\|^2 Z^\top \bSigma^\dagger Z\bigr]
    \leq \frac{\E\|Z\|^4}{\lambda_{\min}^+(\bSigma)}.
\end{equation*}
This is finite whenever $P \in \cP_4(\R^d)$ and $\bSigma \neq 0$. The $\bSigma=0$ case requires $Z=0$ almost surely and hence $\sV_{\sB}(P)=0$. When $\bSigma\succ\bZero$,
$\lambda_{\min}^+(\bSigma)=\lambda_{\min}(\bSigma)$,
giving the stated bound.

Now suppose that $P = \cN(0,\bSigma)$ is Gaussian, and assume without loss of generality that $\bSigma$ is diagonalizable in the standard basis, with eigenvalues $\lambda_1 \geq \lambda_2 \geq \dots \geq \lambda_d$. Restricting to $\range(\bSigma)$, let $r \defeq \rank(\bSigma)$ and write
\begin{equation*}
    Z=\sum_{j=1}^r \sqrt{\lambda_j}G_j e_j,
    \qquad
    G_j\overset{\mathrm{i.i.d.}}{\sim}\cN(0,1).
\end{equation*}
This gives
\begin{equation*}
    Z^\top\bSigma^\dagger Z =
    \sum_{j=1}^r G_j^2, \qquad \|Z\|^2 = \sum_{j=1}^r \lambda_j G_j^2,
\end{equation*}
and so we compute
\begin{align*}
    \E_P\bigl[\|Z\|^2 Z^\top \bSigma^\dagger Z\bigr]
    &=
    \sum_{i=1}^r \lambda_i
    \E\left[G_i^2 \sum_{j=1}^r G_j^2\right]\\
    &=
    \sum_{i=1}^r \lambda_i \bigl(3+(r-1)\bigr)\\
    &=
    (r+2)\tr(\bSigma).
\end{align*}
Therefore, $\sV_{\sB}(P) = (r+2)\tr(\bSigma)-\tr(\bSigma) = (r+1)\tr(\bSigma)$.

If further $P \in \cP^*$ is non-degenerate, we then have $r = \rank(\bSigma) \geq \rank(\bSigma_X) = p$. Moreover,
\begin{align*}
    \tr(\bSigma) \geq \tr(\bSigma_Y) = \tr(\bA_P \bSigma_X \bA_P^\top) + \tr(\bSigma_E) \geq \tr(\bSigma_E).
\end{align*}
Combining with the above gives $\sV_\sB(P) \geq (p+1)\tr(\bSigma_E) \geq p\tr(\bSigma_E)$, as claimed.

It remains to prove the empirical Bures bound. Since each $Z_i \in \range(\bSigma)$ almost surely, we have
$\range(\bSigma_{P_n}) \subseteq \range(\bSigma)$ almost surely. Therefore \cref{lem:Bures-quadratic-upper} gives
\begin{equation*}
    \sB(\bSigma,\bSigma_{P_n})^2 \leq \tr\bigl((\bSigma_{P_n}-\bSigma) \bSigma^\dagger (\bSigma_{P_n}-\bSigma)\bigr).
\end{equation*}
Next, set $\bH_i \defeq Z_i Z_i^\top - \bSigma$, so that $\bSigma_{P_n}-\bSigma = \sum_{i=1}^n \bH_i/n$.
The $\bH_i$ matrices are independent and mean zero, so the cross terms vanish in expectation:
\begin{align*}
    \E\tr\bigl((\bSigma_{P_n} - \bSigma)\bSigma^\dagger(\bSigma_{P_n} - \bSigma)\bigr)
    = \frac{1}{n^2}
    \sum_{i=1}^n
    \E\tr(\bH_i\bSigma^\dagger\bH_i) =
    \frac1n \sV_{\sB}(P),
\end{align*}
where the last equality uses \eqref{eq:Bures-variance-proxy}. This proves the claim.\qed

\subsection{Finite-Sample Regret Bounds
(Proof of \cref{thm:finite-sample-regret-bds})}
\label{prf:finite-sample-regret-bds}

We first establish some empirical estimates needed for the
sub-Gaussian guarantee.

\begin{lemma}[Empirical conditioning and ERM error]
\label{lem:empirical-conditioning}
Fix $P\in\cP^*$ and $\delta\in(0,1)$. Let $(X,Y) \sim P$ and $E=Y-\bA_PX$. Suppose that the random vector $((\bSigma_X^P)^{-1/2}X,(\bSigma_E^P)^{\dagger/2}E)$ is $K$-sub-Gaussian.
Then for $n\gtrsim K^4(d+\log(1/\delta))$, the following hold
simultaneously with probability at least $1-\delta$:
\begin{align*}
    \tfrac34\bSigma_X^P
    &\preceq\bSigma_X^{P_n}\preceq\tfrac54\bSigma_X^P,\\
    \kappa_{P_n}
    &\leq4\kappa_P,\\
    R(\ERM[P_n],P)
    &\lesssim
    \frac{K^4\tr(\bSigma_E^P)(d+\log(1/\delta))}{n}.
\end{align*}
In particular, $P_n\in\cP^*$ on this event.
\end{lemma}
\begin{proof}
Define the random vectors
\[
    \bar X\defeq(\bSigma_X^P)^{-1/2}X,
    \qquad
    \bar E\defeq(\bSigma_E^P)^{\dagger/2}E,
    \qquad
    W\defeq(\bar X,\bar E).
\]
We will consider both $P$ and $P_n$ as their base probability law. Let $\bPi_E$ project onto $\range(\bSigma_E^P)$. Then $E\in\range(\bSigma_E^P)$ almost surely, and the population normal equation gives $\E_P[EX^\top]=\bZero$. Therefore,
\[
    \bSigma_W^P\defeq\E_P[WW^\top]
    =\diag(\bI_p,\bPi_E).
\]
The vector $W$ is supported on
$\R^p\times\range(\bSigma_E^P)$ and has identity second moment
on this subspace, whose dimension is at most $d$.
The feature directions have unit second moment, so $K\geq1$.
Applying \citet[Theorem~4.7.1 and Remark~4.7.3]{Vershynin_2026_Probability}
on this subspace gives an event $\cE$ of probability at least
$1-\delta$ on which
\[
    \left\|\E_{P_n}[WW^\top]-\bSigma_W^P\right\|_\op
    \leq\alpha_n
    \defeq CK^2\sqrt{\frac{d+\log(1/\delta)}n}
    \leq\frac14.
\]
Here $C$ is an absolute constant, and the sample-size assumption
absorbs the linear term in the cited concentration bound.
The second-moment formulation does not require $W$ to be centered.

Define the empirical blocks
\begin{align*}
    \bH_n&\defeq\E_{P_n}[\bar X\bar X^\top]\in\R^{p\times p},\\
    \bC_n&\defeq\E_{P_n}[\bar E\bar X^\top]\in\R^{m\times p},\\
    \bJ_n&\defeq\E_{P_n}[\bar E\bar E^\top]\in\R^{m\times m}.
\end{align*}
On $\cE$ we have
\begin{equation}
\label{eq:empirical-whitened-moment-convergence}
    \|\bH_n-\bI_p\|_\op,
    \|\bC_n\|_\op,
    \|\bJ_n-\bPi_E\|_\op
    \leq\alpha_n\leq \frac{1}{4}.
\end{equation}
Condition on this event for the remainder of the proof.
Since
\[
    \bSigma_X^{P_n}
    =(\bSigma_X^P)^{1/2}\bH_n(\bSigma_X^P)^{1/2},
\]
we obtain the claimed bounds for the empirical feature second-moment matrix, with the lower bound implying that $P_n \in \cP^*$. Also,
$\|\bH_n^{-1}\|_\op\leq(1-\alpha_n)^{-1}\leq4/3$.

The identities $Y=\bA_PX+E$ and $E=(\bSigma_E^P)^{1/2}\bar E$ imply that
\[
    \bSigma_{YX}^{P_n}
    =\bA_P\bSigma_X^{P_n}+\E_{P_n}[EX^\top] = \bA_P\bSigma_X^{P_n} + (\bSigma_E^P)^{1/2}\bC_n(\bSigma_X^P)^{1/2}.
\]
Combining the above with the formula for $\bA_{P_n}$ gives
\begin{align}
    \bA_{P_n}-\bA_P
    &= \bSigma_{YX}^{P_n}(\bSigma_X^{P_n})^{-1} - \bA_P\notag\\
    &= (\bSigma_E^P)^{1/2}\bC_n(\bSigma_X^P)^{1/2}(\bSigma_X^{P_n})^{-1}\notag\\
    &= (\bSigma_E^P)^{1/2}\bC_n\bH_n^{-1}
      (\bSigma_X^P)^{-1/2}.
\label{eq:A_P-empirical-error}
\end{align}
Consequently,
\[
    \|\bA_{P_n}-\bA_P\|_\op
    \leq\frac{\alpha_n}{1-\alpha_n}
      \sqrt{\frac{\lambda_{\max}(\bSigma_E^P)}
                 {\lambda_{\min}(\bSigma_X^P)}}
    \leq\frac13
      \sqrt{\frac{\lambda_{\max}(\bSigma_E^P)}
                 {\lambda_{\min}(\bSigma_X^P)}}.
\]

Using \eqref{eq:A_P-empirical-error}, the empirical fitted
residual can be written as
\[
    Y-\bA_{P_n}X
    = E + (\bA_P - \bA_{P_n})X = (\bSigma_E^P)^{1/2}
       (\bar E-\bC_n\bH_n^{-1}\bar X).
\]
Using the definitions of $\bC_n$, $\bH_n$ and $\bJ_n$, along with symmetry of $\bH_n^{-1}$,
we compute
\begin{align*}
    &\E_{P_n}\bigl[
        (\bar E-\bC_n\bH_n^{-1}\bar X)
        (\bar E-\bC_n\bH_n^{-1}\bar X)^\top
    \bigr]\\
    &\quad=
    \bJ_n
    -\E_{P_n}[\bar E\bar X^\top]\bH_n^{-1}\bC_n^\top
    -\bC_n\bH_n^{-1}\E_{P_n}[\bar X\bar E^\top] +\bC_n\bH_n^{-1}
        \E_{P_n}[\bar X\bar X^\top]
        \bH_n^{-1}\bC_n^\top\\
    &\quad=
    \bJ_n-2\bC_n\bH_n^{-1}\bC_n^\top
    +\bC_n\bH_n^{-1}\bH_n\bH_n^{-1}\bC_n^\top\\
    &\quad=
    \bJ_n-\bC_n\bH_n^{-1}\bC_n^\top.
\end{align*}
Consequently, we have
\begin{align*}
    \bSigma_E^{P_n}
    &=\E_{P_n}\bigl[
        (Y-\bA_{P_n}X)(Y-\bA_{P_n}X)^\top
      \bigr]\\
    &=(\bSigma_E^P)^{1/2}
       (\bJ_n-\bC_n\bH_n^{-1}\bC_n^\top)
       (\bSigma_E^P)^{1/2}\\
    &\preceq(\bSigma_E^P)^{1/2}\bJ_n(\bSigma_E^P)^{1/2}\\
    &\preceq\frac54\bSigma_E^P.
\end{align*}
The last inequality follows from
$\bJ_n\preceq\bPi_E+\alpha_n\bI_m$
and $(\bSigma_E^P)^{1/2}\bPi_E=(\bSigma_E^P)^{1/2}$. Combining the above with \cref{thm:condition-number-formula} gives
\begin{align*}
    \kappa_{P_n}
    &\leq1+\|\bA_{P_n}\|_\op^2
       +\frac{\lambda_{\max}(\bSigma_E^{P_n})}
              {\lambda_{\min}(\bSigma_X^{P_n})}\\
    &\leq1+2\|\bA_P\|_\op^2 + 2\|\bA_{P_n} - \bA_P\|_\op^2 + 
        \frac{\tfrac{5}{4}\lambda_{\max}(\bSigma_E^P)}
             {\frac{3}{4}\lambda_{\min}(\bSigma_X^P)}\\
    &\leq1+2\|\bA_P\|_\op^2 + \left(\frac{2}{9} + \frac{5}{3}\right)
        \frac{\lambda_{\max}(\bSigma_E^P)}
             {\lambda_{\min}(\bSigma_X^P)}
     \\
     &\leq4\kappa_P.
\end{align*}

Finally, \eqref{eq:A_P-empirical-error} gives
\begin{align*}
    R(\ERM[P_n],P)
    &=\|(\bA_{P_n}-\bA_P)(\bSigma_X^P)^{1/2}\|_\Frob^2\\
    &=\|(\bSigma_E^P)^{1/2}\bC_n\bH_n^{-1}\|_\Frob^2\\
    &\leq\frac{\alpha_n^2}{(1-\alpha_n)^2}\tr(\bSigma_E^P)\\
    &\lesssim
      \frac{K^4\tr(\bSigma_E^P)(d+\log(1/\delta))}{n},
\end{align*}
as claimed.
\end{proof}

We now prove the two guarantees separately.
For each, we establish a high-probability event depending only on the i.i.d.\ sample from $P$, under which the claimed bounds
hold uniformly over the admissible choices in both perturbation models.

\textbf{Fourth-moment guarantee.}
Abbreviate $\rho \defeq \eps_n$. By \cref{prop:Bures-variance-properties}, $\E[\sB(\bSigma_{P_n},\bSigma_P)^2] \leq V/n$, so  Markov's inequality gives $\sB(\bSigma_{P_n},\bSigma_P) \leq \sqrt{V/(\delta n)}$ with probability at least $1-\delta$. On this event,
\begin{equation}
\label{eq:empirical-radius}
    \sB(\bSigma_{P_{\mathrm{s},n}},\bSigma_P)
    +\sB(\bSigma_{P_{\mathrm t}},\bSigma_P) \leq \eps + \sqrt{\frac{V}{\delta n}} = \rho
\end{equation}
under either adversary model. Indeed, under the sample-corruption model, we have $P_{\mathrm t} = P$ and $\Wtwo(P_{\mathrm s,n},P_n) \leq \eps$, so the claim follows by the $\Wtwo$--Bures equivalence (\cref{lem:W2-Bures-equivalence}). Likewise, under the distribution-shift model, we have $P_{\mathrm s,n} = P_n$ and $\sB(\bSigma_{P_t},\bSigma_P) \leq \Wtwo(P_t,P) \leq \eps$.

Equation \eqref{eq:empirical-radius} implies that $\sB(\bSigma_{P_{\mathrm{s},n}},\bSigma_{P}), \sB(\bSigma_{P_{\mathrm t}},\bSigma_{P}), \sB(\bSigma_{P_{\mathrm{s},n}},\bSigma_{P_{\mathrm t}}) \leq \rho$, with the last case holding by the triangle inequality. Note that $\rho \leq 2\eps$ under our assumed sample complexity. We now proceed as in the population-limit proof. In particular, we use the equivalent second-moment formulations of
the estimators and regret stability, justified by
\cref{lem:W2-Bures-equivalence}.

For the MRO guarantee, set $\hat{\bA} \defeq \MRO_{\rho}[P_{\mathrm{s},n}]$ and bound
\begin{align*}
    R(\hat{\bA},P_{\mathrm t})
    &\leq
    \sup_{\bSigma\in\cB_{\rho}(\bSigma_{P_{\mathrm{s},n}})}
        R(\hat{\bA},\bSigma) \tag{$\sB(\bSigma_{P_{\mathrm{s},n}},\bSigma_{P_{\mathrm t}}) \leq \rho$}\\
    &\leq
    \sup_{\bSigma\in\cB_{\rho}(\bSigma_{P_{\mathrm{s},n}})}
        R(\bA_P,\bSigma) \tag{MRO optimality}\\
    &\leq
    \sup_{\bSigma\in\cB_{2\rho}(\bSigma_{P})}
        R(\bA_P,\bSigma) \tag{$\sB(\bSigma_{P_{\mathrm{s},n}},\bSigma_{P}) \leq \rho$}\\
    &\leq 4 \kappa_P(2\rho)^2 \tag{\cref{lem:regret-stability}}\\
    &\leq 64 \kappa_P \eps^2, \tag{$\rho \leq 2\eps$}
\end{align*}
as claimed.

For SNR, set $\hat{\bA} \defeq\SNR_\rho[P_{\mathrm{s},n}]$. Since risk and regret
differ by a constant independent of the coefficient matrix, SNR optimality implies
\begin{align*}
    R(\hat{\bA},P_{\mathrm{s},n})
    +\rho^2\|\hat{\bA}\|_\op^2
    &\leq
    R(\bA_P,P_{\mathrm{s},n})
    +\rho^2\|\bA_P\|_\op^2 \leq5\kappa_P\rho^2.
\end{align*}
The last inequality follows from
$\sB(\bSigma_{P_{\mathrm{s},n}},\bSigma_P)\leq\rho$,
\cref{lem:regret-stability}, and
$\|\bA_P\|_\op^2\leq\kappa_P$
from \cref{thm:condition-number-formula}. To convert this to a regret bound under $P_{\mathrm t}$, we apply regret stability
twice, using $P$ as the reference law in both applications:
\begin{align*}
    \sqrt{R(\hat{\bA},P_{\mathrm t})}
    &\leq
    \sqrt{R(\hat{\bA},P)}
    +\bigl(2\sqrt{\kappa_P}+\|\hat{\bA}\|_\op\bigr)
       \sB(\bSigma_{P_{\mathrm t}},\bSigma_P)\\
    &\leq
    \sqrt{R(\hat{\bA},P_{\mathrm{s},n})}
    +\bigl(2\sqrt{\kappa_P}+\|\hat{\bA}\|_\op\bigr)\rho,
\end{align*}
where the last inequality uses
\eqref{eq:empirical-radius}.
By Cauchy--Schwarz and the SNR objective comparison above,
\begin{align*}
    \sqrt{R(\hat{\bA},P_{\mathrm t})}
    &\leq
    \sqrt2\sqrt{
        R(\hat{\bA},P_{\mathrm{s},n})
        +\rho^2\|\hat{\bA}\|_\op^2
    }
    +2\sqrt{\kappa_P}\rho \leq6\sqrt{\kappa_P}\rho \leq 12 \sqrt{\kappa_P} \eps.
\end{align*}
Squaring gives the desired SNR bound.

Thus, both estimators incur target regret
$O(\kappa_P\eps^2)$ under the specified event.
The bounds hold uniformly over all admissible sample
corruptions and population targets in the respective models,
including randomized, sample-dependent choices.
The event has probability at least $1-\delta$ under every
admissible rule, since the reference sample has law
$P^{\otimes n}$.
Taking the supremum over $\mathbb M_{\eps,n}(P)$ therefore
proves the first assertion.

\textbf{Sub-Gaussian guarantee.}
Our assumption on the sample size allows us to apply
\cref{lem:empirical-conditioning}.
Fix a realization on its event of probability at least
$1-\delta$.
Then $P_n\in\cP^*$, $\kappa_{P_n}\leq4\kappa_P$, and $\bSigma_X^P\preceq\tfrac43\bSigma_X^{P_n}$. Moreover, we have
\[
    e_n\defeq R(\bA_{P_n},P)
    \lesssim
    \frac{K^4\tr(\bSigma_E^P)(d+\log(1/\delta))}{n}
    \lesssim\eps^2,
\]
where the last inequality uses the theorem's sample-size
assumption.

We first establish a comparison that will be used in both  perturbation models.
For each $\bA\in\cA$, the regret formula and the squared
triangle inequality imply that
\begin{align}
    R(\bA,P)
    &=
    \|(\bA-\bA_P)(\bSigma_X^P)^{1/2}\|_\Frob^2\notag\\
    &\leq
    2\|(\bA-\bA_{P_n})(\bSigma_X^P)^{1/2}\|_\Frob^2
    +2e_n\notag\\
    &\leq
    \frac83
    \|(\bA-\bA_{P_n})(\bSigma_X^{P_n})^{1/2}\|_\Frob^2
    +2e_n\notag\\
    &=\frac83R(\bA,P_n)+2e_n,
    \label{eq:finite-sample-regret-transfer}
\end{align}
where the last inequality uses $\bSigma_X^P\preceq\tfrac43\bSigma_X^{P_n}$.

In the sample-corruption model, fix
$\hat{\bA}_n\in\{\MRO_\eps,\SNR_\eps\}$ and any admissible
observed source $P_{\mathrm{s},n}$.
Set $\hat{\bA}\defeq\hat{\bA}_n[P_{\mathrm{s},n}]$.
Since $\Wtwo(P_{\mathrm{s},n},P_n)\leq\eps$ and
$P_n\in\cP^*$, \cref{lem:stability-implications} gives
\[
    R(\hat{\bA},P_n)
    \leq R_\eps(\hat{\bA}_n,P_n)
    \lesssim\kappa_{P_n}\eps^2
    \lesssim\kappa_P\eps^2.
\]
Applying \eqref{eq:finite-sample-regret-transfer}, we obtain
\[
    R(\hat{\bA},P_{\mathrm t})
    =R(\hat{\bA},P)
    \leq\frac83R(\hat{\bA},P_n)+2e_n
    \lesssim\kappa_P\eps^2,
\]
using $e_n\lesssim\eps^2$ and $\kappa_P\geq1$.
This proves the desired target regret bound uniformly over admissible corruptions.

In the distribution-shift model, $P_{\mathrm{s},n}=P_n$
and $P_{\mathrm t}=Q\in\cB_\eps(P)$. Since $Q$ need not belong to $\cB_\eps(P_n)$, we will
first transfer regret from $P_n$ to $P$, then from $P$ to $Q$.
For the second transfer, we also need to control the
coefficient norm.

For $\hat{\bA}=\SNR_\eps[P_n]$, comparison with the empirical
risk minimizer $\bA_{P_n}$ gives
\[
    R(\hat{\bA},P_n)+\eps^2\|\hat{\bA}\|_\op^2
    \leq
    R(\bA_{P_n},P_n)+\eps^2\|\bA_{P_n}\|_\op^2
    =
    \eps^2\|\bA_{P_n}\|_\op^2
    \leq\kappa_{P_n}\eps^2.
\]
Here the first inequality uses SNR optimality, and the last
uses \cref{thm:condition-number-formula}.
Since both terms on the left are nonnegative, this bounds
the empirical regret and coefficient norm separately.

For MRO, define the objective $F_n(\bA)
    \defeq
    \sup_{Q'\in\cB_\eps(P_n)}R(\bA,Q')$. We first show that
\begin{equation}
\label{eq:finite-sample-MRO-norm-control}
    F_n(\bA)\geq\eps^2(1+\|\bA\|_\op^2)
    \qquad\forall\,\bA\in\cA.
\end{equation}
By the $\Wtwo$--Bures equivalence (\cref{lem:W2-Bures-equivalence}), we may apply
\cref{thm:MRO-convex-reformulation} with observed second
moment $\bSigma_{P_n}$.
Define the square-root factor $\bS_n\defeq\bSigma_{P_n}^{1/2}$ and let
$\bS_{n,X}\in\R^{p\times d}$ be its feature row block.
For every $\eta>1+\|\bA\|_\op^2$, the matrix inside the
$\tr_p$ term in that theorem is
$\bS_{n,X}^\top\bS_{n,X}$ plus a positive semidefinite matrix.
The $\tr_p$ term is therefore at least
\[
    \tr_p(\bS_{n,X}^\top\bS_{n,X})
    =
    \tr(\bS_{n,X}^\top\bS_{n,X})
    =
    \|\bS_{n,X}\|_\Frob^2,
\]
where the first equality uses
$\rank(\bS_{n,X}^\top\bS_{n,X})\leq p$.
Consequently, the dual objective is at least $\eta\eps^2$,
and the reformulation gives
\[
    F_n(\bA)
    \geq
    \inf_{\eta>1+\|\bA\|_\op^2}\eta\eps^2
    =
    \eps^2(1+\|\bA\|_\op^2).
\]
This proves \eqref{eq:finite-sample-MRO-norm-control}.

Now take $\hat{\bA}=\MRO_\eps[P_n]$.
Since $P_n\in\cB_\eps(P_n)$, MRO optimality gives
\begin{align*}
    R(\hat{\bA},P_n) \leq F_n(\hat{\bA}) \leq F_n(\bA_{P_n}) \leq4\kappa_{P_n}\eps^2.
\end{align*}
The last inequality follows from \cref{lem:regret-stability}
with reference law $P_n$ and coefficient $\bA_{P_n}$.
Combining this upper bound with
\eqref{eq:finite-sample-MRO-norm-control} also gives
\[
    \eps^2(1+\|\hat{\bA}\|_\op^2)
    \leq F_n(\hat{\bA})
    \leq4\kappa_{P_n}\eps^2,
\]
and hence $\|\hat{\bA}\|_\op^2\leq4\kappa_{P_n}$.

Thus, for either choice of
$\hat{\bA}_n\in\{\MRO_\eps,\SNR_\eps\}$ and
$\hat{\bA}\defeq\hat{\bA}_n[P_n]$, we have
\[
    R(\hat{\bA},P_n)\lesssim\kappa_P\eps^2,
    \qquad
    \|\hat{\bA}\|_\op^2\lesssim\kappa_P,
\]
using $\kappa_{P_n}\leq4\kappa_P$.
The empirical-to-population comparison
\eqref{eq:finite-sample-regret-transfer} thus yields
\[
    R(\hat{\bA},P)
    \leq\tfrac83R(\hat{\bA},P_n)+2e_n
    \lesssim\kappa_P\eps^2.
\]
Finally, for every $Q\in\cB_\eps(P)$,
\cref{lem:regret-stability} gives
\begin{align*}
    \sqrt{R(\hat{\bA},Q)}
    &\leq
    \sqrt{R(\hat{\bA},P)}
    +\bigl(2\sqrt{\kappa_P}+\|\hat{\bA}\|_\op\bigr)
       \Wtwo(P,Q)\\
    &\leq
    \sqrt{R(\hat{\bA},P)}
    +\bigl(2\sqrt{\kappa_P}+\|\hat{\bA}\|_\op\bigr)\eps\\
    &\lesssim\sqrt{\kappa_P}\eps.
\end{align*}
Squaring proves that
$R(\hat{\bA},P_{\mathrm t})\lesssim\kappa_P\eps^2$
in the distribution-shift model.

On the chosen sample event, both estimators satisfy the
claimed bounds for every admissible sample corruption
and every population target $Q\in\cB_\eps(P)$ in the
respective models. The event has probability at least $1-\delta$ under every
admissible rule, since the reference sample has law
$P^{\otimes n}$.
Taking the supremum over $\mathbb M_{\eps,n}(P)$ therefore
proves the second assertion.\qed

\section{Exact Reductions for Population Benchmarks in \cref{sec:experiments}}
\label{app:experiment-reductions}

In this section, we derive closed forms and exact constant-size semidefinite program (SDP) reformulations for the population benchmarks in \cref{sec:experiments}. Write $P\defeq P_{\mathrm t}$ and $\tilde P\defeq P_{\mathrm s}$ for the target and source laws, respectively, and let $\rho > 0$ denote the supplied radius, which may differ from the corruption budget $\eps=1$.

Set $q \defeq m-1$. Recall that $m=p=d/2 \geq 2$ and that our $P$ is the law of $(X,Y)$, where
\begin{equation*}
    X \sim \cN(0,\bI_m), \qquad Y = X + U, \qquad U \sim \cN\bigl(0,\diag(0,\bI_q)\bigr),
\end{equation*}
and $X$ and $U$ are independent. For $\alpha,\beta\in\R$, consider the diagonal coefficient matrix given by
\begin{equation*}
    \bA(\alpha,\beta) \defeq \diag(\alpha,\beta \bI_q).
\end{equation*}
With this notation, our source measure $\tilde{P}$ is the law of $(\bA(c,1)X,Y)$, where $c \defeq m^{-1/4}$. Since $\bA_P = \bI_m$ and $\bSigma_X^P = \bI_m$, the target regret of $\bA(\alpha,\beta)$ is given by
\begin{equation}
\label{eq:experiment-clean-regret}
    R(\bA(\alpha,\beta),P)
    =(\alpha-1)^2+q(\beta-1)^2.
\end{equation}
Moreover, the source risk is
\begin{equation}
\label{eq:experiment-corrupted-risk}
    L\bigl(\bA(\alpha,\beta),\tilde P\bigr) = (c\alpha-1)^2 + q\bigl((\beta-1)^2+1\bigr).
\end{equation}

Finally, since $\bSigma_X^{\tilde{P}}=\diag(c^2,\bI_q)\succ\bZero$,
the population ERM solution is unique and satisfies
\[
    \ERM[\tilde{P}]=\bA(c^{-1},1),
    \qquad
    R(\ERM[\tilde{P}],P)=(c^{-1}-1)^2.
\]

\subsection{Reduction to Scalar Optimization Using Rotational Symmetry}
\label{ssec:reduction-symmetry}

As in \cref{lem:DRO-LB-A}, both $P$ and $\tilde P$
are invariant under the compact group of orthogonal transformations 
\[
    \left\{
        \diag(1,\bH,1,\bH):\bH\in\OO^q
    \right\}.
\]
Its invariant coefficient matrices (in the sense of \cref{lem:DRO-symmetry}) are exactly those of the form
$\bA(\alpha,\beta)$ for some $\alpha,\beta \in \R$.
Indeed, taking $\bH=-\bI_q$ eliminates the off-diagonal
blocks, and commuting with every $\bH\in\OO^q$ forces the
remaining $q\times q$ block to be a scalar multiple of $\bI_q$. Hence, by \cref{lem:DRO-symmetry}, each objective
\begin{align*}
    J_{\mathrm{DRO}}(\bA)
    &\defeq\sup_{Q\in\cB_\rho(\tilde P)}L(\bA,Q),\\
    J_{\mathrm{MRO}}(\bA)
    &\defeq\sup_{Q\in\cB_\rho(\tilde P)}R(\bA,Q),\\
    J_{\mathrm{SNR}}(\bA)
    &\defeq L(\bA,\tilde P)+\rho^2\|\bA\|_\op^2
\end{align*}
admits a minimizer of the form $\bA(\alpha,\beta)$.

\subsection{Constant-Size SDP Reformulations for DRO and MRO}

Using this symmetry reduction, we now derive SDP reformulations for the DRO and MRO problems.
Fix $\bA=\bA(\alpha,\beta)$ for now and write
$\gamma \defeq1+\max\{\alpha^2,\beta^2\}$.

Our first goal is to derive a clean dual formulation for $J_{\mathrm{DRO}}(\bA)$ and $J_{\mathrm{MRO}}(\bA)$. We then introduce epigraph variables and simplify the resulting convex programs using Schur complements and a variational formula for $\tr_p$.

\textbf{Step 1: Initial dual problems.}
A convenient square-root factor of the source second moment is
\[
    \tilde{\bS}\defeq
    \begin{pmatrix}
        \diag(c,\bI_q)&\bZero_{m\times m}\\
        \bI_m&\diag(0,\bI_q)
    \end{pmatrix},
    \qquad
    \tilde{\bS}\tilde{\bS}^\top=\bSigma_{\tilde{P}}.
\]
Set $\tilde{\bE}_{\bA} \defeq \tilde{\bS}_Y-\bA\tilde{\bS}_X$
and $\bB_{\bA,\eta} \defeq (\eta-1)\bI_m-\bA\bA^\top$.
Then, the $\Wtwo$--Bures equivalence (\cref{lem:W2-Bures-equivalence}) and the DRO and MRO dual formulas in
\eqref{eq:DRO-inner-strong-duality} and
\cref{thm:MRO-convex-reformulation} give
\begin{align*}
    J_{\mathrm{DRO}}(\bA)
    &=\inf_{\eta>\gamma}
      \left\{\eta\rho^2+
      \tr\bigl(\eta\tilde{\bE}_{\bA}^\top
        \bB_{\bA,\eta}^{-1}\tilde{\bE}_{\bA}\bigr)\right\},\\
    J_{\mathrm{MRO}}(\bA)
    &=\inf_{\eta>\gamma}\left\{
      \eta(\rho^2-\|\tilde{\bS}_X\|_\Frob^2)
      +\tr_m\bigl(\eta\tilde{\bS}_X^\top\tilde{\bS}_X
        +\eta\tilde{\bE}_{\bA}^\top
          \bB_{\bA,\eta}^{-1}\tilde{\bE}_{\bA}\bigr)
      \right\}.
\end{align*}
Here, we used $p=m$ and positive homogeneity of $\tr_m$. To unify these formulas, define
\[
    \bK_\eta\defeq
    \eta\tilde{\bS}^\top\tilde{\bS}
    +\eta\tilde{\bE}_{\bA}^\top
       \bB_{\bA,\eta}^{-1}\tilde{\bE}_{\bA}.
\]
Substituting into the above gives
\begin{align}
    J_{\mathrm{DRO}}(\bA)
    &=\inf_{\eta>\gamma}\left\{
       \eta(\rho^2-\|\tilde{\bS}\|_\Frob^2)+\tr(\bK_\eta)
       \right\},\label{eq:experiment-DRO-Gram-form}\\
    J_{\mathrm{MRO}}(\bA)
    &=\inf_{\eta>\gamma}\left\{
       \eta(\rho^2-\|\tilde{\bS}_X\|_\Frob^2)
       +\tr_m(\bK_\eta-\eta\tilde{\bS}_Y^\top\tilde{\bS}_Y)
       \right\}.\label{eq:experiment-MRO-Gram-form}
\end{align}
The matrices here are structured, due to our choice of source law, leading to further simplifications.

First, writing $\bM_{\bA} \defeq \begin{pmatrix}\bA &-\bI_m\end{pmatrix}$,
\cref{lem:rectangular-resolvent-identities} implies that
\[
\bK_\eta = \eta^2\tilde{\bS}^\top (\eta\bI_d-\bM_{\bA}^\top\bM_{\bA})^{-1}\tilde{\bS}.
\]
Next, note that the source random variables can be written as
\[
    (\tilde X_1,Y_1)=(cG_1,G_1),\qquad
    (\tilde X_i,Y_i)=(G_i,G_i+H_i),\quad i=2,\ldots,m,
\]
with independent standard Gaussian latent variables $G_i$ and $H_i$.
The corresponding factors are
\[
    \bS_1 \defeq \begin{pmatrix}c\\1\end{pmatrix},\qquad
    \bS_b \defeq\begin{pmatrix}1&0\\1&1\end{pmatrix},
\]
with the subscript ``b'' denoting the ``bulk'' of the coordinates. For a scalar coefficient $a \in \R$ and a fixed block factor
$\bS\in\R^{2\times k}$, define
\[
    \bM(a) \defeq \begin{pmatrix}a & -1\end{pmatrix},\qquad
    \bK(a,\bS,\eta)\defeq
    \eta^2\bS^\top(\eta\bI_2-\bM(a)^\top\bM(a))^{-1}\bS,
    \quad \eta>1+a^2.
\]
Write $k_1 \defeq \bK(\alpha,\bS_1,\eta)$, identifying the
$1\times1$ matrix with its entry, and
$\bK_b \defeq \bK(\beta,\bS_b,\eta)$.
In latent-coordinate order
$G_1,(G_2,H_2),\ldots,(G_m,H_m),H_1$, the matrices
$\bK_\eta$ and $\tilde{\bS}_Y^\top\tilde{\bS}_Y$ become
\[
    \diag(k_1,\underbrace{\bK_b,\ldots,\bK_b}_{q\text{ times}},0),
    \qquad
    \diag(1,\underbrace{1_2 1_2^\top,\ldots,1_2 1_2^\top}_{q\text{ times}},0),
    \quad 1_2=(1,1)^\top.
\]
The final zero comes from the unused latent coordinate $H_1$. We also compute
\[
    \|\tilde{\bS}\|_\Frob^2=c^2+1+3q,
    \qquad
    \|\tilde{\bS}_X\|_\Frob^2=c^2+q.
\]

Substituting these decompositions into
\eqref{eq:experiment-DRO-Gram-form} and \eqref{eq:experiment-MRO-Gram-form}, we arrive at
\begin{align}
    J_{\mathrm{DRO}}(\bA)
    &=\inf_{\eta>\gamma}\Bigl\{
        \eta(\rho^2-c^2-1-3q)
        +k_1+q\tr(\bK_b)
      \Bigr\},
      \label{eq:experiment-DRO-block-form}\\
    J_{\mathrm{MRO}}(\bA)
    &=\inf_{\eta>\gamma}\Bigl\{
        \eta(\rho^2-c^2-q)
        +\tr_m\Bigl(\diag\bigl(
            k_1-\eta,
            \underbrace{
                \bK_b-\eta1_2 1_2^\top,\ldots,
                \bK_b-\eta1_2 1_2^\top
            }_{q\text{ copies}}
        \bigr)\Bigr)
      \Bigr\}.
      \label{eq:experiment-MRO-block-form}
\end{align}
Here, we removed the final zero from the matrix inside $\tr_m$. This is permissible because this matrix coincides up to a permutation of rows and columns with $\bK_\eta-\eta\tilde{\bS}_Y^\top\tilde{\bS}_Y$. By the definition of $\bK_\eta$, that matrix is greater in Loewner order than $\eta \tilde{\bS}_X^\top\tilde{\bS}_X \succeq\bZero$ and is thus positive semidefinite. The remaining blocks have dimension $2m-1\geq m$, so removing the zero does not change the eigenvalue sum.

\textbf{Step 2: Introduce epigraph variables and simplify.}
We now introduce epigraph variables $t_1\geq k_1$ and $\bT_b\succeq\bK_b$. To make these lower bound constraints affine in $\bA$ and $\eta$, we use that for $\eta>1+a^2$,
a Schur complement (\cref{lem:schur-complement}) gives
\[
    \bT\succeq\bK(a,\bS,\eta)
    \quad\Longleftrightarrow\quad
    \begin{pmatrix}
        \eta\bI_2-\bM(a)^\top\bM(a)&\eta\bS\\
        \eta\bS^\top&\bT
    \end{pmatrix}\succeq\bZero.
\]
The upper-left block is quadratic in $a$.
A second Schur complement replaces this quadratic term by
an additional scalar block, yielding the affine constraint
$\cQ(a,\bS,\eta,\bT)\succeq\bZero$, where
\begin{equation*}
    \cQ(a,\bS,\eta,\bT)
    \defeq
    \begin{pmatrix}
        \eta\bI_2&\bM(a)^\top&\eta\bS\\
        \bM(a)&1&\bZero_{1\times k}\\
        \eta\bS^\top&\bZero_{k\times1}&\bT
    \end{pmatrix}\in\Sym^{k+3}.
\end{equation*}
In particular, for $\eta>1+a^2$,
\begin{equation*}
    \cQ(a,\bS,\eta,\bT)\succeq\bZero
    \quad\Longleftrightarrow\quad
    \bT\succeq\bK(a,\bS,\eta).
\end{equation*}
For general $\eta\in\R$, its Schur complement at the scalar
block $1$ also shows that
\begin{equation}
\label{eq:experiment-PSD-gamma-implication}
    \cQ(a,\bS,\eta,\bT)\succeq\bZero
    \quad\Longrightarrow\quad\eta\geq1+a^2.
\end{equation}
Define the two instances of this constraint by
\begin{align*}
    \cQ_1(\alpha,\eta,t_1)
    &\defeq\cQ(\alpha,\bS_1,\eta,t_1)\in\Sym^4,\\
    \cQ_b(\beta,\eta,\bT_b)
    &\defeq\cQ(\beta,\bS_b,\eta,\bT_b)\in\Sym^5.
\end{align*}

Retaining the constraint that $\eta>\gamma$ in \eqref{eq:experiment-DRO-block-form} and \eqref{eq:experiment-MRO-block-form}, we replace $k_1,\bK_b$ by the epigraph variables
$t_1,\bT_b$ and minimize over them subject to
$\cQ_1(\alpha,\eta,t_1)\succeq\bZero$ and
$\cQ_b(\beta,\eta,\bT_b)\succeq\bZero$.
Both traces are nondecreasing in Loewner order, and equality
is feasible at $t_1=k_1$, $\bT_b=\bK_b$.
Thus this substitution preserves the values of
\eqref{eq:experiment-DRO-block-form} and
\eqref{eq:experiment-MRO-block-form}. For MRO, we additionally reparameterize using variables
\[
    z_1 \defeq t_1-\eta,\qquad
    \bZ_b \defeq \bT_b-\eta1_2 1_2^\top,
    \qquad
    \bZ \defeq \diag(z_1,\underbrace{\bZ_b,\ldots,\bZ_b}_{q\text{ times}}).
\]
By \cref{lem:variational-matrix-formulas}, we have
\begin{equation}
\label{eq:experiment-KyFan-epigraph}
    \tr_m(\bZ)=
    \min_{\substack{
        \tau\in\R,\ y_1\geq0,\ y_1\geq z_1-\tau\\
        \bY_b\in\Sym^2,\ \bY_b\succeq\bZero,\
        \bY_b\succeq\bZ_b-\tau\bI_2
    }} m\tau+y_1+q\tr(\bY_b).
\end{equation}

Substituting these representations into
\eqref{eq:experiment-DRO-block-form} and
\eqref{eq:experiment-MRO-block-form} and taking the infimum
over $(\alpha,\beta)$ results in the programs below with the additional constraint that $\eta > \gamma$.
We show next that this constraint
can be dropped and that the resulting minima are attained.

\begin{proposition}[Constant-size SDP formulations]
\label{prop:experiment-constant-size-conic}
A population DRO optimizer is obtained by extracting coefficient
variables $(\alpha,\beta)$ from any solution to
\begin{equation}
\label{eq:experiment-DRO-SDP}
\begin{aligned}
    \min_{\alpha,\beta,\eta,t_1,\bT_b}\quad
    &t_1+q\tr(\bT_b)+\eta(\rho^2-c^2-1-3q)\\
    \mathrm{s.t.}\quad
    &\cQ_1(\alpha,\eta,t_1)\succeq\bZero,\\
    &\cQ_b(\beta,\eta,\bT_b)\succeq\bZero,\\
    &\eta\geq0,\qquad\bT_b\in\Sym^2.
\end{aligned}
\end{equation}
Likewise, a population MRO optimizer is obtained by taking $(\alpha,\beta)$ from any solution to
\begin{equation}
\label{eq:experiment-MRO-SDP}
\begin{aligned}
    \min_{\alpha,\beta,\eta,t_1,\bT_b,\tau,y_1,\bY_b}\quad
    &\eta(\rho^2-c^2-q)+m\tau+y_1+q\tr(\bY_b)\\
    \mathrm{s.t.}\quad
    &\cQ_1(\alpha,\eta,t_1)\succeq\bZero,\\
    &\cQ_b(\beta,\eta,\bT_b)\succeq\bZero,\\
    &\eta\geq0,\\
    &y_1\geq0,\qquad y_1\geq z_1-\tau,\\
    &\bY_b\succeq\bZero,\qquad
      \bY_b\succeq\bZ_b-\tau\bI_2,\\
    &\bT_b,\bY_b\in\Sym^2.
\end{aligned}
\end{equation}
Both minima are attained, and the regret of the
returned coefficient pairs at the target law $P$ can be computed via
\eqref{eq:experiment-clean-regret}.
\end{proposition}

\begin{proof}
By \eqref{eq:experiment-PSD-gamma-implication}, both SDPs
impose $\eta\geq\gamma = \gamma(\alpha,\beta)$.
The preceding argument establishes the desired optimal value
identities with the additional restriction $\eta>\gamma$. It remains to remove this strict constraint and prove that minimizers are attained.

For fixed $(\alpha,\beta)$, let $\eta > \gamma$. By taking the remaining variables sufficiently large, we can obtain a feasible point with this strictly feasible value of $\eta$.
Moreover, the feasible set is convex, and the objective is continuous. Thus, the presence of the strict constraint does not affect the infimum.

For attainment, fix an optimal coefficient pair $(\alpha,\beta)$, which exists by
\cref{ssec:reduction-symmetry}. Choose a feasible objective value $C$ for either SDP in the statement, and restrict to feasible points with objective
at most $C$. This does not change the infimum.
We show that the restricted feasible set is compact.

First, $\eta \geq \gamma \geq 1$. Extracting principal submatrices from the $\cQ_1,\cQ_b \succeq \bZero$ constraints gives
\[
    \begin{pmatrix}
        \eta\bI_2 & \eta\bS_1\\
        \eta\bS_1^\top & t_1
    \end{pmatrix}\succeq\bZero, \qquad \begin{pmatrix}
        \eta\bI_2 & \eta\bS_b\\
        \eta\bS^\top_b & \bT_b
    \end{pmatrix}\succeq\bZero.
\]
Since $\eta>0$, \cref{lem:schur-complement} implies that
\[
    t_1\geq \eta \bS_1^\top \bS_1 = \eta(c^2+1),
    \qquad
    \bT_b\succeq\eta\bS_b^\top\bS_b = \eta \begin{pmatrix}
        2 & 1\\
        1 & 1
    \end{pmatrix}.
\]
For MRO, we therefore have
\begin{align*}
    z_1 = t_1 - \eta \geq \eta c^2, \qquad \bZ_b = \bT_b - \eta 1_2 1_2^\top \succeq \eta \begin{pmatrix}
        1 & 0\\
        0 & 0
    \end{pmatrix},
\end{align*}
implying that the combined matrix $\bZ = \diag(z_1,\bZ_b, \dots \bZ_b)$, with $q$ copies of $\bZ_b$, satisfies $\bZ \succeq \bZero$ and $\tr_m(\bZ) \geq \eta(c^2 + q)$. The SDP objective formulas and
\eqref{eq:experiment-KyFan-epigraph} now show that both
objectives are at least $\eta\rho^2$.
By our choice of $C$, the multiplier $\eta$ must therefore lie in the bounded interval $[\gamma, C/\rho^2]$.

For DRO, boundedness of the objective and multiplier bounds
$t_1+q\tr(\bT_b)$.
Since $t_1\geq0$, $\bT_b\succeq\bZero$, and $q\geq1$,
both $t_1$ and $\bT_b$ must be bounded.

For MRO, boundedness of the objective and multiplier bounds
$m\tau+y_1+q\tr(\bY_b)$ from above. In the other direction, the constraints $y_1 \geq 0$ and $\bY_b \succeq \bZero$ imply that
\begin{align*}
    m\tau+y_1+q\tr(\bY_b) \geq m \tau.
\end{align*}
Further, the SDP constraints for $y_1$ and $\bY_b$ and the early observation that $\bZ \succeq \bZero$ imply that
\begin{align*}
    m\tau+ y_1+q\tr(\bY_b) &\geq m\tau + z_1 - \tau + q \tr(\bZ_b - \tau \bI_2)\\
    &\geq\tr(\bZ)-(1+2q - m)\tau\\
    &\geq -q \tau.
\end{align*}
Since $m,q>0$, these bounds control $\tau$ in both directions.
They then bound $y_1$ and $\tr(\bY_b)$ separately, since
$y_1\geq0$ and $\bY_b\succeq\bZero$.
Moreover,
\[
    0\leq z_1\leq y_1+\tau,
    \qquad
    \bZero\preceq\bZ_b\preceq\bY_b+\tau\bI_2,
\]
so the shifted blocks $z_1$ and $\bZ_b$ are bounded.
The identities
$t_1=z_1+\eta$ and
$\bT_b=\bZ_b+\eta1_2 1_2^\top$
then bound the remaining variables $t_1$ and $\bT_b$.

Thus, for either SDP, the restricted feasible set is bounded.
It is also nonempty and closed, because the constraints are
closed and the objective is continuous.
It is thus compact, and so the continuous objective attains
its minimum on this set.
\end{proof}

\subsection{Closed form for SNR}

For SNR, we obtain an explicit solution and closed form for the regret under $P$.

\begin{lemma}[Closed-form SNR solution]
\label{lem:experiment-SNR-closed-form}
Let $\lambda\defeq\rho^2$. Then the unique minimizer of $\bA \mapsto L(\bA,\tilde P)+\lambda\|\bA\|_\op^2$ is $\bA(\alpha_\lambda,\beta_\lambda)$, where
\begin{equation}
\label{eq:experiment-SNR-coefficients}
    (\alpha_\lambda,\beta_\lambda)
    =
    \begin{cases}
        \left(\dfrac{c}{c^2+\lambda},1\right),
        &0\leq\lambda\leq c(1-c),\\[3mm]
        (t_\lambda,t_\lambda),
        \quad
        t_\lambda\defeq\dfrac{q+c}{q+c^2+\lambda},
        &\lambda\geq c(1-c).
    \end{cases}
\end{equation}
Consequently, its exact clean regret is
\begin{equation}
\label{eq:experiment-SNR-regret}
    R(\SNR_\rho[\tilde P],P)
    =
    \begin{cases}
        \left(\dfrac{c}{c^2+\lambda}-1\right)^2,
        &0\leq\lambda\leq c(1-c),\\[3mm]
        m\left(\dfrac{q+c}{q+c^2+\lambda}-1\right)^2,
        &\lambda\geq c(1-c).
    \end{cases}
\end{equation}
The two expressions agree at $\lambda=c(1-c)$.
\end{lemma}

\begin{proof}
By our discussion in \cref{ssec:reduction-symmetry}, it suffices to minimize over
$\bA(\alpha,\beta)$.  Dropping the additive constant $q$ from
\eqref{eq:experiment-corrupted-risk}, the reduced objective is
\begin{equation}
\label{eq:experiment-SNR-reduced}
    F_\lambda(\alpha,\beta)
    =(c\alpha-1)^2+q(\beta-1)^2
     +\lambda\max\{\alpha^2,\beta^2\}.
\end{equation}
Recall that $c = m^{-1/4} \in (0,1]$. Because
$\bSigma_X^{\tilde P}=\diag(c^2,\bI_q)\succ0$, the risk is strongly
convex in the full matrix $\bA$.  Hence the regularized objective has a
unique minimizer.

Introduce an epigraph variable $t \geq 0$ with
constraints $|\alpha|\leq t$ and $|\beta|\leq t$.  For fixed $t$, the quadratic part of
\eqref{eq:experiment-SNR-reduced} is minimized by projecting its
unconstrained minimizers $\alpha = c^{-1}$ and $\beta = 1$ onto $[-t,t]$.  Since both are
positive, the constrained minimizers are
\[
    \alpha(t) \defeq\min\{t,c^{-1}\},
    \qquad
    \beta(t) \defeq \min\{t,1\}.
\]
Thus the epigraph problem reduces to minimizing
\[
    \phi_\lambda(t)
    \defeq
    \lambda t^2+(1-ct)_+^2+q(1-t)_+^2,
    \qquad t\geq0.
\]
Here $\lambda=\rho^2>0$ and $0<c<1$, so $\phi_\lambda$
is strictly convex.
For $t>c^{-1}$ it equals $\lambda t^2$ and increases,
so its minimizer lies in $[0,c^{-1}]$.
Moreover,
\[
    \frac12\phi_\lambda'(t)
    =
    \lambda t-c(1-ct)_+-q(1-t)_+,
    \qquad
    \frac12\phi_\lambda'(1)=\lambda-c(1-c).
\]

If $0<\lambda\leq c(1-c)$, the minimizer lies in
$[1,c^{-1}]$, where the derivative vanishes at
\[
    t=\frac{c}{c^2+\lambda}.
\]
The corresponding coefficients are
$\alpha=t$ and $\beta=1$.
If $\lambda\geq c(1-c)$, the minimizer lies in $[0,1]$,
where the derivative vanishes at
\[
    t=\frac{q+c}{q+c^2+\lambda}.
\]
In this case $\alpha=\beta=t$.
At $\lambda=c(1-c)$, both expressions give $t=1$.
This proves \eqref{eq:experiment-SNR-coefficients}.
Substitution into \eqref{eq:experiment-clean-regret}
proves \eqref{eq:experiment-SNR-regret}.
\end{proof}

\section{Full Experimental Details and Additional Experiments}
\label{app:experiment-extras}

The synthetic experiments for \cref{fig:plots} in \cref{sec:experiments} and \cref{ssec:experiment-varied-rank} below
were run in 8 hours on an \texttt{m8g.large} Amazon EC2
instance with 2 vCPUs and 8 GB of RAM, while the experiment for \cref{ssec:experiment-cesnet} was run in less than one hour on an M1 MacBook Air with an 8-core CPU and 16 GB RAM.
Code for the experiments and plots is available at \url{https://github.com/sbnietert/regret-optimal-regression}.

\subsection{Additional Synthetic Experiment with Varied Residual Rank}
\label{ssec:experiment-varied-rank}

For \textbf{Setting C}, we modify Setting A by fixing $p=m=200$ ($d=400$),
$n=10m^2$, and $\rho=\eps=1$, and varying the
residual effective rank over $10$ equally spaced values
$r\in[1,m-1]$. Specifically, for each $r$, the target law $P_r$ is
the law of $(X,Y)$, where
\[
    X\sim\cN(0,\bI_m),\qquad
    Y=X+U_r,\qquad
    U_r\sim\cN \left(
        0,\diag\!\left(0,1,\frac{r-1}{m-2}\bI_{m-2}\right)
    \right),
\]
and $X$ and $U_r$ are independent. Then
$\bA_{P_r}=\bSigma_X^{P_r}=\bI_m$, and the residual
second moment satisfies
\[
    \tr(\bSigma_E^{P_r})=r,\qquad
    \|\bSigma_E^{P_r}\|_\op=1,\qquad
    \reff(\bSigma_E^{P_r})=r.
\]
In particular, each $\kappa_{P_r}=3$.
The endpoint $r=m-1$ recovers the target law of Setting A
at $m=200$.

\begin{figure}[t]
    \centering
    \includegraphics[width=0.5\linewidth]{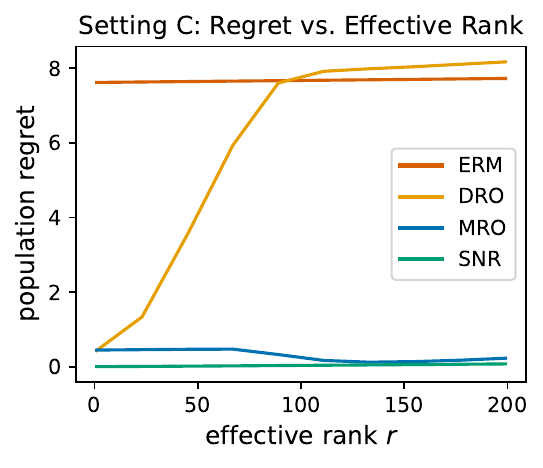}
    \caption{Target population regret
    $R(\hat{\bA}[P_{\mathrm{s},n}],P_r)$
    for Setting C, as a function of the target residual
    effective rank $r=\reff(\bSigma_E^{P_r})$.}
    \label{fig:plot-C}
\end{figure}

For each $r$, we draw $n$ i.i.d.\ target observations
and apply the same corruption as in Setting A.
We fit each estimator $\smash{\hat{\bA} \in \{\ERM,\DRO_\rho,\MRO_\rho,\SNR_\rho\}}$ to the resulting empirical
source law $P_{\mathrm{s},n}$, using the implementations
described in \cref{sec:experiments}. In \cref{fig:plot-C}, we plot the target population regret $R(\hat{\bA}[P_{\mathrm{s},n}],P_r)
    =\|\hat{\bA}[P_{\mathrm{s},n}]-\bI_m\|_\Frob^2$ against $r$. Curves report averages over $10$ independent
trials at each value of $r$, with error bars showing
the minimum and maximum over trials (again imperceptible due to low variance).

\subsection{Network-Traffic Experiment with CESNET Dataset}
\label{ssec:experiment-cesnet}

Finally, we test our estimators on a real-world network-traffic dataset, albeit with synthetically introduced Wasserstein perturbations. We employ the CESNET-TimeSeries24 dataset \citep{cesnettimeseries24}, which contains network-traffic time series data from the CESNET3 network connecting public and research institutions in the Czech Republic. Specifically, it records hourly, institution-level counts of packets and bytes transmitted over the network between October 2023 and July 2024, and we attempt to predict byte totals from packet counts. We use $268$ institutions with nonconstant counts and finite, nonnegative measurements for at least $99\%$ of the $6{,}717$ observed hourly timestamps, and intersect their valid timestamps, obtaining $n=6{,}228$ paired observations. For each response dimension $m\in\{8,16,24,32,48,64,96,128,192,256,268\}$, we take our target response vector to be the first $m$ byte-total channels and its corresponding feature vector to be the first $q=m/2$ matching packet-count channels plus an appended intercept term of $1$, so $p=q+1$. We center and normalize each physical channel according to its mean and standard deviation under the full empirical law of the raw observations, keeping these transformations and the observation set fixed throughout. Denote the resulting standardized target law by $P_m \in \cP_2(\R^{p + m})$, treated as a population measure to isolate the complexity of robust regression from that of handling sampling error.

This setup was selected so that some of the feature and response coordinates are highly correlated, while others are nearly independent, yielding a range of effective ranks. Coordinate-wise normalization is a relatively common preprocessing procedure, and it helps to keep the condition number of the target law from growing too quickly with $m$. In particular, the effective rank of $\bSigma_E^{P_m}$ grows roughly linearly with $m$, while the target condition number $\kappa_{P_m}$ grows more slowly, as shown in \cref{fig:cesnet-geometry}.

To obtain the corresponding corrupted source law $\tilde{P}_m$, we apply the frozen transformation above to the same $n$ observations, but with the first packet-count channel first multiplied by a gain of $0.5$ (all other channels are unchanged). By coupling observations according to their timestamps, one can directly upper bound $\Wtwo(P_m,\tilde{P}_m) \leq \eps \defeq 1.08679$, independent of $m$. We then fit each estimator $\hat{\bA} \in \{\ERM,\DRO_\eps,\MRO_\eps,\SNR_\eps\}$ to $\tilde{P}_m$, using the implementations described in \cref{sec:experiments}. In \cref{fig:cesnet-regret}, we plot the target regret $R(\hat{\bA}[\tilde{P}_m],P_m)$ against $m$. We observe qualitative similarities to the separation observed in our synthetic experiments, with MRO and SNR outperforming DRO and ERM as the dimension grows (although the gap is less pronounced). We emphasize that ample room remains for more refined empirical study, e.g., to identify the most realistic modes of data corruption for specific applications and to compare the performance of these estimators beyond the worst case.

\clearpage
\noindent
\begin{minipage}{\textwidth}
    \centering
    \captionsetup{hypcap=false}
    \includegraphics[width=0.5\linewidth]{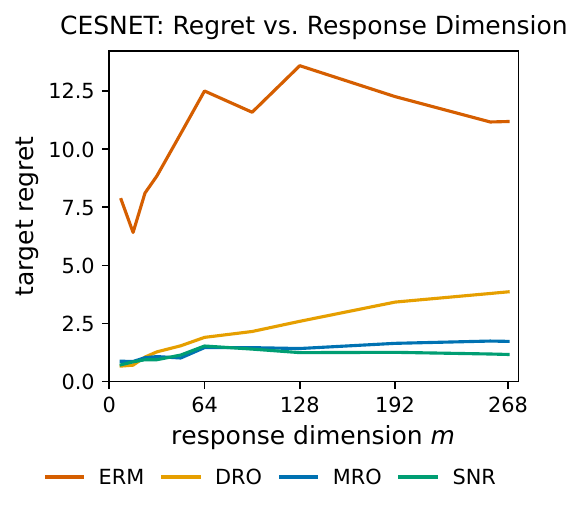}
    \captionof{figure}{Target regret $R(\hat{\bA}[\tilde{P}_m],P_m)$ for our CESNET experiment, as a function of the response dimension $m$.}
    \label{fig:cesnet-regret}

    \par\vspace{1cm}

    \includegraphics[width=\linewidth]{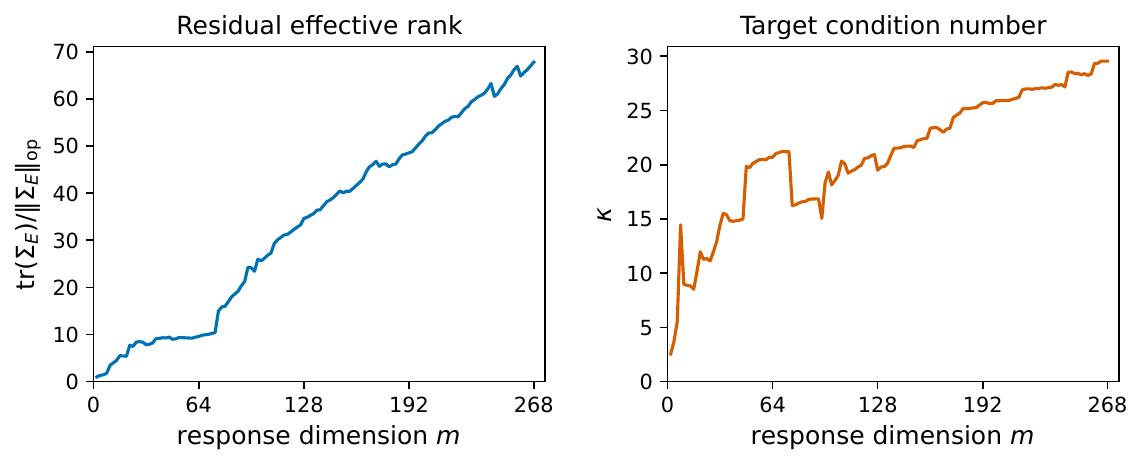}
    \captionof{figure}{Effective rank $\tr(\bSigma_E^{P_m})/\|\bSigma_E^{P_m}\|_\op$ and condition number $\kappa_{P_m}$ of the target law $P_m$ in our CESNET experiment, as a function of the response dimension $m$.}
    \label{fig:cesnet-geometry}
\end{minipage}

\end{document}